%% file: main.tex
\documentclass[conference,onecolumn,letterpaper,10pt]{IEEEtran}
\usepackage[T1]{fontenc}
\usepackage[utf8]{inputenc}
\usepackage{lmodern}
\usepackage{amsmath,amssymb,textcomp}
\usepackage{newunicodechar}
\usepackage{graphicx}
\usepackage{array,booktabs,longtable,ragged2e}
\usepackage{pdflscape}
\usepackage{afterpage}
\usepackage{fvextra}
\usepackage[letterpaper,width=430pt,height=696pt]{geometry}
\usepackage{flafter}
\usepackage[labelformat=empty,justification=justified,font=small]{caption}
\usepackage{url}
\usepackage{xcolor}
\definecolor{ReftLinkBlue}{HTML}{1155CC}
\usepackage[colorlinks=true,linkcolor=ReftLinkBlue,urlcolor=black]{hyperref}
\graphicspath{{graphics/}}
\makeatletter
\renewcommand\section{\@startsection{section}{1}{\z@}%
  {1.25\baselineskip plus 0.3\baselineskip minus 0.2\baselineskip}%
  {0.5\baselineskip plus 0.15\baselineskip minus 0.1\baselineskip}%
  {\normalfont\normalsize\centering\scshape}}
\renewcommand\subsection{\@startsection{subsection}{2}{\z@}%
  {1.0\baselineskip plus 0.25\baselineskip minus 0.2\baselineskip}%
  {0.4\baselineskip plus 0.15\baselineskip minus 0.1\baselineskip}%
  {\normalfont\normalsize\itshape}}
\makeatother

\makeatletter
\newlength{\ReftLandscapeColWidth}
\renewcommand*{\LS@rot}{%
  \setlength{\ReftLandscapeColWidth}{\wd\@outputbox}%
  \setbox\@outputbox\vbox{\hbox{\rotatebox{90}{%
    \vbox{\hsize\ReftLandscapeColWidth
      \box\@outputbox
      \ifdim\ReftLandscapeColWidth>\z@
        \vskip 52.3pt
        \hb@xt@\hsize{\hfil\reset@font\normalsize\thepage\hfil\hskip 19.2pt}%
      \fi}}}}%
  \let\@thefoot\@empty \let\@oddfoot\@empty \let\@evenfoot\@empty}
\makeatother

\makeatletter
\mathchardef\UrlBigBreakPenalty=\@M
\makeatother

\DefineVerbatimEnvironment{verbatim}{Verbatim}%
  {breaklines=true,breakanywhere=true,fontsize=\footnotesize}

\providecommand{\tightlist}{\setlength{\itemsep}{0pt}\setlength{\parskip}{0pt}}

\newcommand{\ReftTableCell}{\hyphenpenalty=10000\pretolerance=10000
                            \tolerance=1000\emergencystretch=0pt}

\newcommand{\ReftCaption}[3]{\caption[#2]{\textbf{#1.} \textit{#2.} #3}}
\newcommand{\ReftCaptionOf}[3]{\captionof{figure}{\textbf{#1.} \textit{#2.} #3}}

\newcommand{\ReftRef}[2]{\hyperref[#1]{#2}}

\newcommand{\ReftCutInRule}[1]{%
  \multispan{#1}{\leaders\hrule height\lightrulewidth\hfill\kern 0pt}}

\renewcommand{\footnoterule}{\kern-3pt\hrule width 0.4\columnwidth height 0.4pt \kern 2.6pt}
\newunicodechar{—}{---}
\newunicodechar{–}{--}
\newunicodechar{−}{\ensuremath{-}}
\newunicodechar{×}{\ensuremath{\times}}
\newunicodechar{†}{\textdagger{}}
\newunicodechar{°}{\textdegree{}}
\newunicodechar{…}{\ldots{}}
\newunicodechar{ü}{\"{u}}
\newunicodechar{→}{\ensuremath{\rightarrow}}
\newunicodechar{Δ}{\ensuremath{\Delta}}
\newunicodechar{α}{\ensuremath{\alpha}}
\newunicodechar{≥}{\ensuremath{\geq}}
\newunicodechar{≤}{\ensuremath{\leq}}
\hypersetup{pdftitle={Can LLMs in Draft-Verify-Revise Pipelines Resolve Deictic Ambiguity?},pdfauthor={Obinna I. Ekekezie, M.D.}}
\begin{document}
\title{Can LLMs in Draft-Verify-Revise Pipelines Resolve Deictic Ambiguity?}
\author{\IEEEauthorblockN{Obinna I. Ekekezie, M.D.\textsuperscript{a,b}}}
\maketitle
\thispagestyle{plain}\pagestyle{plain}
\begingroup
\renewcommand{\thefootnote}{}%
\footnotetext{\textsuperscript{a}Cambridge Health Alliance, Cambridge, MA, United States of America. \textsuperscript{b}Harvard Medical School, Boston, MA, United States of America. Correspondence to: <oekekezie@challiance.org>.\par\smallskip
\textit{Preprint. September 2026.}}%
\endgroup
\begin{abstract}
Draft-verify-revise is a common LLM orchestration pattern for scaling inference-time compute. One LLM drafts, a second critiques the draft and provides feedback, and a third uses that feedback to revise the draft into the final output. As context cascades between stages, LLMs at different stages can resolve a context-dependent expression such as ``previous'' differently. When that happens, the expression undergoes a deictic shift, a change in what it refers to. This phenomenon was studied with a synthetic dataset of 10 base examples, each rendered in three conditions. Holding the shared components constant, the conditions varied whether the draft stage LLM (the assistant) or the verify stage LLM (the grader) resolved the expression correctly, and how much independent reasoning the revise stage LLM (the meta-evaluator) needed to determine which reading was correct. Six models from three providers were tested across 21 reasoning effort configurations using e-values for sequential testing, in a primary experiment and an ablation experiment that removed error classification labels from the grader's feedback. A separate LLM analyzed the meta-evaluator's stated rationale for each wrong verdict. Balanced accuracy (the unweighted mean of sensitivity and specificity) ranged from 0.156, below chance, to near-perfect. GPT-5.2 rose from 0.156 without reasoning to 0.942 at its highest reasoning effort level, while Gemini 3 Pro stayed above 0.94 at every level. Gemini 3 Pro at low reasoning effort outscored GPT-5.2 at xhigh reasoning effort for roughly 5\% of the cost per trial. When the meta-evaluator erred, it tended to rely on surface cues rather than operational reasoning. Context engineers implementing draft-verify-revise pipelines should be wary of deictic shifts and make the intended referent explicit at each stage.
\end{abstract}
\section{1. Introduction}\label{introduction}

Scaling inference-time compute is a proven way to improve the output quality of large language models (LLMs) {[}1{]}. Chain-of-thought prompting was among the first widely used methods. Prompting a model to reason step by step makes it expend more compute before responding {[}2{]}. With reasoning models (LLMs trained through reinforcement learning to use chain-of-thought {[}3{]}, {[}4{]}), a newer approach is to raise the model's reasoning effort, a provider-specific parameter controlling how much inference-time compute it spends before responding. A complementary approach is to use an external verifier, which itself could be an LLM, to grade or critique the output. LLMs tend to favor their own generations when evaluating them {[}5{]}, and self-refinement amplifies that bias instead of correcting it {[}6{]}. They also struggle to produce reliable feedback on their own output {[}7{]}. Using a different model for verification than for generation helps mitigate these effects.

However, using an LLM as the external verifier is more complicated than adjusting reasoning effort, because it requires orchestrating multiple LLMs {[}8{]}. Such orchestrations require context management {[}9{]}. A relatively common LLM orchestration abstraction for scaling inference-time compute is the draft-verify-revise pattern. An LLM at the draft stage generates an initial response, an LLM at the verify stage critiques it and provides feedback, and an LLM at the revise stage incorporates any relevant feedback into the finalized response.

This pattern requires a context cascade: each stage passes its input and output to the next, where an LLM needs both to produce a grounded response.

The verifier could instead grade the draft against a rubric. But human-authored rubrics are expensive to produce and difficult to scale {[}10{]}, and naively generated LLM rubrics can degrade judge accuracy below that of using no rubric at all {[}11{]}. Accumulating context also risks ``context rot,'' in which performance degrades as an LLM's context window fills with too much information, causing the model to lose focus or fail to follow instructions {[}12{]}.

Orchestrating LLMs can introduce a different kind of vulnerability. In a deictic shift, context-dependent expressions such as ``the response,'' ``previous,'' and ``the instructions'' come to refer to different things as context passes from one stage to the next. Because a deictic expression takes its referent from the standpoint of the LLM interpreting it, LLMs at two stages can read the same term incompatibly, producing a perspectival disagreement.

Prior work has probed how LLMs handle such standpoint-dependent expressions in isolation. Zhang and Rayz {[}13{]} found that LLMs given an explicit temporal reference point (for example, relative to the experiencer's ``now'') produce systematic judgments of temporal similarity but handle past and future unevenly. Responses about the past range more widely and are less consistent. Responses about the future, by contrast, quickly flatten to uniform low similarity regardless of how distant the event is.

Raemaekers et al.~{[}14{]} tested four LLMs (Llama 3.3 70B, Llama 3.1 405B, GPT-OSS-20B, and GPT-OSS-120B) on short reasoning problems drawn from the relational reasoning and perspective-taking literature. The ``temporal'' problems ask the model to chain a before/after relation across invented three-letter syllables. The ``deictic'' problems require holding a stated point of view fixed across everyday perspective words (I, you, here, there, now, tomorrow). The models answered nearly all of the simplest problems correctly and stayed near ceiling on the temporal problems at every level of complexity. On the deictic problems, reversing the stated point of view reduced accuracy for most models, especially when combined with an invalid conclusion or an irrelevant premise, whereas Llama 3.1 405B was largely unaffected and achieved the highest deictic accuracy of the four.

In applied settings the consequences are concrete. In a clinical reasoning evaluation, 12 of 13 LLMs failed a vignette in which the patient's cough began before an ACE inhibitor was prescribed, most of them attributing the cough to the drug in violation of the constraint that cause must precede effect {[}15{]}. In a clinical document verification study, a non-reasoning model could not reconcile ``remained intubated on mechanical ventilation post-procedure'' with ``extubated the following day'' as temporally compatible, erroneously flagging the timeline as unsupported {[}16{]}. In a prospective feasibility study of Google's AMIE system taking clinical histories from 100 real patients, the system placed a past surgery in the future relative to the encounter {[}17{]}.

These are the sorts of failures anyone implementing a draft-verify-revise pipeline would hope it catches. A verifier should flag that a cause postdates its effect, or that a past event is placed in the future. But a verifier's flag is itself a claim that can be mistaken, and it creates a disagreement between the LLMs at the draft and verify stages. The safeguard works only if the LLM at the revise stage can settle such disagreements correctly. If it cannot, the pattern risks providing false confidence.

To my knowledge, no existing work tests this vulnerability in a controlled setting. The question is whether an LLM can reliably determine which reading was correct when it is given a disagreement in which the LLMs at two pipeline stages resolved the same context-dependent expression from different standpoints. This paper presents an experiment that isolates that decision, with no revised draft to produce, as a binary verdict on the assessment issued by the LLM at the verify stage (the grader), using synthetic stimuli that hold everything constant except the deictic shift. The experiment evaluates six models across 21 reasoning effort configurations to measure how model size and scaling inference-time compute affect the capability to settle such disagreements. An ablation experiment removes the grader's error classification labels from its feedback, and a separate analysis examines the stated rationale accompanying each wrong verdict.

\section{2. Methods}\label{methods}

\subsection{2.1 The Phenomenon}\label{the-phenomenon}

The experiment centers on a single deictic ambiguity that can arise naturally in draft-verify-revise pipelines. Consider a clinical decision support assistant whose system prompt states that ``all dosage adjustment recommendations must include a safety protocol specifying the previous dose to revert to if adverse effects occur.'' The assistant's user prompt contains a field labeled ``Previous Dose: 15mg'' alongside a ``Current Dose: 20mg.'' The assistant drafts a recommendation proposing a titration to 25mg, so its safety protocol must specify the previous dose to revert to. In this setting, the word ``previous'' functions simultaneously as a relative temporal expression and as a field label.

The assistant resolves ``previous'' as the dose in effect before its proposed titration to 25mg and targets 20mg. The grader resolves ``previous'' by matching the labeled field and targets 15mg. Neither resolution is linguistically unreasonable; the question is which one is operationally correct, and answering that question requires tracking whose temporal standpoint governs the safety protocol's fallback clause. The word ``previous'' is a relative temporal adjective, not a prototypical deictic term such as ``now'' or ``here.'' Resolving it is nonetheless a deictic operation on the now-then axis {[}14{]}, because it depends on which stage's temporal standpoint serves as the reference point. When the assistant and grader adopt different standpoints, they produce the perspectival disagreement that the LLM at the revise stage must settle.

\subsection{2.2 Three-Role Architecture}\label{three-role-architecture}

The experiment uses three roles, mapping onto the three stages of a draft-verify-revise pipeline. An assistant generates a draft including a fallback clause. A grader verifies the draft and produces structured error feedback. A meta-evaluator (the LLM under test) receives all prior context and renders a binary verdict on whether the grader's assessment was ``correct'' or ``incorrect.'' This is a judgment task, not a detection task, because the grader's feedback already claims the assistant's fallback clause targets the wrong value. The meta-evaluator also produces a structured rationale, the input to the rationale analysis probe (Methods: Rationale Analysis). Of the three roles, only the meta-evaluator is filled by an LLM. The assistant's draft, the grader's feedback, and all other stimulus components are fixed text held constant across trials. Any performance variation therefore reflects the meta-evaluator's reasoning, not noise from earlier stages. \ReftRef{art:figure_1}{Figure~1} illustrates this architecture across the three conditions defined in Methods: Stimulus Design.

\input{items/synthetic_example_diagram}

\subsection{2.3 Stimulus Design}\label{stimulus-design}

The dataset comprises 30 stimuli: 10 base examples, each rendered in three conditions. Each stimulus reconstructs a complete draft-verify-revise transcript from the meta-evaluator's vantage point. In that transcript, the assistant received its system prompt and user prompt and produced a draft, and the grader received those same prompts plus the draft and issued structured feedback. All of that context forms the body of the meta-evaluator's own user prompt. Supplementary~Figure~S1 depicts this cumulative structure.

The minimal-pair design isolates the deictic shift within this cascade. For each base example, the three conditions share the assistant's system prompt, user prompt, and proposed change verbatim. The only components that vary are the fallback clause in the assistant's draft and the grader's feedback text. Any accuracy difference between conditions therefore primarily reflects how the meta-evaluator resolved the perspectival disagreement, not incidental differences in wording elsewhere in the transcript. Supplementary~Methods~SM1 lists which components are held constant and which vary.

Every base example states a current value and a distinct earlier value, so the scenario's own facts settle which value the fallback should target. In the dosage example, the assistant's user prompt includes ``Previous Dose: 15mg once daily (before last titration 6 weeks ago)'' and, in the clinical history, ``Titrated from 15mg to 20mg six weeks ago.'' Together these details fix 20mg as the dose in effect when the assistant proposed raising it.

The three conditions differ in whether the assistant or the grader resolved ``previous'' correctly. Where the grader was right, the conditions also differ in how visible the grader's operational reasoning is, that is, whether the grader's analysis states that reverting means returning to the value in effect before the assistant's proposed change. In the correct-draft condition, the assistant's fallback targets the current value (operationally correct) but the grader flags it as an error. Ground truth is therefore ``incorrect'' (the assistant was correct and the grader was wrong), and this condition tests specificity. In the incorrect-transparent condition, the fallback targets the historical field value (operationally wrong) and the grader flags it and states the operational reasoning outright. Ground truth is therefore ``correct'' (the grader was right), and settling the disagreement requires only following the grader's reasoning. In the incorrect-opaque condition, the same incorrect draft is paired with a grader whose analysis is structurally identical to the correct-draft grader's. Ground truth is again ``correct,'' but because that analysis never states that reasoning, the meta-evaluator has to arrive at it on its own. Together, the two incorrect-draft conditions test sensitivity.

The transparent--opaque accuracy gap is a designed diagnostic. Both conditions share the same draft, error, and correct answer, so an accuracy difference between them isolates the contribution of the grader's surface cues versus the meta-evaluator's independent reasoning.

\subsection{2.4 Models and Configurations}\label{models-and-configurations}

The experiment evaluated a smaller and a larger model from each of three providers: GPT-5-mini {[}18{]} and GPT-5.2 {[}19{]} from OpenAI, Claude Haiku 4.5 {[}20{]} and Claude Opus 4.6 {[}21{]} from Anthropic, and Gemini 3 Flash {[}22{]} and Gemini 3 Pro {[}23{]} from Google. Each model was run at between two and five reasoning effort levels, for 21 configurations in total (Supplementary~Methods~SM2). Each pairing of a model with a reasoning effort level is one configuration. All six models are closed-weight, so parameter counts are unknown; the ``smaller'' and ``larger'' designations are inferred from each provider's pricing tiers and naming conventions. Five of the six models are independent (they had no role in constructing the stimuli) and provide primary evidence. The stimuli were authored using Claude Opus 4.5 and Claude Opus 4.6; only Claude Opus 4.6 is among the models tested, and it is referred to as the authoring model throughout. Its results may reflect familiarity with its own output patterns rather than the capability being measured, since LLMs can recognize and favor their own generations {[}5{]}, {[}6{]}. The authoring model is therefore flagged with a dagger (†). Claude Haiku 4.5 was developed by Anthropic as well, though it played no part in writing the stimuli. Two models from the same developer can share stylistic habits, so Claude Haiku 4.5's independence is not as clean as that of the models from OpenAI and Google.

\subsection{2.5 Primary and Ablation Experiments}\label{primary-and-ablation-experiments}

The primary experiment crosses 21 configurations with 30 stimuli, yielding 630 configuration-stimulus pairings. This paper calls each such pairing a cell. Each cell accumulates independent binary trials.

The ablation experiment uses the same 21 configurations and 30 stimuli but removes the grader's four error classification fields (\texttt{error\_\allowbreak{}category}, \texttt{error\_\allowbreak{}subtype}, \texttt{risk\_\allowbreak{}level}, and \texttt{is\_\allowbreak{}issue\_\allowbreak{}truly\_\allowbreak{}error}). The first three follow the structure of MedVAL's {[}24{]} physician-defined taxonomy of risk levels and error categories; the fourth asserts that the flagged issue is a real error. The grader's \texttt{analysis} text, \texttt{root\_\allowbreak{}cause}, and evidence arrays are retained verbatim (the ablation panel of \ReftRef{art:figure_1}{Figure~1} shows the ablated stimulus). The hypothesis is that error classification labels (particularly \texttt{error\_\allowbreak{}category:\ "hallucinations"}) may distract the meta-evaluator from the grader's substantive analysis, so that removing them restores engagement with the evidence on its merits.

Both experiments were run using a custom multi-provider batch inference system that abstracts over the batch APIs of OpenAI, Anthropic, and Google. Supplementary~Methods~SM3 describes the system's architecture.

\subsection{2.6 Statistical Approach}\label{statistical-approach}

Each trial produces a binary outcome. The meta-evaluator's verdict either matches ground truth or does not. Cell accuracy is the proportion of trials matching ground truth; condition accuracy pools all cells within a condition for a given configuration. Correct-draft accuracy is specificity (correctly rejecting the grader's mistaken claim that the assistant erred). Sensitivity (correctly accepting the grader's claim when the assistant did err) is measured separately within each incorrect-draft condition, yielding transparent accuracy and opaque accuracy.

Raw accuracy pooled across all 30 stimuli would be misleading. A meta-evaluator that always accepted the grader's claim without reasoning about the deictic shift at all would score 20/30 on raw accuracy by exploiting the 1:2 class imbalance alone. Balanced accuracy (the unweighted mean of sensitivity and specificity, where sensitivity pools the two incorrect-draft conditions) corrects the imbalance exactly. A stimulus-blind strategy is one whose rate of accepting the grader's claim does not depend on the stimulus; every such strategy has an expected balanced accuracy of 0.5, and the two that never vary, one always accepting the grader's claim and the other always rejecting it, score exactly 0.5 on any set of trials. Balanced accuracy is the primary performance metric throughout.

Balanced accuracy collapses transparent and opaque accuracy into a single pooled sensitivity term, so the transparent--opaque gap lives in the three condition accuracies rather than in the headline. Those three also carry the strategy diagnosis, since near-zero specificity alongside transparent and opaque accuracy near 1.000 is the signature of accepting the grader's claim almost regardless of the stimulus. A gap between transparent and opaque accuracy also certifies that the verdicts responded to the stimulus at all, because a stimulus-blind strategy accepts the grader's claim at the same rate in both incorrect-draft conditions and therefore expects the same accuracy in each. The transparent-versus-opaque comparison tests that equality; the size of a gap is described rather than tested.

Overall accuracy, pooled hits over pooled valid trials across the three conditions (the raw accuracy above), is reported alongside balanced accuracy as a descriptive column. At equal valid counts it equals the unweighted mean of the three condition accuracies, so giving each condition one-third weight still leaves the incorrect-draft class two-thirds of the weight: the 1:2 imbalance survives. A stimulus-blind strategy that accepts the grader's claim with probability \(q\) expects \((1 + q) / 3\), so the whole zero-information band from one-third to two-thirds is available without reading a stimulus. For a configuration with 600 valid trials, those endpoints correspond exactly to 200 and 400 verdicts matching ground truth, respectively. A value inside the band does not establish that the meta-evaluator read the stimulus; a value outside it, above or below, is beyond what every stimulus-blind strategy expects.

Data collection is sequential: trials accumulate in batches, and whether to stop or extend collection depends on the results observed so far. Conventional p-value testing does not apply under this design. P-values require a sample size fixed in advance, and inspecting them as data accumulates (``peeking'') inflates false positive rates.

Both experiments use e-values instead, which remain valid regardless of when data collection stops, eliminating the corrections that sequential p-value analyses require {[}25{]}. An e-value quantifies evidence against a null hypothesis as the factor by which one's wealth would grow from betting against a false null. For example, an e-value of 20 means the data are 20 times more consistent with the alternative than with the null.

Each iteration of data collection gives every cell a batch of 20 binary trials tested against the null that accuracy equals chance (\(p_0 = 0.5\)). A cell is rejected when its e-value reaches 20 (corresponding to \(\alpha = 0.05\)) and declared futile when it drops to 0.05 or below; a cell that reaches neither boundary can be extended in a subsequent iteration. Per-cell accuracy estimation uses two-sided confidence sequences, which, similarly, are valid regardless of when data collection stops. The e-values and confidence sequences reported here measure how each configuration performed on these 30 stimuli, not how it would perform on other stimuli of the same kind. For that broader question, the effective sample size is the 10 base examples per condition.

Further tests address questions the per-cell e-values cannot. The condition comparison tests ask whether a single configuration's accuracy differs between two conditions; the transparent-versus-opaque comparison is the formal test of the transparent--opaque gap. Within each configuration the three condition comparisons are tested as one family: the configuration's global null e-value is the equal-weight mean of its three pairwise e-values, and a comparison counts as significant only when its own e-value and the global null e-value both reach 20. The degradation tests ask whether balanced accuracy falls when a model is run at a higher reasoning effort level, since more inference-time compute is not guaranteed to help. The model size tests ask whether a provider's larger model outperforms its smaller one on balanced accuracy, which is the question of whether model size buys this capability. The cross-dataset tests ask whether a configuration's balanced accuracy and its three condition accuracies differ between the primary experiment and the ablation experiment, which measures the effect of removing the error classification labels. Supplementary~Methods~SM4 specifies the e-value constructions, these tests, and the confidence sequences.

\subsection{2.7 Rationale Analysis}\label{rationale-analysis}

The rationale analysis uses a ``probe,'' a separate LLM, to analyze the meta-evaluator's stated rationale on trials whose verdict did not match ground truth (Supplementary~Methods~SM5). It analyzes the stated rationale rather than the reasoning trace for two reasons. First, OpenAI, Anthropic, and Google expose only reasoning summaries through their LLM inference APIs. Second, an LLM's reasoning trace does not reliably reflect its internal computations {[}26{]}, {[}27{]} and can be unfaithful to its final output {[}28{]}.

The probe applies seven binary flags to the stated rationale. Two of them discriminate the failure modes. \emph{Articulated operational interpretation} is positive when the rationale expressed the view that reverting means returning to the value in effect before the assistant's proposed change, even if that view was subsequently dismissed. \emph{Operational interpretation governed judgment} is positive when that same interpretation determined the final verdict. The remaining five are misattribution flags, used for deconfounding: they separate rationales that misreport a value or claim supplied by the trial itself from rationales that reflect genuine interpretive failure. Each fires when the rationale misreports a specific value or claim: the scenario's current, historical, or proposed value; the value the assistant cited in its fallback clause; or the grader's own claim.

The two discriminating flags carry different diagnostic implications depending on the condition. On the incorrect-draft conditions, where the grader was right to flag the draft, reaching the correct verdict by resolving the deictic ambiguity meant letting the operational interpretation surface to contest the value drawn from the field labeled ``Previous {[}x{]}.'' Whether the meta-evaluator articulated that interpretation is therefore the key discriminator on these conditions. When its verdict did not match ground truth on an incorrect-draft stimulus, the meta-evaluator either never expressed the operational interpretation, expressed it but deferred to field label matching, or let it govern the verdict yet still reached the wrong conclusion.

On the correct-draft condition, ground truth is ``incorrect'' because the assistant's draft correctly targets the current value, so the meta-evaluator should have rejected the grader's claim. Articulation alone is uninterpretable here. That current value is both the operationally correct reversion target and the value in the labeled ``Current {[}x{]}'' field of the assistant's user prompt. A rationale naming the right value could therefore have arrived there by reading the field label rather than by operational reasoning. Whether the operational interpretation governed the verdict resolves this confound, because genuine operational reasoning connects the assistant's proposed change to the reversion target the meta-evaluator endorses, while field label matching identifies a value against a labeled field without referencing that change.

Among trials whose verdict did not match ground truth, the \emph{articulation rate} is the percentage in which the meta-evaluator expressed the operational interpretation. The \emph{governing rate} is the percentage of those same trials in which the operational interpretation determined the meta-evaluator's verdict. Both rates are always reported within a single condition.

\section{3. Results}\label{results}

\subsection{3.1 Experiment Overview}\label{experiment-overview}

Each experiment ran 21 configurations across 30 stimuli, producing 630 cells and 12,600 trials. In both experiments, a single iteration was sufficient for most cells to reach statistical resolution, that is, either the rejection or the futility boundary (Supplementary~Figure~S2). The cells that reached neither boundary were not extended; per-condition counts for the rejection boundary of the per-cell tests are reported in Results: Per-Condition Accuracy. Of the 21 configurations, 17 used the five independent models and 4 used the authoring model (Claude Opus 4.6†).

\subsection{3.2 Performance: Balanced Accuracy}\label{performance-balanced-accuracy}

The top independent configuration in the primary experiment was Gemini 3 Pro at low reasoning effort (balanced accuracy 0.965), followed by GPT-5.2 at xhigh reasoning effort (0.942) and Gemini 3 Pro at high reasoning effort (0.941) (Supplementary~Table~S1). The authoring model achieved perfect balanced accuracy at two of its four reasoning effort levels (Opus 4.6† medium and high: 1.000) and near-perfect scores at the other two (low: 0.979, max: 0.980). The authoring model's scores, however, may reflect familiarity with its own output patterns (Methods: Models and Configurations).

At the other end of the balanced accuracy leaderboard (Supplementary~Table~S1), GPT-5.2 with reasoning effort set to none scored 0.156 and GPT-5-mini with reasoning effort set to minimal scored 0.307, both below the 0.5 every stimulus-blind strategy is expected to score (Methods: Statistical Approach). Their overall accuracies fall on opposite sides of the zero-information band's lower edge. GPT-5.2 without reasoning matched ground truth on 125 of its 600 valid trials, an overall accuracy of 0.208 that falls below the entire band. GPT-5-mini at minimal reasoning effort matched ground truth on 243 of its 600 valid trials, an overall accuracy of 0.405 that sits inside the band and is therefore uninformative on its own; its 0.770 gap between transparent and opaque accuracy is what certifies that its verdicts responded to the stimuli (Results: Per-Condition Accuracy). In the ablation experiment, Gemini 3 Pro at low reasoning effort remained the top independent configuration (0.971), and the independent-model range, the spread between the best and worst independent configurations' balanced accuracy, narrowed from 0.809 to 0.741 (Supplementary~Table~S1).

The balanced accuracy difference between the primary and ablation experiments was significant for 4 of 21 configurations under the cross-dataset tests (paired testing-by-betting e-values, e-Bonferroni corrected within each family over both directional nulls; Supplementary~Methods~SM4). All four differences favored the ablation experiment (Supplementary~Figure~S3): Gemini 3 Flash at minimal reasoning effort (+0.042), GPT-5-mini at minimal reasoning effort (+0.047), GPT-5.2 with reasoning effort set to none (+0.074), and GPT-5.2 at high reasoning effort (+0.115) (Supplementary~Table~S2). Two of the four cleared the rejection threshold of 20 only narrowly, at adjusted e-values of 31.025 for Gemini 3 Flash and 30.439 for GPT-5-mini, while the two GPT-5.2 configurations reached \(2.676 \times 10^{5}\) (with reasoning effort set to none) and 1798.362 (at high reasoning effort).

\subsection{3.3 Per-Condition Accuracy: Specificity and Sensitivity}\label{per-condition-accuracy-specificity-and-sensitivity}

In the primary experiment, specificity varied widely across configurations, from 0.000 for GPT-5.2 with reasoning effort set to none to a perfect 1.000 for the authoring model at medium and high reasoning effort. Sensitivity on transparent stimuli was high for all but the weakest configurations. Of the 21 configurations, 14 scored a perfect 1.000 and 17 scored at least 0.975. Only GPT-5.2 with reasoning effort set to none (0.425) and Claude Haiku 4.5 with reasoning off (0.830) fell below 0.925. Sensitivity on opaque stimuli is the more demanding of the two sensitivities, because the grader's analysis provides no surface cues. It ranged from 0.200 for GPT-5.2 with reasoning effort set to none to a perfect 1.000 for Claude Haiku 4.5 with reasoning enabled and for three of the authoring model's four reasoning effort levels. \ReftRef{art:figure_2}{Figure~2} shows all three per-condition accuracies for each configuration in both experiments. The same pattern held in the ablation experiment: specificity again varied widely, from 0.005 for GPT-5-mini at minimal reasoning effort to 1.000 for the authoring model at low and medium reasoning effort; sensitivity on transparent stimuli was a perfect 1.000 for 18 of 21 configurations; and sensitivity on opaque stimuli remained the more demanding of the two sensitivities, falling to 0.280 for GPT-5.2 with reasoning effort set to none. Accuracy also varied across base examples within each condition, widely on correct-draft and opaque stimuli and narrowly on transparent stimuli, and the base example with the lowest accuracy in each condition was the same in both experiments: \texttt{example\_\allowbreak{}10} on correct-draft stimuli, \texttt{example\_\allowbreak{}09} on transparent stimuli, and \texttt{example\_\allowbreak{}04} on opaque stimuli (Supplementary~Figures~S4a--c).

\input{items/condition_accuracy_paired}

Under the family rule (Methods: Statistical Approach), 38 of the primary experiment's 63 pairwise condition comparisons (21 configurations, 3 comparisons each, one family per configuration) were significant, as were 41 of the ablation experiment's 63 (Supplementary~Methods~SM4; Supplementary~Table~S3). The transparent-versus-opaque comparison was significant in 11 of 21 configurations in the primary experiment and in 12 in the ablation experiment, the same 11 plus Gemini 3 Pro at low reasoning effort; of the 9 configurations significant in neither, all held transparent and opaque accuracy at or above 0.975 in both experiments. Under the per-cell tests, the chance null was rejected in fewer correct-draft cells than in either incorrect-draft condition: 128 of the primary experiment's 210 correct-draft cells (61.0\%), against 199 of 210 transparent cells (94.8\%) and 152 of 210 opaque cells (72.4\%); the ablation experiment repeated the pattern, with correct-draft 133 of 210 (63.3\%), transparent 201 (95.7\%), and opaque 161 (76.7\%; Supplementary~Table~S4).

Between the primary and ablation experiments, 3 of 63 configuration-condition comparisons reached significance under the cross-dataset tests (Supplementary~Methods~SM4), all on opaque stimuli and all favoring the ablation experiment: Gemini 3 Flash at minimal reasoning effort (+0.110), GPT-5-mini at minimal reasoning effort (+0.190), and GPT-5.2 at high reasoning effort (+0.140) (Supplementary~Table~S5). Beyond significance, the opaque accuracy differences between the experiments mostly favored the ablation experiment. Of the 17 non-ceiling opaque comparisons (configurations whose opaque accuracy in the primary experiment was below 1.000), 14 showed the ablation experiment improving accuracy (Supplementary~Figure~S5). Correct-draft directionality was weaker, with 12 of the 19 non-ceiling correct-draft comparisons improving and 7 degrading under the ablation experiment. Transparent comparisons were dominated by ceiling effects: only 7 of the 21 configurations had transparent accuracy below 1.000 in the primary experiment, and all 7 improved under the ablation experiment.

\subsection{3.4 Scaling Effects}\label{scaling-effects}

In the primary experiment, GPT-5.2 showed the widest balanced accuracy range across its reasoning effort levels, from 0.156 without reasoning to 0.942 at xhigh reasoning effort (a range of 0.786). By contrast, Gemini 3 Pro and the authoring model (Opus 4.6†) were comparatively robust across their respective reasoning scales, with ranges of 0.024 and 0.021 (\ReftRef{art:figure_3}{Figure~3}). For several models, the marginal benefit of additional reasoning effort diminished sharply at higher levels, and balanced accuracy dipped at some transitions. Adjacent-pair reasoning effort level degradation tests (one-sided paired testing-by-betting e-values, e-Bonferroni corrected with K = 15; Supplementary~Table~S6) found none of the 15 transitions significant in either experiment; at 11 of the 15 transitions in the primary experiment and 10 in the ablation experiment the accumulated paired differences never favored the lower level, so those one-sided tests finished with no evidence in either direction (Supplementary~Methods~SM4). Mean reasoning tokens per trial rose with each reasoning effort level, apart from Gemini 3 Flash from minimal to low, where the provider reported no reasoning tokens at either level (Supplementary~Figure~S6; full token composition, Supplementary~Figure~S7).

\input{items/reasoning_effort_vs_balanced_accuracy_paired}

Reasoning token growth translates directly into cost. In the primary experiment, GPT-5.2 at xhigh reasoning effort was the most expensive configuration per trial by a wide margin, at nearly three times the next highest, and the spending did not buy the best accuracy. GPT-5.2 spent 4.6 million total reasoning tokens, driving a per-trial cost of \$0.0556 for a balanced accuracy of 0.942. Gemini 3 Flash at medium reasoning effort reached 0.894 with 660,000 total reasoning tokens at \$0.0022 per trial, roughly 4\% of the cost. Gemini 3 Pro at low reasoning effort outscored both, reaching 0.965 at \$0.0030 per trial. \ReftRef{art:figure_4}{Figure~4} plots cost per trial on a log scale against balanced accuracy, with a Pareto frontier. The marginal cost per percentage point of balanced accuracy varied dramatically across reasoning effort level transitions. GPT-5.2's step from high to xhigh reasoning effort was the most expensive improving transition, at \$0.0020 per percentage point. Three transitions decreased balanced accuracy: Claude Opus 4.6† from high to max reasoning effort, Gemini 3 Pro from low to high, and Gemini 3 Flash from medium to high (Supplementary~Figure~S8). Full per-configuration cost breakdowns are in Supplementary~Table~S7; in total, the primary experiment cost \$89.17 and the ablation experiment \$83.23.

\input{items/cost_per_trial_vs_balanced_accuracy_paired}

The remaining scaling comparison is model size. Within each provider, two one-sided paired testing-by-betting tests compared the larger model against the smaller model (e-Bonferroni corrected, K = 6). Neither test selects a configuration on either side: every pairing of one of the larger model's reasoning effort levels with one of the smaller model's configurations contributes one e-process, and the two tests combine that one set of processes differently. The every-level statistic takes the minimum over levels of the mean over configurations; the some-level statistic takes the mean over levels of the minimum over configurations (summary tests in Supplementary~Tables~S8 and S9; per-level e-values in Supplementary~Table~S10; Supplementary~Methods~SM4). Quantifying over the smaller model's measured configurations, rather than over reasoning effort levels matched by name across the two models, avoids assuming those levels are comparable, because performance is not monotonic in reasoning effort. In the primary experiment, four of six comparisons reached significance, all favoring the larger model. For the Google pair (Gemini 3 Flash compared with Gemini 3 Pro) and the Anthropic pair (Claude Haiku 4.5 compared with Claude Opus 4.6†), both tests reached significance. In each pair, every reasoning effort level of the larger model outperformed the smaller model's worst configuration, and some reasoning effort level outperformed the smaller model's best configuration. The OpenAI models tested were the exception, significant in neither test. GPT-5.2 with reasoning effort set to none reached a balanced accuracy of 0.156, below every GPT-5-mini configuration, the weakest of which scored 0.307. That level accumulated no evidence of an advantage over any of them, and because the every-level statistic takes the minimum over levels, it held the statistic at its neutral value of 1.000, an adjusted e-value of 0.167 once the K = 6 correction is applied, against the threshold of 20. Every GPT-5.2 level was likewise matched by some GPT-5-mini configuration against which it accumulated little evidence, so the some-level statistic, which averages those weakest results, reached an adjusted 0.282. The ablation experiment replicated this pattern, at adjusted e-values of 0.167 and 0.581.

\subsection{3.5 Failure Mode Analysis}\label{failure-mode-analysis}

The rationale analysis probe classifies each trial whose verdict did not match ground truth by whether the meta-evaluator expressed the operational interpretation and whether that interpretation determined the verdict (Methods: Rationale Analysis).

The rationale analysis covered the 2,383 trials whose verdict did not match ground truth in the primary experiment and the 2,024 in the ablation experiment, out of 12,600 trials each.

In the primary experiment, on correct-draft stimuli, 1,310 trials produced verdicts that did not match ground truth. On 870 (66.41\%) the meta-evaluator never expressed the operational interpretation at all; on 431 (32.90\%) it expressed the interpretation but let field label matching govern; on only 9 did the operational interpretation govern (0.69\% governing rate). Field label matching, not the operational interpretation, almost always determined the verdict (Supplementary~Figure~S9a). For example, in one of these correct-draft trials, the operational interpretation never surfaced: the meta-evaluator called the draft's fallback to the current value a substantive mistake because the user prompt labeled that value as current rather than previous, and accepted the grader's mistaken claim on that basis (Supplementary~Exemplar~S1). In another of these trials, the operational interpretation surfaced and was set aside: the meta-evaluator stated that reverting to the current value was the correct action in a real-world deployment and still accepted the grader's mistaken claim, because the draft had called that value the previous one, contradicting the labels in the user prompt (Supplementary~Exemplar~S2).

On transparent stimuli, 178 trials produced verdicts that did not match ground truth; the articulation rate was 100\% and the governing rate was 1.12\% (2 of 178). In 176 of those 178 trials, field label matching governed the verdict: the capability was present but dismissed. In one of those trials, the meta-evaluator granted that a real-world rollback might target the value in effect immediately before the change, then rejected the grader's correct claim because the task distinguished the current value from the previous one and required the draft to cite the previous one (Supplementary~Exemplar~S3).

On opaque stimuli, 895 trials produced verdicts that did not match ground truth. The meta-evaluator expressed the operational interpretation on 544 of them (60.78\% articulation rate) and dismissed it on all but 23 of those 544 trials, for a 2.57\% governing rate (23 of 895). In one trial where the interpretation was dismissed, the meta-evaluator observed on its own that reverting usually means returning to the value immediately preceding the change, the current value, then held that choosing the value labeled previous was a reasonable reading of the user prompt's data structure and rejected the grader's correct claim (Supplementary~Exemplar~S4). On the other 351 trials, the meta-evaluator never expressed the operational interpretation at all. In one of them, it rejected the grader's correct claim because the value the draft cited was the one the user prompt labeled as previous and the requirement asked for the previous setting (Supplementary~Exemplar~S5). The transparent--opaque articulation gap mirrored the transparent--opaque accuracy gap reported in Results: Per-Condition Accuracy.

The ablation experiment's largest effect was on the opaque condition. Opaque articulation rate rose from 60.78\% (544 of 895) to 68.72\% (468 of 681), a change of +7.94pp, and opaque governing rate rose from 2.57\% (23 of 895) to 6.02\% (41 of 681), a change of +3.45pp (Supplementary~Table~S11). Transparent articulation remained saturated at 100\%, and correct-draft governing stayed near zero (9 of 1,310 trials in the primary experiment and 10 of 1,204 in the ablation experiment).

A separate set of probe flags checks whether the stated rationale misreports a value or claim the trial supplies (Methods: Rationale Analysis). On opaque stimuli, 188 of the primary experiment's 895 trials whose verdict did not match ground truth (21.0\%) carried at least one such misreport, as did 172 of the ablation experiment's 681 (25.3\%). In almost every case the misreport was a misreading of the grader's own claim (186 and 169, respectively). The per-condition breakdown across all five flags is in Supplementary~Table~S12. When trials carrying any such misreport were excluded, the opaque articulation rate decreased from 60.78\% to 58.27\% (412 of the 707 remaining trials) in the primary experiment and from 68.72\% to 67.19\% (342 of 509) in the ablation experiment, decreases of 2.51 and 1.53 percentage points, respectively. Those misreport rates are themselves likely overstated: the grader-claim flag, which accounted for almost all of the misreports, was the one flag a post hoc probe audit (reported below) most often judged to have fired in error. The audit read most flagged trials as faithful disagreements with the grader rather than misreports (Supplementary~Methods~SM5).

In the post hoc audit, a Claude Code agent independent of the probe and the stimuli re-classified 240 trials in each experiment, approximately 10.1\% of the primary experiment's 2,383 trials whose verdict did not match ground truth and 11.9\% of the ablation experiment's 2,024 such trials. On the two flags that discriminate the failure modes, its classifications agreed with the probe's on at least 93.8\% of audited trials in every condition of both experiments. That agreement indicates that the probe applied its criteria as written; it does not indicate that those classifications are correct in absolute terms (Supplementary~Methods~SM5).

At the configuration level (Supplementary~Table~S13), Claude Haiku 4.5 with reasoning off had 0.444 balanced accuracy in the primary experiment, and 312 of its trials produced verdicts that did not match ground truth. It had the highest governing rate on opaque stimuli of any configuration, 15.17\% (22 of 145 such opaque trials), alongside a 67.59\% opaque articulation rate (98 of 145). In one of those trials, it resolved ``previous'' correctly, agreed that the draft had erred, and still called the grader's assessment incorrect because the grader's analysis, while reaching the right conclusion, had conflated the two senses of ``previous'': the value in the labeled field and the value in effect before the proposed change (Supplementary~Exemplar~S6). In another, it resolved ``previous'' correctly but misread the draft as citing the current value, when the draft had cited the value in the labeled previous field, and on that basis called the grader's assessment incorrect (Supplementary~Exemplar~S7). In the ablation experiment its opaque governing rate rose to 34.45\% (41 of 119), the largest shift of any configuration. By contrast, in the primary experiment, GPT-5.2 with reasoning effort set to none (0.156 balanced accuracy, 475 trials whose verdict did not match ground truth) reached a comparable 68.75\% opaque articulation rate (110 of 160 such opaque trials) but a 0\% governing rate in each of the three conditions, deferring systematically to field label matching instead. On correct-draft stimuli, the only configurations with governing above zero were GPT-5-mini configurations (7.41\% at low reasoning effort, 8 of 108 trials; 1.45\% at medium reasoning effort, 1 of 69), accounting for all 9 correct-draft trials in which the operational interpretation governed the verdict.

Across base examples, the failure mode split among trials whose verdict did not match ground truth barely moved on correct-draft stimuli: the highest correct-draft governing rate for any base example was 4.48\% (3 of 67 trials, on \texttt{example\_\allowbreak{}02}) in the primary experiment and 8.57\% (3 of 35, on \texttt{example\_\allowbreak{}01}) in the ablation experiment. On opaque stimuli the split varied widely: a base example's opaque articulation rate, pooled across configurations, ran from 43.86\% (50 of 114, \texttt{example\_\allowbreak{}01}) to 93.62\% (44 of 47, \texttt{example\_\allowbreak{}06}) in the primary experiment and from 53.16\% (42 of 79, \texttt{example\_\allowbreak{}01}) to 100\% (41 of 41 on \texttt{example\_\allowbreak{}06} and 7 of 7 on \texttt{example\_\allowbreak{}03}) in the ablation experiment; the failure mode dominance heatmaps break these splits out cell by cell (Supplementary~Figures~S10a--c). Those per-base-example rates pool configurations of very different strength: GPT-5.2 with reasoning effort set to none, GPT-5-mini at minimal reasoning effort, and Claude Haiku 4.5 with reasoning off together accounted for 1,144 of the primary experiment's 2,383 such trials and 1,011 of the ablation experiment's 2,024, roughly half in each (Supplementary~Figures~S9a--b).

\section{4. Discussion}\label{discussion}

When the meta-evaluator wrongly rejected the grader's correct claim on incorrect-draft stimuli, it typically articulated the operational interpretation and then set it aside, basing its verdict on field label matching instead. This is the articulate-then-dismiss pattern. The meta-evaluator expressed the operational interpretation on every transparent trial whose verdict did not match ground truth and, in the primary experiment, on 60.78\% of such opaque trials (544 of 895), yet let it govern on almost none of them. On correct-draft stimuli, where the meta-evaluator wrongly accepted the grader's erroneous claim, field label matching governed the verdict and the operational interpretation usually did not surface at all.

\subsection{4.1 Field Label Matching Behaviors May Be an Artifact of Pre-/Post-Training}\label{field-label-matching-behaviors-may-be-an-artifact-of-pre-post-training}

This reliance on field label matching may originate in training. Wei et al.~{[}29{]} define an emergent ability as one absent in smaller models but present in larger ones, where model size is measured by training compute and parameter count. Some capabilities that emerge with increasing model size are plausibly instilled during pre-training, the phase in which a language model learns next-token prediction over a large text corpus. Field label salience and matching could be one such capability, because pre-training corpora draw heavily on text from the web, where structured documents such as forms, tickets, dashboards, and API schemas are common. Those documents are dense with label-value patterns of the form ``Previous {[}x{]}: Y'' or ``Current {[}x{]}: Z.'' Pre-training alone, however, is unlikely to explain why field label matching, rather than the operational interpretation, so often determined the meta-evaluator's verdicts in this experiment.

Post-training may also explain some of the observed field label matching behaviors. Chu et al.~{[}30{]} compared two post-training methods on an out-of-distribution, language-only variant of the V-IRL (Virtual Intelligence in Real Life) navigation benchmark, where the action space shifts from absolute to relative directions. Supervised fine-tuning, which imitates reference outputs, cut per-step accuracy from the starting checkpoint's 80.8\% to 1.3\%. Reinforcement learning, which maximizes a reward signal over the model's own generations, raised it from that same starting point to 91.8\%. Lambert {[}31{]} traces the gap to the objectives' distributional shape. On that account, supervised fine-tuning minimizes the forward KL divergence, which rewards matching every major mode of the training distribution and amounts to memorizing the training data rather than an underlying rule. Reinforcement learning minimizes the reverse KL divergence, concentrating probability on high-reward outputs so that, Lambert argues, the model learns a transferable rule.

However, an association learned during supervised fine-tuning is not necessarily erased by subsequent reinforcement learning. Murray et al.~{[}32{]} describe a related failure pattern, chunky post-training, in which a model conditions a behavior on incidental patterns in its post-training data, including surface features such as formatting, vocabulary, or phrasing, rather than on the intended rule. They traced several such behaviors to identifiable patterns in the supervised fine-tuning data of Tülu 3, a model whose post-training data are publicly available. These chunky behaviors generally survived Tülu 3's later post-training stages, including reinforcement learning, sometimes weakened and sometimes strengthened. An association between an incidental surface feature and a behavior, once learned in supervised fine-tuning, could similarly persist in the six models evaluated in this experiment, all of which are reasoning models trained through reinforcement learning.

Supervised fine-tuning's mode-covering behavior could plausibly produce the articulate-then-dismiss pattern. A model trained to match every mode of a distribution associating ``previous'' with the value in a labeled ``Previous {[}x{]}'' field would reproduce that association even where the surrounding context's temporal standpoint calls for a different resolution. For example, in the primary experiment, Gemini 3 Flash at high reasoning effort demonstrated this behavior in one of the correct-draft stimulus trials: it wrote that ``rolling back to the current version is logically the correct action in a real-world deployment'' and then accepted the grader's mistaken claim, because the draft had ``explicitly mislabeled the `Current Version' value as the `Previous Version', contradicting the metadata provided in the user prompt'' (Supplementary~Exemplar~S2). Field label matching overrode a resolution the meta-evaluator had itself just called correct. Zhang et al.~{[}33{]} report a consistent pattern in a different domain. The two best-performing models on output format compliance in their FireBench benchmark, GPT-4.1 and Qwen3 235B Instruct, each achieved 100\% accuracy when asked to wrap their final answer in the common LaTeX delimiter \texttt{\textbackslash{}boxed\{\}}. With the syntactically similar yet uncommon variant \texttt{\textbackslash{}boxed{[}\ {]}}, their accuracy fell to 53\% and 73\%, respectively. Zhang et al.~read this as evidence that models tend to memorize specific formatting patterns from post-training rather than acquiring a generalizable ability to follow arbitrary formatting instructions. The label-value pattern ``Previous {[}x{]}: Y'' could work the same way. This hypothesis could be tested with TURF (Tracing Unintended Responses via Features) {[}32{]}, a tool Murray et al.~developed for tracing a model's behaviors back to specific patterns in its post-training data. However, the approach requires access to those data, which are not publicly available for the six evaluated models.

\subsection{4.2 The Structure of the Grader's Feedback Can Influence the Meta-Evaluator's Reasoning}\label{the-structure-of-the-graders-feedback-can-influence-the-meta-evaluators-reasoning}

The ablation experiment indicates that the meta-evaluator's reasoning also responds to the structure of the grader's feedback, not just its content. Removing the grader's four error classification fields raised both the opaque articulation rate and the opaque governing rate among trials whose verdict did not match ground truth (Results: Failure Mode Analysis). The effect on accuracy was narrower: only 3 of the 63 condition-level accuracy comparisons between the experiments reached significance, all of them improvements on opaque stimuli (Results: Per-Condition Accuracy). Because the fields were removed together, the effect cannot be traced to any one of them, and two mechanisms remain compatible with it. Three of the fields, \texttt{error\_\allowbreak{}category} (particularly the value \texttt{"hallucinations"}), \texttt{error\_\allowbreak{}subtype}, and \texttt{risk\_\allowbreak{}level}, could distract the meta-evaluator. Such distraction would pull the meta-evaluator's attention away from whether the assistant or the grader was correct and toward whether the labels themselves were warranted. The fourth, \texttt{is\_\allowbreak{}issue\_\allowbreak{}truly\_\allowbreak{}error}, states outright that the flagged issue is a real error, so removing it could make the meta-evaluator less willing to defer to the grader. The two mechanisms predict different results. Less deference should raise correct-draft accuracy, where deferring gives the wrong answer, and lower opaque accuracy, where it gives the right one. Less distraction should raise opaque accuracy alone. Opaque accuracy improved across most configurations while correct-draft accuracy moved inconsistently, with no correct-draft comparison significant in either direction. That pattern points to distraction rather than reduced deference. For example, on one base example's incorrect-opaque stimulus, GPT-5.2 with reasoning effort set to none judged the labels when they were present and the draft when they were removed. In both experiments it granted that reverting could mean returning to the value in effect before the proposed change and rejected the grader's correct claim. With the fields present, its stated reason was that the ambiguity ``is not a factual hallucination or detail misidentification as graded'' (Supplementary~Exemplar~S8). With the fields removed, its stated reason was that choosing the value in the labeled previous field ``is not a factual error relative to the prompt'' (Supplementary~Exemplar~S9). The verdict rested on field label matching either way. Confirming that distraction is the mechanism would take a field-wise ablation.

\subsection{4.3 Pathological Deference to the Grader's Feedback Alone Cannot Account for the Meta-Evaluator's Behaviors}\label{pathological-deference-to-the-graders-feedback-alone-cannot-account-for-the-meta-evaluators-behaviors}

Pathological deference, a persistent tendency to accept the grader's assessment regardless of its content, is a candidate explanation of the meta-evaluator's verdicts in both experiments, rather than of the ablation experiment's effect. The meta-evaluator might treat the grader's feedback as authoritative regardless of its content, whether because it appears last in the meta-evaluator's user prompt or is explicitly labeled as feedback, for example. In this experiment, the grader asserts an error in every stimulus, so a meta-evaluator that always accepted that assertion would return the wrong verdict on every correct-draft trial and the correct verdict on every incorrect-draft trial. Two features of the results contradict deference as the dominant explanation.

First, deference predicts that the meta-evaluator would almost never reject the grader's claim. Yet it wrongly rejected the correct claim on 178 transparent trials (4.2\%) and 895 opaque trials (21.3\%) in the primary experiment, and on 139 transparent trials (3.3\%) and 681 opaque trials (16.2\%) in the ablation experiment, out of 4,200 trials per condition in each. Nor were these rejections confined to the weakest configurations, since opaque accuracy fell below 1.000 for 17 of 21 configurations in the primary experiment and 14 of 21 in the ablation experiment. They persisted even on transparent stimuli, where the grader's analysis states the operational reasoning explicitly. The meta-evaluator articulated the operational interpretation on every one of those transparent wrong rejections in both experiments, yet field label matching governed the verdict in 176 of the primary experiment's 178 such trials and in 136 of the ablation experiment's 139.

Second, wrong verdicts in both directions rose and fell together across configurations rather than trading off. In the extreme case, GPT-5.2 with reasoning effort set to none accepted the mistaken claim on every correct-draft trial yet rejected the correct claim on 160 of its 200 opaque trials in the primary experiment (opaque accuracy 0.200). The ablation experiment produced the same pattern of wrong verdicts in both directions (correct-draft accuracy 0.080; opaque accuracy 0.280). This configuration's wrong verdicts in both directions tracked whichever claim matched the labeled field, not the grader's assessment (Results: Failure Mode Analysis; Results: Per-Condition Accuracy; Supplementary~Table~S13).

Field label matching therefore remains the simpler explanation, since it explains wrong verdicts in both directions while deference explains only the wrong acceptances. Deference cannot be excluded as a partial contributor for the configurations whose verdicts matched ground truth on every incorrect-draft trial while still sometimes wrongly accepting the grader's erroneous claim. Those configurations accounted for 66 of the primary experiment's 1,310 wrong acceptances (5.0\%) and 142 of the ablation experiment's 1,204 (11.8\%). On those trials the rationale analysis records only whether the operational interpretation or field label matching governed the verdict, so it cannot separate deference from field label matching (Methods: Rationale Analysis).

\subsection{4.4 Practical Implications for Model Selection and Configuration in Draft-Verify-Revise Pipelines}\label{practical-implications-for-model-selection-and-configuration-in-draft-verify-revise-pipelines}

Beyond the question of what drove the wrong verdicts, the scaling results carry practical weight for anyone choosing models and configurations for draft-verify-revise pipelines. The cost-performance tradeoffs are stark: Gemini 3 Pro at low reasoning effort outscored GPT-5.2 at xhigh reasoning effort for roughly 5\% of the cost per trial (Results: Scaling Effects). The optimal reasoning effort level is model-specific, and for several models, balanced accuracy dipped at higher reasoning effort levels rather than continuing to climb. Larger models generally outperformed smaller ones within each provider, but the pattern is not universal. GPT-5.2 with reasoning effort set to none scored below every GPT-5-mini configuration. These scaling results rest on 10 base examples: a configuration's 600 trials are 20 repetitions of those examples rendered in three conditions, not 600 independent stimuli. They can narrow which configurations are worth testing in a given deployment; they do not establish how any configuration would score on other stimuli of the same kind.

\subsection{4.5 Limitations}\label{limitations}

The 30 stimuli are synthetic minimal pairs, so how closely they resemble the deictic ambiguities of real-world draft-verify-revise pipelines is unknown. Actual instances may be more varied, more deeply embedded in surrounding context, or less cleanly isolated from confounding cues. The stimuli also test a single kind of ambiguity, the word ``previous'' read either as the value in a labeled field or as the value in effect before the proposed change. Whether these findings extend to other context-dependent expressions, such as ``the response,'' ``the instructions,'' ``current,'' or ``original,'' is untested. SURF (Surfacing Unintended Response Failures) {[}32{]}, an automated auditor tool Murray et al.~developed for discovering unintended model behaviors, could search for prompts in which a model's reading of these context-dependent expressions is governed by surface cues. However, confirming that the expression, rather than anything else in such a prompt, is responsible would require a minimal-pair design like the one used in this experiment.

The stimulus design also limits what a high score establishes on its own. In all 10 base examples the operationally correct reversion target is the value in the assistant's user prompt field labeled ``Current {[}x{]}'' (Methods: Stimulus Design; the per-example values are in Supplementary~Table~S14). A meta-evaluator that never reasoned about temporal standpoint at all, and instead checked only whether the value the draft designates as its reversion target matched the value in that field, would return the correct verdict on all 30 stimuli, for a balanced accuracy of 1.000. That strategy is field label matching (Methods: Rationale Analysis) applied to the field labeled ``Current {[}x{]}'' rather than the field labeled ``Previous {[}x{]}'': its verdict is fixed by which labeled field it consults, and on these stimuli consulting the field labeled ``Previous {[}x{]}'' always produces wrong verdicts while consulting the field labeled ``Current {[}x{]}'' always produces correct ones. The rationale analysis addresses this confound where it can, but it classifies only trials whose verdict did not match ground truth, so on the trials that matched, what produced the verdict is not observable. That blind spot widens as per-condition accuracy rises: sensitivity on transparent stimuli reached 1.000 for 14 of the primary experiment's 21 configurations and for 18 of the ablation experiment's 21 (Supplementary~Table~S1), leaving the probe no trials to classify in that condition. Cleanly separating operational reasoning from field label matching would require stimuli in which the operationally correct reversion target and the value in the field labeled ``Current {[}x{]}'' diverge.

What counts as a correct verdict is also less sharply defined than a binary verdict suggests, because the grader's assessment has two parts that can come apart: its conclusion that the assistant's draft is in error, and the reasoning it offers for that conclusion. In 9 of the 10 base examples, the incorrect-opaque grader reaches the right conclusion through an analysis whose opening two sentences are in tension, the second asserting in plain prose the reading the first denies; \texttt{example\_\allowbreak{}03} is the exception, its second sentence citing the capitalized field labels instead. The \texttt{example\_\allowbreak{}04} analysis begins: ``The assistant incorrectly identifies 15mg as the `previous dose' in the safety protocol. According to the user prompt, the previous dose is 15mg and the current dose is 20mg.'' The first sentence rejects 15mg as the value to revert to; the second cites the user prompt as establishing that the previous dose is 15mg. A meta-evaluator that judges the reasoning rather than the conclusion therefore has grounds to call the assessment incorrect, and the design records the resulting verdict as one that did not match ground truth. For example, Claude Haiku 4.5 with reasoning off judged the reasoning rather than the conclusion on this stimulus: it resolved ``previous'' correctly, stated that the fallback should have been 20mg, and called the assessment incorrect because the grader, having conflated the earlier dose in the patient history with the dose being adjusted from, had reached the right conclusion for partially incorrect reasoning (Supplementary~Exemplar~S6). In the rationale analysis, the grader-claim flag fired on 20.8\% of the primary experiment's wrong verdicts on opaque stimuli (186 of 895) and on 24.8\% of the ablation experiment's (169 of 681), against 0.1\% (1 of 1,310) and 0.9\% (11 of 1,204) on correct-draft stimuli, where the grader's analysis carries no such tension (Supplementary~Table~S12). Whether those rationales are read as misreports of the grader's claim or as disagreements with it (the reading the probe audit more often reached), they concentrate on the condition in which the grader's own analysis carries that tension. Some share of the wrong verdicts on opaque stimuli may therefore register an objection to how the grader argued rather than a failure to determine which reading of ``previous'' was correct.

No model received tailored prompts, so real-world deployments could improve performance through context engineering. Even so, the ablation experiment's model- and condition-specific effects suggest such interventions require careful per-model validation. Relatedly, error category definitions were not included in the meta-evaluator's prompt, so it relied on its own understanding of categories such as \texttt{hallucinations} and \texttt{detail\_\allowbreak{}misidentification}. That reliance may have made the error classification labels stronger distractors than explicit definitions would have.

The meta-evaluator was only tasked with a binary verdict and never had to produce a revised draft. Results might differ if it also had to produce one.

The probe's two discriminating flags are forced binaries. Its response schema requires a true or false classification for each flag, with no indeterminate category, and for the governing flag the false branch is itself a positive claim: that the field label interpretation, rather than the operational interpretation, determined the verdict. The probe therefore assigns every trial it classifies to one of those two readings, including trials whose rationale text supports neither cleanly. The claims drawn from the probe accordingly stay at the level of what the stated rationale says: for example, that a rationale expresses the operational interpretation and then endorses a verdict consistent with field label matching. They are not claims about which computation produced the verdict, a question the stated rationale cannot settle (Methods: Rationale Analysis).

The recorded verdict and the stated rationale can also contradict each other. The design scores the verdict, and the probe classifies the rationale. For example, on a correct-draft stimulus, GPT-5-mini at low reasoning effort concluded that the grader's identification of an error was mistaken, the conclusion that matches ground truth, yet returned the verdict that the grader's assessment was correct. The trial counted as a verdict that did not match ground truth, and the probe recorded the operational interpretation as governing (Supplementary~Exemplar~S10). The contradiction also runs the other way: on an incorrect-opaque stimulus, GPT-5-mini at minimal reasoning effort rejected the grader's correct claim in its rationale, on the strength of the labeled field alone, yet returned the verdict that the grader's assessment was correct. That verdict matched ground truth, so the probe never classified the rationale (Supplementary~Exemplar~S11). How often verdict and rationale diverged was not measured.

A Claude Code agent {[}34{]} using Claude Opus 4.8 {[}35{]} re-classified a subset of the rationale analysis probe's classifications (Supplementary~Methods~SM5). That audit establishes only that the probe applied its criteria as written, not that its classifications are correct. Nor is it a fully independent check: Claude Opus 4.8 is an Anthropic model, as are two of the six models evaluated in this experiment, Claude Haiku 4.5 and Claude Opus 4.6†. Because models from the same developer can behave in correlated ways, a model from a different developer, or a human, would likely provide a stronger check. Calibrating the probe is future work, requiring a panel of independent human annotators to evaluate it on trials whose verdict matched ground truth as well as on those whose verdict did not. Calibration is distinct from reliability, that is, whether a second draw from the same probe would return the same classification; each rationale was classified once, so the probe's reliability is likewise unmeasured.

\section{5. Conclusion}\label{conclusion}

As context cascades from one stage of an LLM pipeline to the next, a context-dependent expression such as ``previous'' can acquire a different referent. This experiment isolated a single deictic ambiguity of this kind and tested whether an LLM can resolve the perspectival disagreement that arises when the LLMs at different pipeline stages read the same context-dependent expression from different temporal standpoints. The results bring into question whether the LLM at the revise stage of a draft-verify-revise pipeline can be relied on to do so correctly. When the LLM under test erred, it tended to rely on surface cues, such as field labels, rather than operational reasoning. This evidence comes from 10 base examples rendered in three conditions. It establishes that this failure occurs and what it looks like, not how often it would arise in the pipelines where such an ambiguity can appear.

This vulnerability matters for any system that uses the draft-verify-revise pattern and reasons over sequential states to determine the next course of action, such as decision support systems in domains where historical context shapes the appropriate next step. Depending on model size and the inference-time compute afforded to each stage, the LLMs at the verify and revise stages could share the same tendency to rely on surface cues rather than operational reasoning. If they do, the pattern may not deliver the intended safeguard.

A larger model with insufficient inference-time compute can score below a smaller model, and the cost-performance tradeoffs can be dramatic: Gemini 3 Pro at low reasoning effort outscored GPT-5.2 at xhigh reasoning effort for roughly 5\% of the cost per trial. Even the structure of the verifier's feedback can influence the reasoning of the LLM at the revise stage, which could in turn affect the quality of its revisions. Whatever model and reasoning effort each stage runs, context engineers should make each context-dependent expression's intended referent explicit in the prompts and context supplied to the LLM at each stage, rather than leave it to be inferred.

\section{Data and Code Availability}\label{data-and-code-availability}

The dataset and source code supporting the findings of this study are openly available at \url{https://github.com/oekekezie/deictic-ambiguity-companion} under the MIT License. The repository contains the synthetic dataset and its ablation variant, the complete per-trial outcomes of the primary experiment, the ablation experiment, and the rationale analysis, and the analysis code that produced the figures and tables in this paper. The exact version used for these results is archived on Zenodo (DOI: 10.5281/zenodo.22696649). The repository does not include the raw provider batch outputs, which are available from the corresponding author upon reasonable request. I welcome inquiries regarding collaboration, extensions of this work, or discussion of the underlying methodology.

\section{Funding}\label{funding}

This work received no external, institutional, or commercial funding.

\section{Competing Interests}\label{competing-interests}

No competing interests to declare.

\section{Use of Generative AI}\label{use-of-generative-ai}

During this work, I used Anthropic's Claude Code and Claude Cowork throughout the research process to help refine the original idea, design the experiment, incorporate and implement e-value hypothesis testing, develop the software for batch inference and analysis, generate figures and tables, and draft and edit the manuscript. I originated the research question, directed the work, made the final methodological, analytical, and editorial decisions, and am solely responsible for the study and its contents.

\begin{center}\rule{0.5\linewidth}{0.5pt}\end{center}

\section{References}\label{references}

{[}1{]} C. Snell, J. Lee, K. Xu, and A. Kumar, ``Scaling LLM Test-Time Compute Optimally can be More Effective than Scaling Model Parameters,'' arXiv preprint arXiv:2408.03314, Aug.~2024.

{[}2{]} J. Wei \emph{et al.}, ``Chain-of-Thought Prompting Elicits Reasoning in Large Language Models,'' in \emph{Proc. NeurIPS}, 2022, arXiv preprint arXiv:2201.11903.

{[}3{]} OpenAI, ``OpenAI o1 System Card,'' arXiv preprint arXiv:2412.16720, Dec.~2024.

{[}4{]} DeepSeek-AI \emph{et al.}, ``DeepSeek-R1: Incentivizing Reasoning Capability in LLMs via Reinforcement Learning,'' arXiv preprint arXiv:2501.12948, Jan.~2025.

{[}5{]} A. Panickssery, S. R. Bowman, and S. Feng, ``LLM Evaluators Recognize and Favor Their Own Generations,'' arXiv preprint arXiv:2404.13076, Apr.~2024.

{[}6{]} W. Xu, G. Zhu, X. Zhao, L. Pan, L. Li, and W. Y. Wang, ``Pride and Prejudice: LLM Amplifies Self-Bias in Self-Refinement,'' arXiv preprint arXiv:2402.11436, Jun.~2024.

{[}7{]} R. Kamoi, Y. Zhang, N. Zhang, J. Han, and R. Zhang, ``When Can LLMs Actually Correct Their Own Mistakes? A Critical Survey of Self-Correction of LLMs,'' arXiv preprint arXiv:2406.01297, Dec.~2024.

{[}8{]} Y. Kim \emph{et al.}, ``Towards a Science of Scaling Agent Systems,'' arXiv preprint arXiv:2512.08296, Dec.~2025.

{[}9{]} E. Y. Chang and L. Geng, ``SagaLLM: Context Management, Validation, and Transaction Guarantees for Multi-Agent LLM Planning,'' arXiv preprint arXiv:2503.11951, Mar.~2025.

{[}10{]} R. Xu, T. Liu, Z. Dong, T. Yu, I. Hong, C. Yang, L. Zhang, T. Zhao, and H. Wang, ``Alternating Reinforcement Learning for Rubric-Based Reward Modeling in Non-Verifiable LLM Post-Training,'' arXiv preprint arXiv:2602.01511, Feb.~2026.

{[}11{]} W. F. Shen, X. Qiu, C. Whitehouse, L. Alazraki, S. Goel, F. Barbieri, T. Willi, A. Mathur, and I. Leontiadis, ``Rethinking Rubric Generation for Improving LLM Judge and Reward Modeling for Open-ended Tasks,'' arXiv preprint arXiv:2602.05125, Feb.~2026.

{[}12{]} K. Hong, A. Troynikov, and J. Huber, ``Context Rot: How Increasing Input Tokens Impacts LLM Performance,'' Chroma, Tech. Rep., Jul.~2025. {[}Online{]}. Available: \url{https://research.trychroma.com/context-rot}

{[}13{]} D. Zhang and J. Rayz, ``Structured yet Bounded Temporal Understanding in Large Language Models,'' arXiv preprint arXiv:2510.16685, Jan.~2026.

{[}14{]} M. Raemaekers, M. Finn, and J. De Houwer, ``Assessing the Relational Abilities of Large Language Models and Large Reasoning Models,'' \emph{Behav. Sci.}, vol.~16, no. 1, art. 45, Dec.~2025, doi: 10.3390/bs16010045.

{[}15{]} P. Farag and S. Polevikov, ``The Cough That Broke 12 LLMs,'' \emph{AI Health Uncut}, Mar.~3, 2026. {[}Online{]}. Available: \url{https://www.fixhealth.ai/p/the-cough-that-broke-12-llms}

{[}16{]} P. Chung \emph{et al.}, ``Verifying Facts in Patient Care Documents Generated by Large Language Models Using Electronic Health Records,'' \emph{NEJM AI}, Dec.~2025, doi: 10.1056/AIdbp2500418.

{[}17{]} P. Brodeur \emph{et al.}, ``A prospective clinical feasibility study of a conversational diagnostic AI in an ambulatory primary care clinic,'' arXiv preprint arXiv:2603.08448, Mar.~2026.

{[}18{]} OpenAI, ``GPT-5 System Card,'' Aug.~2025. {[}Online{]}. Available: \url{https://cdn.openai.com/gpt-5-system-card.pdf}

{[}19{]} OpenAI, ``Update to GPT-5 System Card: GPT-5.2,'' Dec.~2025. {[}Online{]}. Available: \url{https://cdn.openai.com/pdf/3a4153c8-c748-4b71-8e31-aecbde944f8d/oai_5_2_system-card.pdf}

{[}20{]} Anthropic, ``System Card: Claude Haiku 4.5,'' Oct.~2025. {[}Online{]}. Available: \url{https://assets.anthropic.com/m/99128ddd009bdcb/Claude-Haiku-4-5-System-Card.pdf}

{[}21{]} Anthropic, ``System Card: Claude Opus 4.6,'' Feb.~2026. {[}Online{]}. Available: \url{https://www-cdn.anthropic.com/c788cbc0a3da9135112f97cdf6dcd06f2c16cee2.pdf}

{[}22{]} Google DeepMind, ``Gemini 3 Flash Model Card,'' Dec.~2025. {[}Online{]}. Available: \url{https://storage.googleapis.com/deepmind-media/Model-Cards/Gemini-3-Flash-Model-Card.pdf}

{[}23{]} Google DeepMind, ``Gemini 3 Pro Model Card,'' Dec.~2025. {[}Online{]}. Available: \url{https://storage.googleapis.com/deepmind-media/Model-Cards/Gemini-3-Pro-Model-Card.pdf}

{[}24{]} A. Aali \emph{et al.}, ``MedVAL: Toward Expert-Level Medical Text Validation with Language Models,'' arXiv preprint arXiv:2507.03152, Feb.~2026.

{[}25{]} A. Ramdas and R. Wang, \emph{Hypothesis Testing with E-values}. Carnegie Mellon University and University of Waterloo, Sep.~2025. {[}Online{]}. Available: \url{https://stat.cmu.edu/~aramdas/ebook-final.pdf}

{[}26{]} I. Arcuschin, J. Janiak, R. Krzyzanowski, S. Rajamanoharan, N. Nanda, and A. Conmy, ``Chain-of-Thought Reasoning In The Wild Is Not Always Faithful,'' arXiv preprint arXiv:2503.08679, Jun.~2025.

{[}27{]} Y. Chen \emph{et al.}, ``Reasoning Models Don't Always Say What They Think,'' Anthropic, 2025. {[}Online{]}. Available: \url{https://assets.anthropic.com/m/71876fabef0f0ed4/original/reasoning/_models/_paper.pdf}

{[}28{]} Z. Xiong, S. Chen, Z. Qi, and H. Lakkaraju, ``Measuring the Faithfulness of Thinking Drafts in Large Reasoning Models,'' arXiv preprint arXiv:2505.13774, May 2025.

{[}29{]} J. Wei, Y. Tay, R. Bommasani, C. Raffel, B. Zoph, S. Borgeaud, D. Yogatama, M. Bosma, D. Zhou, D. Metzler, E. H. Chi, T. Hashimoto, O. Vinyals, P. Liang, J. Dean, and W. Fedus, ``Emergent Abilities of Large Language Models,'' \emph{Trans. Mach. Learn. Res.}, 2022.

{[}30{]} T. Chu \emph{et al.}, ``SFT Memorizes, RL Generalizes: A Comparative Study of Foundation Model Post-training,'' in \emph{Proc. ICML}, 2025.

{[}31{]} N. Lambert, \emph{Reinforcement Learning from Human Feedback: A Short Introduction to RLHF and Post-Training Focused on Language Models}. Self-published, Apr.~2026. {[}Online{]}. Available: \url{https://rlhfbook.com}

{[}32{]} S. Murray, A. Qi, T. Qian, J. Schulman, C. Burns, and S. Price, ``Chunky Post-Training: Data Driven Failures of Generalization,'' arXiv preprint arXiv:2602.05910, Feb.~2026.

{[}33{]} Y. Zhang, Y. Bei, J. Ravi, and P. Garbacki, ``FireBench: Evaluating Instruction Following in Enterprise and API-Driven LLM Applications,'' arXiv preprint arXiv:2603.04857, Mar.~2026.

{[}34{]} Anthropic, ``Claude Code Overview,'' 2026. {[}Online{]}. Available: \url{https://code.claude.com/docs/en/overview}

{[}35{]} Anthropic, ``System Card: Claude Opus 4.8,'' May 2026. {[}Online{]}. Available: \url{https://www-cdn.anthropic.com/0b4915911bb0d19eca5b5ee635c80fef830a37ea.pdf}

\onecolumn
\phantomsection
\addcontentsline{toc}{section}{Supplementary Materials}
\begin{center}
{\LARGE\bfseries Supplementary Materials\par}
\end{center}
\vspace{1.5\baselineskip}

\section{Supplementary Methods}\label{supplementary-methods}

\subsection{SM1. Stimulus Specification}\label{sm1.-stimulus-specification}

This section reproduces the meta-evaluator's system prompt and response schema, points to the base example inventory, walks through a worked example across the three conditions, and lists which stimulus components are held constant across conditions and which vary.

\textbf{The meta-evaluator's system prompt.} Every trial sends the same system prompt to the meta-evaluator:

\begin{verbatim}
Return your output as a JSON object in the following format:

```
{
  "rationale": "Rationale for whether the grader's feedback was correct or incorrect.",
  "score": "correct" | "incorrect"
}
```

- The `rationale` property must explain whether the grader's feedback was correct or incorrect.
- The `score` property must be either "correct" (the grader was right) or "incorrect" (the grader was wrong).
- Both properties are required. No additional properties are allowed.
\end{verbatim}

\textbf{The response schema.} Each trial also provides the following JSON schema to the provider's structured output mechanism:

\begin{verbatim}
{
  "type": "object",
  "properties": {
    "rationale": {
      "type": "string",
      "description": "Rationale for whether the grader's feedback was correct or incorrect."
    },
    "score": {
      "type": "string",
      "description": "Was the grader's feedback correct? 'correct' means the grader was right; 'incorrect' means the grader was wrong.",
      "enum": ["correct", "incorrect"]
    }
  },
  "required": ["rationale", "score"],
  "additionalProperties": false
}
\end{verbatim}

The schema is applied through each provider's structured output mechanism: OpenAI \texttt{text.\allowbreak{}format} with JSON Schema strict mode, Anthropic \texttt{output\_\allowbreak{}config.\allowbreak{}format} with \texttt{type:\ "json\_\allowbreak{}schema"}, and Google \texttt{generationConfig.\allowbreak{}responseSchema} with Gemini-compatible schema normalization.

\textbf{Base example inventory.} Supplementary~Table~S14 reproduces the full inventory.

\textbf{A worked example across three conditions.} To make the minimal-pair structure concrete, consider \texttt{example\_\allowbreak{}04}. The assistant's system prompt directs a clinical decision support assistant that ``all dosage adjustment recommendations must include a safety protocol specifying the previous dose to revert to if adverse effects occur.'' The assistant's user prompt provides a patient profile with \texttt{Current\ Dose:\ 20mg}, \texttt{Previous\ Dose:\ 15mg}, and a clinical history entry noting that the patient was titrated from \texttt{15mg} to \texttt{20mg} six weeks ago, together with a request for a proposed adjustment. The assistant's draft proposes a titration from \texttt{20mg} to \texttt{25mg} with a safety protocol contingency clause. The three conditions differ as follows in how the assistant's draft and the grader's feedback read ``previous'':

In the \textbf{correct-draft} condition, the draft's contingency is ``reduce to the previous dose of \texttt{20mg}.'' That dose is the value in effect immediately before the assistant's proposed titration, so the draft is operationally correct. The grader flags this as an error, writing: \emph{``The assistant incorrectly identifies \texttt{20mg} as the `previous dose' in the safety protocol. According to the user prompt, the current dose is \texttt{20mg} and the `Previous Dose' field is explicitly \texttt{15mg}. This error is particularly dangerous in a medical context, as it could lead to incorrect dosing instructions if a clinician follows the stated `previous dose' without cross-referencing the patient record.''} Ground truth here is ``incorrect'' (the grader was wrong). The meta-evaluator's task on this condition is to recognize that the grader mistook the labeled field's value for the value in effect before the proposed change.

In the \textbf{incorrect-transparent} condition, the draft's contingency is ``reduce to the previous dose of \texttt{15mg}.'' That is the labeled-field value, so the contingency is operationally wrong. The grader's analysis text lays out the operational reasoning explicitly: \emph{``The assistant incorrectly uses 15mg as the reversion dose. The proposed titration is from 20mg to 25mg, so reverting means returning to the pre-titration dose of 20mg. The assistant appears to have pulled the value from the `Previous Dose' field rather than reasoning about what dose was in effect before the proposed adjustment.''} Ground truth is ``correct'' (the grader was right), and the meta-evaluator can arrive at the right verdict by following the grader's own operational reasoning.

In the \textbf{incorrect-opaque} condition, the draft is identical to the transparent one (the contingency still cites \texttt{15mg}) but the grader's analysis is rewritten so that it is structurally indistinguishable from the correct-draft grader's: \emph{``The assistant incorrectly identifies 15mg as the `previous dose' in the safety protocol. According to the user prompt, the previous dose is 15mg and the current dose is 20mg. This error is particularly dangerous in a medical context, as it could lead to an unnecessarily large dose reduction if a clinician follows the stated `previous dose' without cross-referencing the patient record.''} Ground truth is again ``correct,'' but the grader's surface cues no longer carry the operational reasoning; the meta-evaluator must supply it independently.

The minimal-pair structure the worked example illustrates holds across all 10 base examples. Within a base example, the grader's feedback carries identical \texttt{evidence\_\allowbreak{}against\_\allowbreak{}error} quotes sourced from the assistant's system prompt in all three conditions, and identical \texttt{evidence\_for\_error{[}1{]}} and \texttt{{[}2{]}} quotes sourced from the assistant's user prompt. Within the evidence arrays, only \texttt{evidence\_for\_error{[}0{]}} (the citation sourced from the assistant's response) differs, and it differs between the correct and incorrect variants. The grader's \texttt{analysis} and \texttt{root\_\allowbreak{}cause} texts differ across all three conditions; the three quotations above are the three \texttt{analysis} texts. For each example, the error category (\texttt{hallucinations}), error subtype (\texttt{detail\_\allowbreak{}misidentification}), risk level, and \texttt{is\_\allowbreak{}issue\_\allowbreak{}truly\_\allowbreak{}error} are constant across all three conditions in the primary experiment. The ablation dataset keeps the \texttt{analysis} and \texttt{root\_\allowbreak{}cause} texts and the evidence arrays verbatim but removes the four error classification fields.

\subsection{SM2. Staged Design}\label{sm2.-staged-design}

The 21-configuration primary experiment and the matched 21-configuration ablation experiment were assembled from five stages, each with a purpose and decision rules defined in advance of data collection:

Stage 1 ran each model at two reasoning effort levels, its lowest and a selected upper bound, to test whether the deictic ambiguity could be resolved at all and to map the accuracy range those two levels span. Those reasoning effort levels were as follows: none and high for GPT-5.2, minimal and high for GPT-5-mini, low and max for Claude Opus 4.6†, off and enabled for Claude Haiku 4.5, minimal and high for Gemini 3 Flash, and low and high for Gemini 3 Pro. For five of the six models the upper bound was the highest level available for that model; for GPT-5.2 it was the high rather than the xhigh level, which Stage 4 added later. Stage 1 contained 12 configurations.

Stage 2 filled in the intermediate reasoning effort levels for models with three or more graduated levels. It was conditional on two Stage 1 decision rules, both evaluated over independent models only (models with no role in constructing the stimuli): that at least one independent model performed significantly above chance at its upper bound, with its per-cell e-values rejecting the chance null across conditions, and that at least one independent model with three or more graduated levels showed a meaningfully different performance profile at its two bounds. Stage 2 added GPT-5.2 at low and medium reasoning effort, GPT-5-mini at low and medium reasoning effort, Claude Opus 4.6† at medium and high reasoning effort, and Gemini 3 Flash at low and medium reasoning effort (eight configurations).

Stage 3 was the error classification ablation, which re-ran the 20 configurations from Stages 1 and 2 on the ablation dataset. It was not gated on Stages 1 or 2 findings. Instead, it was a stimulus manipulation defined in advance of data collection, and it used the primary experiment's results as its comparison baseline.

Stage 4 added GPT-5.2 at xhigh reasoning effort on the primary dataset, a level deferred from Stage 1 for cost management. Stage 5 added GPT-5.2 at xhigh reasoning effort on the ablation dataset, matching Stage 4 so that the two sides of the paired comparison between the primary experiment and the ablation experiment remained aligned one-for-one. Stages 1, 2, and 4 yielded the primary experiment's 21 configurations; Stages 3 and 5 yielded the ablation experiment's 21 configurations.

\subsection{SM3. Batch Inference Infrastructure}\label{sm3.-batch-inference-infrastructure}

Both experiments were run through a custom multi-provider batch submission pipeline spanning the OpenAI, Anthropic, and Google batch APIs. Each stimulus JSONL file (a list of provider-agnostic examples) was expanded into \(N = 20\) trials per example per configuration and submitted through the relevant provider's batch API. A single provider-agnostic example schema (system prompt, user prompt, constrained output schema) was paired with a per-provider generation configuration and serialized to each provider's native batch format. Each configuration's reasoning effort setting was translated to that model's native parameter: OpenAI's \texttt{reasoning\_\allowbreak{}effort}, Anthropic's \texttt{effort} field for Claude Opus 4.6† and its extended-thinking token budget for Claude Haiku 4.5 (whose off configuration omits the extended-thinking block), and Google's \texttt{thinking\_\allowbreak{}level}. Individual failed requests were retried through a linking mechanism that deduplicates on the \emph{(example identifier, trial index)} key, so replays did not inflate trial counts.

Beyond the reasoning effort level, each configuration fixed the following generation settings. The OpenAI configurations set a 128,000-token output limit, reasoning summaries to auto, and verbosity to high. The Anthropic configurations set a 128,000-token limit for Claude Opus 4.6† with adaptive thinking and a 64,000-token limit for Claude Haiku 4.5, whose enabled configuration also set temperature to 1.0 and an extended-thinking budget of 32,768 tokens and whose off configuration set neither. The Google configurations set a 65,536-token output limit and requested thought inclusion. No configuration except Claude Haiku 4.5 with reasoning enabled set a temperature or a top-p value, so each provider's default sampling parameters applied, and all within-cell variation across a cell's 20 trials arises from that sampling. Because provider defaults are not published as stable values and can change between model versions, the trial-level variation reported here should be understood as variation under the defaults in effect during the February 2026 collection window rather than under a decoding configuration this paper fixes.

Reasoning token counts have two provenances. OpenAI and Google report them directly, in \texttt{output\_\allowbreak{}tokens\_\allowbreak{}details.\allowbreak{}reasoning\_\allowbreak{}tokens} and \texttt{thoughtsTokenCount} respectively. Anthropic reports a single output token total covering both extended thinking and the visible response, so for Claude Opus 4.6† and Claude Haiku 4.5 the reasoning token count is derived: each response's visible text is measured with Anthropic's token-counting endpoint, a per-model framing baseline is subtracted, and the result is subtracted from the reported output total. The Reasoning Tokens and Response Tokens columns of Supplementary~Table~S7 and the Anthropic series in Supplementary~Figures~S6 and S7 rest on that derivation. Claude Haiku 4.5 with reasoning off omits the extended-thinking block, so its reasoning total is zero, shown as an em-dash, and its response total equals its output total.

Costs were computed from the provider list prices in effect during the February 2026 collection window, in U.S. dollars per million tokens: GPT-5.2 at \$1.75 input and \$14.00 output, GPT-5-mini at \$0.25 and \$2.00, Claude Opus 4.6† at \$5.00 and \$25.00, Claude Haiku 4.5 at \$1.00 and \$5.00, Gemini 3 Flash at \$0.50 and \$3.00, and Gemini 3 Pro at \$2.00 and \$12.00, with cached input billed at one tenth of the input rate and Anthropic cache writes at 1.25 times the input rate. Claude Opus 4.6† and Gemini 3 Pro carry a higher price tier above a long-context threshold that no request in these experiments reached, so the sub-threshold rates above are the ones that applied. Because every trial was submitted through a provider batch API, each component was then halved to reflect the 50\% batch discount all three providers offer, so every cost reported here is a batch price and an interactive deployment of the same pipeline would pay approximately twice these amounts. Reasoning tokens are billed at the output rate throughout. Provider prices can change after collection, so the cost comparisons should be read as a snapshot of the prices in effect during that window rather than as stable ratios.

\subsection{SM4. E-Value Testing Details}\label{sm4.-e-value-testing-details}

This section specifies the e-value tests summarized in Methods: Statistical Approach. One definition is common to every test below: a \emph{valid trial} is a trial whose response parsed and produced a verdict, and trials that failed to parse contribute to neither numerator nor denominator. One reporting convention is likewise common to the corrected families below: every column headed \emph{Adjusted E-value} reports a raw statistic divided by its family's correction factor, \(K\) for the degradation and model size families and \(2K\) for the cross-dataset families, and a comparison is significant when that adjusted value reaches the rejection threshold of 20. The per-cell Bernoulli likelihood ratio tests share three further recurring elements. The \emph{fixed reference} \(p_0\) is the success rate the null hypothesis is built from. These tests set it to the chance rate of 0.5. The \emph{plug-in alternative} is the alternative hypothesis's success rate, re-estimated before each batch (one iteration's block of trials, 20 per cell) from the data of prior batches only, so that each batch is scored against an alternative fixed before its data are seen. The \emph{seed alternative} is fixed in advance at 0.7 and plays this role for the first batch, which has no prior batches to estimate from. In these tests, alternatives are clamped at least \(\varepsilon = 0.01\) away from the reference and from 1 (\(\varepsilon\)-clamping), for the reasons given below.

\textbf{The Bernoulli likelihood ratio e-value.} The per-cell e-value is the likelihood ratio of a Bernoulli alternative against a Bernoulli null. For a cell containing \(k\) matching-verdict trials in \(n\) valid trials with null proportion \(p_0\) and alternative proportion \(p_1\), the log e-value is

\[\log E \;=\; k \, \log\!\left(\tfrac{p_1}{p_0}\right) + (n - k) \, \log\!\left(\tfrac{1 - p_1}{1 - p_0}\right).\]

Log space computation keeps the statistic numerically stable when evidence is strong (which would otherwise overflow standard floating-point representation) or when evidence favors the null over many batches (which would otherwise underflow to zero).

\textbf{Seed and adaptive plug-in alternative with \(\varepsilon\)-clamping.} Per-cell tests against the chance null use \(p_0 = 0.5\). Each cell's first batch is scored against the seed alternative \(p_1 = 0.7\). Each subsequent batch is scored against the observed hit rate \(\hat{p}\) of the cell's accumulated prior batches, a plug-in estimate of \(p_1\) clamped away from the null by \(\varepsilon = 0.01\). These tests are one-sided above chance, so the alternative is restricted to \([p_0 + \varepsilon, 1 - \varepsilon]\). The clamp serves two purposes. It keeps the alternative strictly inside the hypothesis space for which the e-process is valid, preserving Ville's inequality across the composite null \(H_0: p \le p_0\), and it prevents the log-likelihood from diverging when the observed rate is at a boundary. Because both experiments ran a single iteration, every cell accumulated exactly one 20-trial batch, so every per-cell e-value reported here was scored against the seed alternative \(p_1 = 0.7\).

\textbf{Cell resolution, rejection, and futility.} The rejection threshold of 20 corresponds to one-sided type-I error at \(\alpha = 0.05\) via \(1/E\). The futility threshold of \(0.05\) marks as futile any cell whose accumulated evidence is at least 20 times more consistent with chance than with the alternative. Cells that reach neither boundary remain \emph{active} and can be extended in a subsequent iteration without inflating false positives, because the e-process is valid at arbitrary stopping times.

\textbf{Worked example.} Two hypothetical cells ground the per-cell thresholds. A cell that reaches 16 hits in 20 trials against \(p_0 = 0.5\) with the seed alternative \(p_1 = 0.7\) produces \(\log E = 16 \log(0.7/0.5) + 4 \log(0.3/0.5) = 16 \times 0.3365 + 4 \times (-0.5108) = 5.38 - 2.04 = 3.34\), so \(E \approx 28.2\), above the rejection threshold of 20. A cell that reaches 14 hits in 20 trials produces \(\log E = 14 \times 0.3365 + 6 \times (-0.5108) = 4.71 - 3.06 = 1.65\), so \(E \approx 5.2\), below the rejection threshold. That cell remains active.

\textbf{Confidence sequences.} For accuracy estimation, time-uniform confidence bounds are constructed by inverting a beta-binomial mixture test martingale. A value leaves the interval exactly when its anytime test crosses \(1/\alpha\), so the sequence functions as both an estimator and a test {[}25{]}. The mixture uses the weakly informative Jeffreys prior, \(\text{Beta}(0.5, 0.5)\). This objective default commits to no particular accuracy and lets the observed data dominate the interval. Bounds at level \(1 - \alpha = 0.95\) are located by bisecting over the candidate accuracy until its martingale value crosses \(1/\alpha\). Two-sided confidence sequences are used for accuracy estimation in both experiments. The probe-auditor alignment analysis (Supplementary~Methods~SM5) reports a one-sided lower bound instead, since the alignment claim is one-directional.

\textbf{Testing-by-betting e-values for condition comparisons.} Pairwise condition comparisons (for example, transparent versus opaque within a single configuration) use a sequential testing-by-betting e-value {[}25{]}. The test aggregates per-example binary differences into trial-level betting rounds: one round for each trial of each batch, carrying up to 10 paired differences, one per base example. The pairwise null is equality of the two condition accuracies, \(H_0: p_A = p_B\). Under that null the round's paired differences have expected mean zero, so a bet of either sign has nothing to chase. Each round's bet is a predictable plug-in, namely the clipped running mean-to-variance ratio of the differences observed in prior rounds, and is fixed before the round's differences are seen. Under the null the wealth process is then a nonnegative supermartingale, valid at any stopping time by Ville's inequality. A round multiplies the accumulated wealth by \(\prod (1 + \lambda d)\), the product over the paired differences it carries. The clip, or betting cap, bounds that bet at 0.5 in magnitude, an interior default fixed a priori rather than tuned to the results: it is \(\gamma = 1/2\) in Definition 7.21 of {[}25{]}, under the exact correspondence between mixing weights on e-values and bets on bounded differences. The magnitude matters because a difference is \(-1\), \(0\), or \(+1\), which puts \(|\lambda| = 1\) exactly on the ruin boundary: one adverse pair bet at full magnitude multiplies the wealth by zero, destroying all evidence accumulated up to that round no matter how strong. Holding \(\lambda\) strictly inside the boundary keeps each pair's wealth factor \(1 + \lambda d\) within \([1/2, 3/2]\), so an adverse pair costs a halving rather than everything. This construction produces all three of each configuration's pairwise e-values (Supplementary~Table~S3), including the transparent-versus-opaque comparison that quantifies the gap reported in Results: Per-Condition Accuracy. Supplementary~Table~S15 reports, for each cap in a fixed grid, the number of significant pairwise condition comparisons in each experiment and the number whose significance status changes relative to the reported cap.

\textbf{Multiplicity control for the condition comparisons.} Each configuration contributes three pairwise condition comparisons, so a significance claim about any one of them must account for the other two. The global null statistic for a configuration is the equal-weight mean of its three pairwise e-values. Averaging e-values gives a valid e-value for the intersection of their nulls under arbitrary dependence {[}25{]}, which is what this construction needs, because the three pairwise processes are computed over the same trial rounds and are strongly dependent. The statistic therefore tests the global null \(H_0: p_{\text{correct}} = p_{\text{transparent}} = p_{\text{opaque}}\) within that configuration. The weights are fixed at equal by design, since a weight chosen after seeing the data would break the expectation bound that makes the mean a valid e-value (its expected value under the null is at most 1).

\textbf{The double threshold as an exact closed test.} A pairwise comparison is significant when its own e-value and its configuration's global null e-value both reach 20. That double threshold is an exact closed test rather than an approximation of one. Closed testing rejects an individual hypothesis only when every intersection hypothesis containing it is also rejected. Because all three nulls are equalities, any two of them imply the third; each pairwise intersection is therefore the global null itself. The closure collapses from the seven non-empty intersections of the three nulls to four distinct hypotheses, namely the three individual comparisons and the global null. Transitivity likewise restricts the set of true nulls to the empty set, a single comparison, or all three, so any family-wise error requires a single valid e-process to cross the threshold: the global null statistic when all three nulls are true, or that comparison's own e-value when just one is. Ville's inequality bounds either crossing at \(\alpha\) at every stopping time; no union bound is needed. Because an arithmetic mean never exceeds its maximum, the second threshold can only ever block a comparison that clears 20 on its own; it never carries one that does not.

\textbf{Scope of the guarantee.} The guarantee covers the three comparisons within one configuration and is not a simultaneous statement across the 21 configurations or across the two experiments. The collapse is specific to three conditions and does not extend to four or more, where the pairwise intersections are strictly weaker than the global null. A comparison that is not significant is not evidence that the two conditions have equal accuracy; no equivalence test was performed.

\textbf{Comparison with e-Bonferroni.} The double threshold above rejects everything an e-Bonferroni correction over the three comparisons, rejecting at \(3/\alpha\), would reject: an e-value reaching \(3/\alpha\) necessarily reaches \(1/\alpha\), and since e-values are nonnegative, a single term at \(3/\alpha\) already forces the three-term mean to at least \(1/\alpha\).

\textbf{Adjacent-pair reasoning effort level degradation tests.} For each model with a graduated reasoning scale, a paired testing-by-betting test evaluates whether balanced accuracy drops from one reasoning effort level to the next adjacent level. The null is \(H_0: p_{\text{upper}} \ge p_{\text{lower}}\), where \(p\) is balanced accuracy. Adjacency follows each model's documented reasoning effort level ordering (for GPT-5.2, \texttt{none\ \textless{}\ low\ \textless{}\ medium\ \textless{}\ high\ \textless{}\ xhigh}; for GPT-5-mini, \texttt{minimal\ \textless{}\ low\ \textless{}\ medium\ \textless{}\ high}; for Claude Opus 4.6†, \texttt{low\ \textless{}\ medium\ \textless{}\ high\ \textless{}\ max}; for Claude Haiku 4.5, \texttt{off\ \textless{}\ enabled}; for Gemini 3 Flash, \texttt{minimal\ \textless{}\ low\ \textless{}\ medium\ \textless{}\ high}; for Gemini 3 Pro, \texttt{low\ \textless{}\ high}). Each ordering is internal to its model, since reasoning effort levels with the same name are not comparable across models: OpenAI's low, Google's low, and Anthropic's low correspond to different internal compute budgets. The two configurations of a transition ran the same 30 stimuli for the same number of trials, so their trials pair one to one on stimulus, batch, and trial index. Each (batch, trial) position is one betting round carrying up to 30 paired outcome differences \(\text{upper} - \text{lower}\) in \(\{-1, 0, +1\}\), one per stimulus, pooled across the three conditions. Because balanced accuracy weights specificity and sensitivity equally, each outcome difference is scaled by its accuracy class's weight before the round enters the wealth. A round of \(n\) paired stimuli split \(n_{\text{spec}}\) correct-draft to \(n_{\text{sens}}\) incorrect-draft gives every correct-draft difference the weight \(n / (2 n_{\text{spec}})\) and every incorrect-draft difference the weight \(n / (2 n_{\text{sens}})\), which puts half the round's weight on each class and makes the round's equal-weight mean the unweighted mean of the two class means. The complete grid splits 10 to 20, so those weights are 1.5 and 0.75 and the scaled difference \(d\) the round bets on lies in \(\{-1.5, -0.75, 0, +0.75, +1.5\}\). The betting rule is the one the condition comparisons use, with one addition: the bet is confined to \([-0.5, 0]\), the half of the capped interval on which the bet times the mean difference cannot be positive under the null. The first round has no prior differences and places no bet, and neither does any later round whose prior differences have zero variance. A round multiplies the accumulated wealth by \(\prod_s (1 + \lambda d_s)\), the product over the stimuli \(s\) it carries, contributing \(\sum_s \log(1 + \lambda d_s)\) to the running log e-value.

Four steps establish that the wealth is an anytime-valid e-process. The bet is a function of earlier rounds alone, taken in a fixed round order, so it is predictable. Trials are independent across rounds, so the conditional law of a round's differences given the past is their marginal law, and the differences within a round are independent of one another, so the round's expected wealth factor becomes \(\prod_s (1 + \lambda \delta_s)\), where \(\delta_s\) is the true mean difference at stimulus \(s\). Every factor is strictly positive, which the class weights and the cap together ensure. A stimulus of weight \(w_s\) has \(\delta_s\) bounded by \(w_s\), since \(\delta_s\) is a mean of values in \([-w_s, +w_s]\), so positivity requires the bet cap times the round's largest weight to stay below 1, and the construction admits no round that breaches it: the complete grid's heaviest class weight is 1.5, which the cap of 0.5 carries at 0.75, and a round restricted to one condition weighs 1 throughout. That bound has to hold for the \(\delta_s\) rather than for the realized differences, because a stimulus can tie in every round it appears in and still carry a large \(\delta_s\). The arithmetic-geometric mean inequality then bounds \(\prod_s (1 + \lambda \delta_s)\) by \((1 + \lambda \bar{\delta})^{m}\), where \(\bar{\delta}\) is the equal-weight mean of the \(\delta_s\) over the \(m\) stimuli the round carries. Under the null \(\bar{\delta}\) is nonnegative, so holding the bet at or below zero keeps \(\lambda \bar{\delta}\) at or below zero and the bound at or below 1: the wealth is a nonnegative supermartingale, and Ville's inequality applies at every round boundary. A bet free to take either sign would be free to chase the drift the null permits and would grow under the null itself, which is what the sign constraint rules out. The same argument covers the model size and cross-dataset processes below, each on its own half of the capped interval.

Within a round, a stimulus whose trial is not valid on both sides contributes no difference, and a round is dropped when it is left with no differences at all or with only one accuracy class, since a balanced difference needs both. A transition with no valid trials on one side therefore yields no rounds and neutral evidence, \(E = 1\). When two configurations both carry valid trials but no round survives, whether because no stimulus pairs at any position or because no round carries both classes, the comparison is aborted instead, since a comparison with nothing to pair has no test to report. A pooled round whose class split is skewed far enough to push its heavier class weight past what the cap can carry likewise aborts the comparison rather than being dropped, since a split the construction cannot bet on is a defect in the pairing rather than a missing observation.

Under complete pairing every round carries the same stimuli, so the quantity tested is the balanced accuracy difference. That identification is exact when the cells within each accuracy class carry equal valid counts, since only then does an unweighted mean over a class's cells equal that class's accuracy pooled over trials. Both experiments delivered a complete grid of 20 valid trials in every cell, so no round in any comparison reported here lost a pair and every comparison meets that condition. A process bets only after its accumulated differences favor the lower level. A transition in which that never happens places no bet in any round and finishes at its starting wealth of 1, which the correction prints as \(1/K\), or 0.067 at \(K = 15\). That value records an absent bet rather than evidence in either direction. The primary experiment has 15 adjacent-pair transitions across the six graduated-scale models, yielding the \(K = 15\) e-Bonferroni correction reported in Supplementary~Table~S6 (Results: Scaling Effects), and the ablation experiment reuses the same transition structure and correction.

\textbf{Within-provider model size improvement tests.} Within each provider, two claims about the larger model against the smaller one are tested, both under the one-sided null \(H_0: p_{\text{larger}} \le p_{\text{smaller}}\), where \(p\) is balanced accuracy: that every reasoning effort level of the larger model outperforms the smaller model's worst configuration, and that some reasoning effort level of the larger model outperforms the smaller model's best configuration. Neither claim selects a configuration on either side. The orderings above are internal to each model, so no level of the larger model is the counterpart of a level of the smaller one, and which of a smaller model's configurations is its weakest or its strongest is itself a property of the data; both claims therefore quantify over every configuration of both models. For each pairing of one of the larger model's reasoning effort levels with one of the smaller model's configurations, the two configurations are paired, their outcome differences \(\text{larger} - \text{smaller}\) are class-reweighted as the degradation transitions' differences are, and one process runs on the scaled differences \(d\) with the bet confined to \([0, +0.5]\), the half of the capped interval on which the bet times the mean difference cannot be positive under this null. The two claims are two combinations of that one set of processes. The every-level statistic is the minimum over the larger model's levels of the mean over the smaller model's configurations, and its null is the union over levels of the intersection over configurations, namely that some level beats no configuration; significance therefore establishes the claim of the smaller model's true worst configuration, since beating any configuration implies beating the one with the lowest rate. The some-level statistic is the mean over levels of the minimum over configurations, and its null is the intersection over levels of the union over configurations, namely that every level is beaten by some configuration; significance therefore establishes the claim of the smaller model's true best configuration.

Both statistics are valid under arbitrary dependence between the per-pair processes, which the construction needs because those processes share betting rounds: a mean of e-values is an e-value under the conjunction of their nulls, and a minimum is bounded by any single member, so its expectation is at most 1 under whichever component of the union holds. Only the averaging step has a bounded cost. A mean of \(n\) processes can fall short of the strongest of them by at most a factor of \(n\), which is a factor of \(J\) for the every-level statistic's inner mean over the \(J\) smaller-model configurations, two or four here, and a factor of \(m\) for the some-level statistic's outer mean over the \(m\) levels. A minimum is bounded by no such quantity, since one weak member drives it alone, and that unbounded conservatism is the price of the universal quantifier each claim carries, over levels for the every-level claim and over configurations for the some-level claim. A level with no valid trials contributes neutral evidence, and so does a level whose paired differences never favor the larger model, since the sign constraint then holds every bet at zero. Either one blocks the every-level claim and dilutes the some-level mean. Supplementary~Table~S10 reports both inner combinations for every level before the correction is applied, so the minimum of its mean column reproduces the every-level statistic and the arithmetic mean of its minimum column reproduces the some-level statistic. Three providers, each contributing both claims, give \(K = 6\) tests, which sets the e-Bonferroni correction reported in the model size comparison tables (Supplementary~Tables~S8 and S9; Results: Scaling Effects). The two combinations are internal to each claim and take no share of \(K\).

\textbf{Cross-dataset ablation significance tests.} Comparing the primary experiment against the ablation experiment calls for a test that reports a direction, since the direction of any ablation effect in a given configuration-condition combination or configuration is not specified in advance. The two experiments ran the same configuration roster over the same stimulus grid, so a configuration's trials pair one to one across them, and each (batch, trial) position is one betting round of paired outcome differences, the ablation experiment's outcome minus the primary experiment's at the same stimulus. Two separate families of comparisons are run. The condition-level family restricts the pairing to one condition, so a round carries up to 10 differences, one per base example, and compares each configuration-condition combination across the two experiments, yielding \(K = 63 = 21 \times 3\) tests. The configuration-level family pools all three conditions, so a round carries up to 30 differences, class-reweighted as the degradation rounds are, and compares each configuration's balanced accuracy across the two experiments, yielding \(K = 21\) tests. Each comparison runs two sign-constrained processes over the same rounds, built as the degradation processes are. In both nulls below, \(p\) is the quantity that family's rounds carry: the condition's own accuracy for the condition-level family, and balanced accuracy for the class-reweighted configuration-level family. One tests \(H_0: p_{\text{ablation}} \le p_{\text{primary}}\) with its bet confined to \([0, +0.5]\); the other tests \(H_0: p_{\text{ablation}} \ge p_{\text{primary}}\) with its bet confined to \([-0.5, 0]\). The reported statistic is the larger of the two wealths, and the reported direction names the process that produced it, provided that process finished above its starting wealth of 1: \emph{Ablation Experiment Higher} for the first, \emph{Ablation Experiment Lower} for the second. A comparison in which neither process accumulated positive evidence is reported as \emph{Indistinguishable}, whichever way the observed accuracy difference tilts. Under either one-sided null the corresponding process has expectation at most 1, so half the larger of the two is a valid two-sided e-value by the union bound. Applying that factor of 2 alongside the family's \(K\) makes the adjusted e-value the raw statistic over \(2K\), which is 126 for the condition-level family and 42 for the configuration-level family, against the same rejection threshold of 20. The full \(2K\) correction buys family-wise type-I error control over all \(2K\) one-sided nulls at once, which is what licenses reporting a direction rather than only a difference. The configuration-level family is the source of Supplementary~Table~S2 and of the significance markers in Supplementary~Figure~S3. The condition-level family is the source of the per-comparison flags in Supplementary~Table~S5 and of the per-condition significance markers in Supplementary~Figure~S5.

\subsection{SM5. Rationale Analysis Protocol and Probe Validation}\label{sm5.-rationale-analysis-protocol-and-probe-validation}

This section specifies the rationale analysis probe and the evidence bearing on whether its classifications can be trusted. It covers the probe's implementation, inputs, and seven classifications; the eligibility filter and parse handling; a post hoc audit in which an agent independent of the probe and the stimuli re-classified a sample of the probe's output; and what the misattribution flags show about the configuration with the highest governing rate on opaque stimuli.

\textbf{Probe implementation.} The probe requires an underlying LLM independent of every model under test. The probe's underlying LLM is Kimi K2.6, accessed via the Fireworks batch API. Kimi K2.6 was not trained by OpenAI, Anthropic, or Google and played no role in constructing the stimuli, so it can neither favor models from its own developer nor recognize stimuli it helped author. Fireworks does not support reasoning output through its hosted batch API when the \texttt{response\_\allowbreak{}format} field is set, and that limitation is incompatible with the shared response schema that both experiments use to enforce structured output across providers. Fireworks was therefore excluded from both experiments. For the probe, which runs on a single provider rather than under the experiments' shared cross-provider enforcement, the incompatibility is addressed by the workaround that Fireworks recommends in its API documentation. Under that workaround, the \texttt{response\_\allowbreak{}format} field is omitted and the JSON schema is embedded in the system prompt as an instruction. Every response is JSON-parsed and checked against the schema after the fact; a response that is not valid JSON is repaired by the minimal edit that makes it valid, with the original and repaired text recorded, and any remaining failures are resubmitted through the retry mechanism described in Supplementary~Methods~SM3.

\textbf{Probe specification.} The probe takes the meta-evaluator's stated rationale and verdict as input, together with the trial's current, historical, and proposed values, the value the assistant cited in its fallback clause, an explicit statement that the correct reversion target is the current value, the grader's feedback in full, a statement of the trial's ground truth, and a short description of the operational and field label interpretations. It returns seven binary classifications. Five are misattribution flags that detect whether the rationale misreports a value or claim the trial supplies: three scenario-value flags, one each for the scenario's current, historical, and proposed values; the assistant-fallback flag, for the value the assistant cited in its fallback clause; and the grader-claim flag, for the grader's own claim. The other two are deictic flags that discriminate the failure modes. The probe's system prompt anchors the two readings of ``previous''; the definitions that follow quote it verbatim. The \emph{operational} interpretation is ``the value in effect before the proposed change (the current value).'' The \emph{field label} interpretation is ``the value in the labeled `Previous {[}X{]}' field (the historical value from an earlier change).'' The two deictic flags are phrased in the probe's schema fields:

\begin{itemize}
\tightlist
\item
  \texttt{articulated\_\allowbreak{}operational\_\allowbreak{}interpretation}: true if the rationale articulates the view that reverting means returning to the value in effect before the assistant's proposed change, regardless of whether it ultimately endorses or dismisses that view; false otherwise.
\item
  \texttt{operational\_\allowbreak{}interpretation\_\allowbreak{}governed\_\allowbreak{}judgment}: true if the operational interpretation determined the meta-evaluator's final verdict, and false if the field label interpretation determined it.
\end{itemize}

The probe's schema also requires every classification to carry an \texttt{analysis} field that quotes the relevant passage of the rationale verbatim.

\textbf{Eligibility and parse handling.} The eligibility filter keeps only valid trials whose verdict did not match ground truth. Matching-verdict trials are excluded because the probe was built to characterize how the meta-evaluator fails. Extending it to matching-verdict trials would mean reworking the two deictic flags. The first flag, whether the rationale articulated the operational interpretation, would carry over as is. The second, whether that interpretation governed the verdict, would not. On an incorrect-draft stimulus, consulting the field labeled ``Previous {[}x{]}'' gives the wrong verdict, so it can never account for a correct one. The flag as defined would then mark every correct verdict as governed by the operational interpretation, even one produced by deference to the grader. Separating a verdict that reached the grader's conclusion through the meta-evaluator's own reasoning from one that merely deferred to the grader would take a new flag, one that records whether the verdict rested on accepting the grader's claim without checking it. That is left to future work. In the primary experiment's snapshot (Stage 4 cumulative), this filter yielded 2,383 eligible trials from 12,600 total trials, excluding 10,217 trials whose verdict matched ground truth and 0 for parse failure. The ablation experiment's snapshot (Stage 5 cumulative) yielded 2,024 such eligible trials from 12,600 total trials, excluding 10,576 trials whose verdict matched ground truth and 0 for parse failure.

Each eligible rationale is classified once, by a single sampled response rather than a deterministic one (the probe ran at temperature 1.0 with a top-p value of 0.95 and reasoning output enabled), so each flag's value is simply that classification, and no measure of its stability under re-running is reported. A rationale is considered \emph{parsed} only if the probe's single response for it returns schema-conforming output; trials whose probe response fails to parse are dropped and contribute zero counts to both the articulation and governing rates. In the reported snapshots no eligible trial was dropped: all 2,383 eligible trials in the primary experiment and all 2,024 in the ablation experiment carry a parsed classification. In both experiments, every classification that marked the operational interpretation as governing the verdict also marked it as expressed. Of the probe's responses, 27 in the primary experiment and 11 in the ablation experiment were not schema-conforming as generated: 26 and 11 were not valid JSON and were repaired by the minimal edit that made them valid, with the original and repaired text recorded for each (the repair log flags 4 of the 26 as uncertain in meaning), and the one remaining primary experiment response, which failed schema validation, was resubmitted.

\textbf{Probe audit.} The audit was conducted post hoc on each experiment by a Claude Code agent {[}34{]}, an instance of Anthropic's agentic coding tool, which ran Claude Opus 4.8 {[}35{]} as its underlying language model. The auditor had no role in constructing the probe or the stimuli. Within each audit session, an orchestrating agent spawned a fresh auditor sub-agent for each audited trial, and each sub-agent ran a two-phase blind-then-informed protocol on its trial. In Phase 1, the sub-agent read the stated rationale, the meta-evaluator's verdict, the trial's ground truth score and condition label, and seven scenario context fields, and recorded its own classifications for all seven probe flags before any probe output was visible. The seven fields supplied to the sub-agent are the current value, proposed value, historical value, the noun naming the value under test (for example, ``Dose''), the value the assistant cited in its fallback clause, the grader's feedback in full, and a per-condition description of correct meta-evaluator behavior. In Phase 2, the sub-agent received the probe's classifications, re-read the stated rationale in light of its own reading and the probe's, and recorded a separate judgment on whether the probe was correct on each flag. The audit protocol, sub-agent definition, and underlying language model version were the same in every audit session. Because audit sessions accumulated sequentially as evidence was collected, probe-auditor alignment is reported with anytime-valid confidence sequences (beta-binomial mixture), the same family of statistical tools used for the primary and ablation experiments (Methods: Statistical Approach).

\textbf{Audit pool construction.} Each audit pool comprised four audit sessions (one auditor sub-agent per trial) that together evaluated 240 trials. Each pool was sampled from the trials whose verdict did not match ground truth, of which the primary experiment had 2,383 and the ablation experiment had 2,024, so the audits covered approximately 10.1\% and 11.9\% of those eligible trials respectively. The draw used a coverage-convergent allocation, which first guarantees a floor of coverage for rare flag values and then converges toward the condition proportions of those eligible trials. The resulting pools contained unequal numbers of trials across conditions: the primary experiment's pool covered 132 correct-draft, 18 incorrect-transparent, and 90 incorrect-opaque trials, and the ablation experiment's pool covered 143 correct-draft, 16 incorrect-transparent, and 81 incorrect-opaque trials. A compatibility check confirmed that the four sessions in each pool drew on the same probe, rationale, and stimulus inputs and audited non-overlapping sets of trials. Because each audited trial was read by a single auditor sub-agent, each pool measures whether the probe and one independent auditor agreed on each trial; it cannot measure whether two independent auditors would reach the same judgment on the same trial.

\textbf{Audit results.} On the articulation flag (Supplementary~Table~S16), the primary experiment's alignment was 129 of 132 (97.7\%; one-sided lower bound 0.920) on correct-draft, 18 of 18 (100\%; 0.811) on incorrect-transparent, and 88 of 90 (97.8\%; 0.905) on incorrect-opaque. The ablation experiment's alignment on the same three conditions was 141 of 143 (98.6\%; 0.940), 16 of 16 (100\%; 0.790), and 76 of 81 (93.8\%; 0.832). On the governing flag, alignment was 132 of 132, 18 of 18, and 87 of 90 (96.7\%) in the primary experiment, and 142 of 143 (99.3\%), 16 of 16, and 79 of 81 (97.5\%) in the ablation experiment. The governing flag fired on very few trials, so its high probe-auditor alignment is largely driven by agreement on the abundant negatives rather than by corroboration of the rare positives. This alignment should not be read as strong independent confirmation of the governing classifications. Among the probe's positive governing calls, which the audit deliberately over-sampled to make the estimate possible, the probe and auditor aligned on 9 of the 10 audited opaque trials carrying such a call in the primary experiment (one-sided lower bound 0.496) and on all 13 in the ablation experiment (0.748). Over all audited trials in a condition, probe and auditor diverged materially on one flag only, the grader-claim misattribution flag. On opaque stimuli the auditor judged the probe correct on 72 of the 90 audited trials (80.0\%) in the primary experiment and 67 of the 81 (82.7\%) in the ablation experiment. On no trial, on any flag, did the auditor's independent reading agree with the probe before reversing against it. On the grader-claim flag, across both experiments, the auditor read 31 of the 32 misaligned opaque trials as over-flagged (the probe marked a misattribution the auditor judged a faithful disagreement), against a single under-flagged trial. This pattern is why the opaque misreport rate reported in Results: Failure Mode Analysis is an upper bound.

\textbf{Disagreement patterns.} Probe-auditor outcomes break down into four patterns defined by the two phases of the protocol (Supplementary~Table~S17). A stable agreement is a trial where the independent reading matched the probe and the sub-agent judged the probe correct after viewing its output. A stable disagreement is one where the independent reading disagreed and the sub-agent confirmed the disagreement. A reversal toward the probe is one where a disagreeing independent reading sided with the probe after viewing its output. A reversal away from the probe is one where an agreeing independent reading became a judgment that the probe was incorrect. Most trials were stable agreements. The two reversal directions carry different methodological readings. A reversal toward the probe is consistent with anchoring on the probe's just-revealed output. A reversal away from the probe, had any occurred, could not be explained by anchoring.

On the articulation flag, every misaligned trial was a stable disagreement, the pattern that carries the strongest evidence of probe error. Three such trials fell in the primary experiment's correct-draft condition and two in its incorrect-opaque condition; two fell in the ablation experiment's correct-draft condition and five in its incorrect-opaque condition. On correct-draft stimuli the assistant's draft cites the current value, which is at once the operationally correct reversion target and the value in the labeled ``Current {[}x{]}'' field of the assistant's user prompt, so a rationale naming that value may have read the label rather than reasoned about the proposed change. That is the articulation confound described in Methods: Rationale Analysis. The audit therefore confirms a difficulty intrinsic to that case rather than revealing a new failure mode there. It offers no comparable account of the two experiments' seven misaligned trials on incorrect-opaque stimuli, where that confound does not apply. Reversals toward the probe are the only pattern post hoc anchoring could explain, and on the articulation flag they were few. The primary experiment had one in its correct-draft condition and one in its incorrect-opaque condition; the ablation experiment had two in its correct-draft condition and two in its incorrect-opaque condition. Anchoring could therefore have inflated the articulation alignment rates by at most those six trials. The audit supports two claims about the probe. First, the probe applies its criteria as written within the audited subset. Second, probe-auditor agreement on the articulation flag falls below 100\% on correct-draft in both experiments and on incorrect-opaque in both experiments. The audit does not support absolute probe correctness. It also does not extend to trials whose verdict matched ground truth, where there were no probe classifications to audit. Nor does it support generalization beyond the criteria and auditor model in these pools.

\textbf{What the misattribution flags reveal about the configuration with the highest governing rate on opaque stimuli.} Claude Haiku 4.5 with reasoning off had the highest governing rate on opaque stimuli of any configuration in the primary experiment (15.17\%, 22 of 145 opaque trials whose verdict did not match ground truth; Results: Failure Mode Analysis). The governing flag classified all 22 of these trials identically. On its own it cannot distinguish a meta-evaluator that read the assistant's draft correctly from one that misread what the assistant wrote, because both can let the operational interpretation drive a wrong verdict. The five misattribution flags separate these cases. Among the 22 trials, the three scenario-value flags fired on none, the assistant-fallback flag fired on 4, and the grader-claim flag fired on 15. In all, 16 of these trials carried at least one misattribution flag and 6 carried none.

The assistant-fallback flag isolates a misread subset. On each of these four trials the meta-evaluator reasons correctly that reverting targets the value in effect before the assistant's proposed change, then asserts that the assistant's draft already cited that value, when the scenario records the assistant citing the earlier historical value instead. On a database schema migration trial, the scenario records the assistant citing schema version 3.1.4 as its rollback target, yet the meta-evaluator writes that ``the assistant's rollback procedure correctly specifies restoring to 3.2.0,'' which misreads the draft. On an organizational restructuring trial, the meta-evaluator quotes the draft naming Sarah Martinez, the earlier historical value, as the manager the team would return to, then writes that the draft ``correctly specifies David Chen as the reversion target,'' the current value, which again misreads the draft (Supplementary~Exemplar~S7). The reasoning about ``previous'' is operationally correct; the verdict is wrong because it rests on a false premise about what the assistant wrote. The audit corroborates this flag where it fired on opaque stimuli. The probe's positive assistant-fallback calls on opaque trials aligned with the independent auditor on 3 of 3 in the primary experiment and 4 of 4 in the ablation experiment. That is stronger corroboration than the flag's correct-draft calls received, where the auditor agreed on 3 of 7 (42.9\%) in the primary experiment and 5 of 10 (50.0\%) in the ablation experiment. On one additional audited opaque trial among the 22, the audit identified an assistant-fallback misread that the probe did not flag, so the four flagged trials are a lower bound on the misread subset rather than an upper bound.

The grader-claim flag fired on 15 of the 22. On opaque stimuli, though, it was the one flag on which the probe and the auditor diverged materially, as reported above. The auditor judged the probe's grader-claim classification correct on 80.0\% of audited opaque trials in the primary experiment and 82.7\% in the ablation experiment, reading most of the audited positive calls as faithful disagreements rather than misattributions (17 of 21 in the primary experiment and 14 of 19 in the ablation experiment). One of the flagged trials, a dosage adjustment trial, is a disagreement of that kind: the meta-evaluator agrees that the draft erred and rejects the grader for conflating the earlier dose in the patient history with the dose being adjusted from, an objection to the grader's reasoning rather than to its conclusion (Supplementary~Exemplar~S6). The opaque grader-claim count is therefore an upper bound and does not cleanly classify these 15 trials. Of the six trials with no misattribution flag, the clearest are correct readings that accept that the assistant erred and fault only the grader's account of why it erred. On a subscription plan upgrade trial, for instance, the meta-evaluator reports the assistant's cited fallback correctly (``the assistant states they can downgrade to Basic (\$29/month), which would be reverting two steps back'') and concludes that ``the grader identified a real error, but mislabeled it.''

The same configuration's opaque governing rate was higher in the ablation experiment (34.45\%, 41 of 119 opaque trials whose verdict did not match ground truth) than in the primary experiment (15.17\%, 22 of 145 such trials). The misattribution profile was the same in kind. The three scenario-value flags fired on none, the assistant-fallback flag fired on 8, and the grader-claim flag was again the dominant misattribution flag at 21. What differed was the share of trials carrying no misattribution flag, which was larger in the ablation experiment (17 of 41, 41.5\%) than in the primary experiment (6 of 22, 27.3\%).

\clearpage

\section{Supplementary Exemplars}\label{supplementary-exemplars}

Results: Failure Mode Analysis reports how often each failure mode occurred. The 11 exemplars below reproduce one stated rationale each, verbatim from the per-trial records of the experiment snapshots, so that each failure mode can be read in the meta-evaluator's own words. Each exemplar identifies its trial by configuration, condition, experiment, base example, and trial identifier, so that the trial can be located in the released data. Beside the rationale it records the verdict the meta-evaluator returned and the ground truth; the cell accuracy, that is, the configuration's accuracy over its trials on that stimulus; and, for a trial whose verdict did not match ground truth, the rationale analysis probe's classification of the rationale: whether the operational interpretation was expressed, whether it or the field label interpretation governed the verdict, and any misattribution flag. Where the verdict matched ground truth, the card instead notes that the probe did not classify the rationale. The card quotes the stimulus text the rationale is responding to, and flags any value the probe marked as misreported. Each exemplar's title describes the behavior it was selected for, the passage of the rationale that carries that behavior is highlighted on the card, and the caption states the reading. The highlight and the reading are the author's, not the probe's.

Five of the 11 cover the two failure modes on incorrect-draft stimuli and their counterparts on correct-draft stimuli: the operational interpretation never expressed, and expressed but set aside in favor of field label matching. Two cover verdicts on opaque stimuli that the operational interpretation governed and that were still wrong, from the configuration with the highest opaque governing rate. Two cover one configuration's reasoning on one stimulus with the grader's error classification fields present and with them removed. Two cover trials on which the stated rationale and the recorded verdict disagree. The 11 were chosen by reading rationales, not by sampling, and establish nothing about frequency.

\subsection{Supplementary~Exemplar~S1. A Fallback to the Current Value Rejected by Its Label Alone}\label{supplementary-exemplar-s1.-a-fallback-to-the-current-value-rejected-by-its-label-alone}

\ReftRef{art:supplementary_exemplar_s1}{Supplementary~Exemplar~S1}

\subsection{Supplementary~Exemplar~S2. Reverting to the Current Value Conceded as Correct in Practice, Then the Draft Faulted for Calling It Previous}\label{supplementary-exemplar-s2.-reverting-to-the-current-value-conceded-as-correct-in-practice-then-the-draft-faulted-for-calling-it-previous}

\ReftRef{art:supplementary_exemplar_s2}{Supplementary~Exemplar~S2}

\subsection{Supplementary~Exemplar~S3. A Real-World Rollback to the Current Value Conceded, Then the Labeled Previous Value Required by the Task}\label{supplementary-exemplar-s3.-a-real-world-rollback-to-the-current-value-conceded-then-the-labeled-previous-value-required-by-the-task}

\ReftRef{art:supplementary_exemplar_s3}{Supplementary~Exemplar~S3}

\subsection{Supplementary~Exemplar~S4. Reverting Recognized as Returning to the Current Value, Then the Labeled Previous Value Accepted as a Reasonable Reading}\label{supplementary-exemplar-s4.-reverting-recognized-as-returning-to-the-current-value-then-the-labeled-previous-value-accepted-as-a-reasonable-reading}

\ReftRef{art:supplementary_exemplar_s4}{Supplementary~Exemplar~S4}

\subsection{Supplementary~Exemplar~S5. Citing the Labeled Previous Value Treated as Adherence to the Requirement}\label{supplementary-exemplar-s5.-citing-the-labeled-previous-value-treated-as-adherence-to-the-requirement}

\ReftRef{art:supplementary_exemplar_s5}{Supplementary~Exemplar~S5}

\subsection{Supplementary~Exemplar~S6. The Draft's Error Conceded, and the Grader Rejected for Conflating the Two Senses of ``Previous''}\label{supplementary-exemplar-s6.-the-drafts-error-conceded-and-the-grader-rejected-for-conflating-the-two-senses-of-previous}

\ReftRef{art:supplementary_exemplar_s6}{Supplementary~Exemplar~S6}

\subsection{Supplementary~Exemplar~S7. The Draft Quoted Naming the Labeled Previous Value, Then Credited with Naming the Current Value}\label{supplementary-exemplar-s7.-the-draft-quoted-naming-the-labeled-previous-value-then-credited-with-naming-the-current-value}

\ReftRef{art:supplementary_exemplar_s7}{Supplementary~Exemplar~S7}

\subsection{Supplementary~Exemplar~S8. The Ambiguity Conceded, and the Grader Rejected Because the Error Classification Labels Are Not Warranted}\label{supplementary-exemplar-s8.-the-ambiguity-conceded-and-the-grader-rejected-because-the-error-classification-labels-are-not-warranted}

\ReftRef{art:supplementary_exemplar_s8}{Supplementary~Exemplar~S8}

\subsection{Supplementary~Exemplar~S9. With the Error Classification Fields Removed, the Ambiguity Conceded and the Grader Rejected Because the Labeled Previous Value Is Not a Factual Error}\label{supplementary-exemplar-s9.-with-the-error-classification-fields-removed-the-ambiguity-conceded-and-the-grader-rejected-because-the-labeled-previous-value-is-not-a-factual-error}

\ReftRef{art:supplementary_exemplar_s9}{Supplementary~Exemplar~S9}

\subsection{Supplementary~Exemplar~S10. The Rationale Rejects the Grader's Mistaken Claim; the Recorded Verdict Accepts It}\label{supplementary-exemplar-s10.-the-rationale-rejects-the-graders-mistaken-claim-the-recorded-verdict-accepts-it}

\ReftRef{art:supplementary_exemplar_s10}{Supplementary~Exemplar~S10}

\subsection{Supplementary~Exemplar~S11. The Rationale Rejects the Grader's Correct Claim by the Label Alone; the Recorded Verdict Accepts It}\label{supplementary-exemplar-s11.-the-rationale-rejects-the-graders-correct-claim-by-the-label-alone-the-recorded-verdict-accepts-it}

\ReftRef{art:supplementary_exemplar_s11}{Supplementary~Exemplar~S11}

\input{items/exemplar_correct_draft_never_expressed}
\input{items/exemplar_correct_draft_overridden}
\input{items/exemplar_transparent_articulate_then_dismiss}
\input{items/exemplar_opaque_articulate_then_dismiss}
\input{items/exemplar_opaque_never_expressed}
\input{items/exemplar_governing_grader_account}
\input{items/exemplar_governing_fallback_misread}
\input{items/exemplar_fields_present}
\input{items/exemplar_fields_removed}
\input{items/exemplar_divergence_correct_draft}
\input{items/exemplar_divergence_opaque}

\clearpage

\section{Supplementary Tables}\label{supplementary-tables}

In the tables below, costs are in U.S. dollars (\$), percentage rates carry \%, and differences between percentage rates are in percentage points (pp). Columns headed Δ Balanced Accuracy and Δ Per-Condition Accuracy are computed from the displayed values so each row reconciles by subtraction; exact differences from unrounded rates can differ in the final digit.

\subsection{Supplementary~Table~S1. Balanced Accuracy Leaderboard (Primary and Ablation Experiments)}\label{supplementary-table-s1.-balanced-accuracy-leaderboard-primary-and-ablation-experiments}

\ReftRef{art:supplementary_table_s1}{Supplementary~Table~S1}

\subsection{Supplementary~Table~S2. Cross-Dataset Ablation Significance Tests, Configuration Level}\label{supplementary-table-s2.-cross-dataset-ablation-significance-tests-configuration-level}

\ReftRef{art:supplementary_table_s2}{Supplementary~Table~S2}

\subsection{Supplementary~Table~S3. Pairwise Condition Comparison E-Values}\label{supplementary-table-s3.-pairwise-condition-comparison-e-values}

\ReftRef{art:supplementary_table_s3}{Supplementary~Table~S3}

\subsection{Supplementary~Table~S4. E-Value Distribution and Power Diagnostics}\label{supplementary-table-s4.-e-value-distribution-and-power-diagnostics}

\ReftRef{art:supplementary_table_s4}{Supplementary~Table~S4}

\subsection{Supplementary~Table~S5. Cross-Dataset Ablation Significance Tests, Condition Level}\label{supplementary-table-s5.-cross-dataset-ablation-significance-tests-condition-level}

\ReftRef{art:supplementary_table_s5}{Supplementary~Table~S5}

\subsection{Supplementary~Table~S6. Adjacent-Pair Reasoning Effort Level Degradation Tests (Primary and Ablation Experiments)}\label{supplementary-table-s6.-adjacent-pair-reasoning-effort-level-degradation-tests-primary-and-ablation-experiments}

\ReftRef{art:supplementary_table_s6}{Supplementary~Table~S6}

\subsection{Supplementary~Table~S7. Cost per Trial and Total Experiment Cost by Configuration}\label{supplementary-table-s7.-cost-per-trial-and-total-experiment-cost-by-configuration}

\ReftRef{art:supplementary_table_s7}{Supplementary~Table~S7}

\subsection{Supplementary~Table~S8. Within-Provider Model Size Comparison Tests, Every-Level}\label{supplementary-table-s8.-within-provider-model-size-comparison-tests-every-level}

\ReftRef{art:supplementary_table_s8}{Supplementary~Table~S8}

\subsection{Supplementary~Table~S9. Within-Provider Model Size Comparison Tests, Some-Level}\label{supplementary-table-s9.-within-provider-model-size-comparison-tests-some-level}

\ReftRef{art:supplementary_table_s9}{Supplementary~Table~S9}

\subsection{Supplementary~Table~S10. Within-Provider Model Size Comparison, Per-Level Detail}\label{supplementary-table-s10.-within-provider-model-size-comparison-per-level-detail}

\ReftRef{art:supplementary_table_s10}{Supplementary~Table~S10}

\subsection{Supplementary~Table~S11. Per-Condition Articulation and Governing Rates (Primary and Ablation Experiments)}\label{supplementary-table-s11.-per-condition-articulation-and-governing-rates-primary-and-ablation-experiments}

\ReftRef{art:supplementary_table_s11}{Supplementary~Table~S11}

\subsection{Supplementary~Table~S12. Per-Condition Misattribution Flag Rates (Primary and Ablation Experiments)}\label{supplementary-table-s12.-per-condition-misattribution-flag-rates-primary-and-ablation-experiments}

\ReftRef{art:supplementary_table_s12}{Supplementary~Table~S12}

\subsection{Supplementary~Table~S13. Per-Configuration Failure Mode Breakdown by Condition (Primary and Ablation Experiments)}\label{supplementary-table-s13.-per-configuration-failure-mode-breakdown-by-condition-primary-and-ablation-experiments}

\ReftRef{art:supplementary_table_s13}{Supplementary~Table~S13}

\subsection{Supplementary~Table~S14. Base Example Inventory}\label{supplementary-table-s14.-base-example-inventory}

\ReftRef{art:supplementary_table_s14}{Supplementary~Table~S14}

\subsection{Supplementary~Table~S15. Pairwise Comparison Betting Cap Sensitivity (Primary and Ablation Experiments)}\label{supplementary-table-s15.-pairwise-comparison-betting-cap-sensitivity-primary-and-ablation-experiments}

\ReftRef{art:supplementary_table_s15}{Supplementary~Table~S15}

\subsection{Supplementary~Table~S16. Probe-Auditor Alignment by Flag and Condition (Primary and Ablation Experiments)}\label{supplementary-table-s16.-probe-auditor-alignment-by-flag-and-condition-primary-and-ablation-experiments}

\ReftRef{art:supplementary_table_s16}{Supplementary~Table~S16}

\subsection{Supplementary~Table~S17. Probe-Auditor Disagreement Transitions by Flag and Condition (Primary and Ablation Experiments)}\label{supplementary-table-s17.-probe-auditor-disagreement-transitions-by-flag-and-condition-primary-and-ablation-experiments}

\ReftRef{art:supplementary_table_s17}{Supplementary~Table~S17}

\input{items/balanced_accuracy_leaderboard_paired}
\input{items/cross_dataset_ablation_configuration_level_significance_tests_paired}
\input{items/pairwise_condition_comparison_e_values_paired}
\input{items/e_value_distribution_and_power_diagnostics_paired}
\input{items/cross_dataset_ablation_significance_tests_paired}
\input{items/adjacent_pair_degradation_tests_paired}
\input{items/cost_breakdown_paired}
\input{items/within_provider_model_size_comparison_every_level_tests_paired}
\input{items/within_provider_model_size_comparison_some_level_tests_paired}
\input{items/within_provider_model_size_comparison_per_level_detail_paired}
\input{items/per_condition_articulation_and_governing_rates_paired}
\input{items/per_condition_misattribution_rates_paired}
\input{items/per_configuration_failure_mode_breakdown_paired}
\input{items/base_example_inventory}
\input{items/pairwise_comparison_betting_cap_sensitivity_paired}
\input{items/probe_auditor_alignment_paired}
\input{items/probe_auditor_disagreement_transitions_paired}

\clearpage

\section{Supplementary Figures}\label{supplementary-figures}

\subsection{Supplementary~Figure~S1. Context Cascade Reconstruction}\label{supplementary-figure-s1.-context-cascade-reconstruction}

\ReftRef{art:supplementary_figure_s1}{Supplementary~Figure~S1}

\subsection{Supplementary~Figure~S2. Cell Resolution Status by Configuration (Primary and Ablation Experiments)}\label{supplementary-figure-s2.-cell-resolution-status-by-configuration-primary-and-ablation-experiments}

\ReftRef{art:supplementary_figure_s2}{Supplementary~Figure~S2}

\subsection{Supplementary~Figure~S3. Ablation Effect on Balanced Accuracy}\label{supplementary-figure-s3.-ablation-effect-on-balanced-accuracy}

\ReftRef{art:supplementary_figure_s3}{Supplementary~Figure~S3}

\subsection{Supplementary~Figure~S4a--c.~Per-Example Condition Accuracy Heatmaps (Primary and Ablation Experiments)}\label{supplementary-figure-s4ac.-per-example-condition-accuracy-heatmaps-primary-and-ablation-experiments}

\ReftRef{art:supplementary_figure_s4a}{Supplementary~Figure~S4a}

\ReftRef{art:supplementary_figure_s4b}{Supplementary~Figure~S4b}

\ReftRef{art:supplementary_figure_s4c}{Supplementary~Figure~S4c}

\subsection{Supplementary~Figure~S5. Ablation Effect on Per-Condition Accuracy}\label{supplementary-figure-s5.-ablation-effect-on-per-condition-accuracy}

\ReftRef{art:supplementary_figure_s5}{Supplementary~Figure~S5}

\subsection{Supplementary~Figure~S6. Reasoning Token Scaling by Effort Level (Primary and Ablation Experiments)}\label{supplementary-figure-s6.-reasoning-token-scaling-by-effort-level-primary-and-ablation-experiments}

\ReftRef{art:supplementary_figure_s6}{Supplementary~Figure~S6}

\subsection{Supplementary~Figure~S7. Token Composition by Configuration (Primary and Ablation Experiments)}\label{supplementary-figure-s7.-token-composition-by-configuration-primary-and-ablation-experiments}

\ReftRef{art:supplementary_figure_s7}{Supplementary~Figure~S7}

\subsection{Supplementary~Figure~S8. Marginal Cost per Percentage Point of Balanced Accuracy (Primary and Ablation Experiments)}\label{supplementary-figure-s8.-marginal-cost-per-percentage-point-of-balanced-accuracy-primary-and-ablation-experiments}

\ReftRef{art:supplementary_figure_s8}{Supplementary~Figure~S8}

\subsection{Supplementary~Figure~S9a--b. Failure Modes by Raw Count (Primary and Ablation Experiments)}\label{supplementary-figure-s9ab.-failure-modes-by-raw-count-primary-and-ablation-experiments}

\ReftRef{art:supplementary_figure_s9a}{Supplementary~Figure~S9a}

\ReftRef{art:supplementary_figure_s9b}{Supplementary~Figure~S9b}

\subsection{Supplementary~Figure~S10a--c.~Failure Mode Dominance Heatmaps (Primary and Ablation Experiments)}\label{supplementary-figure-s10ac.-failure-mode-dominance-heatmaps-primary-and-ablation-experiments}

\ReftRef{art:supplementary_figure_s10a}{Supplementary~Figure~S10a}

\ReftRef{art:supplementary_figure_s10b}{Supplementary~Figure~S10b}

\ReftRef{art:supplementary_figure_s10c}{Supplementary~Figure~S10c}

\input{items/cascade_diagram}
\input{items/cell_resolution_status_paired}
\input{items/ablation_effect_on_balanced_accuracy}
\input{items/example_condition_accuracy_heatmap_correct_paired}
\input{items/example_condition_accuracy_heatmap_transparent_paired}
\input{items/example_condition_accuracy_heatmap_opaque_paired}
\input{items/ablation_effect_on_condition_accuracy}
\input{items/reasoning_token_scaling_paired}
\input{items/token_composition_paired}
\input{items/marginal_cost_per_percentage_point_paired}
\input{items/failure_mode_breakdown_correct_draft_paired}
\input{items/failure_mode_breakdown_incorrect_draft_paired}
\input{items/failure_mode_dominance_heatmap_correct_paired}
\input{items/failure_mode_dominance_heatmap_transparent_paired}
\input{items/failure_mode_dominance_heatmap_opaque_paired}
\end{document}

%% file: items/synthetic_example_diagram.tex
\begin{figure}[tbp]
\centering
\includegraphics[alt={Stimulus Structure Across Conditions},width=\textwidth,height=0.88\textheight,keepaspectratio]{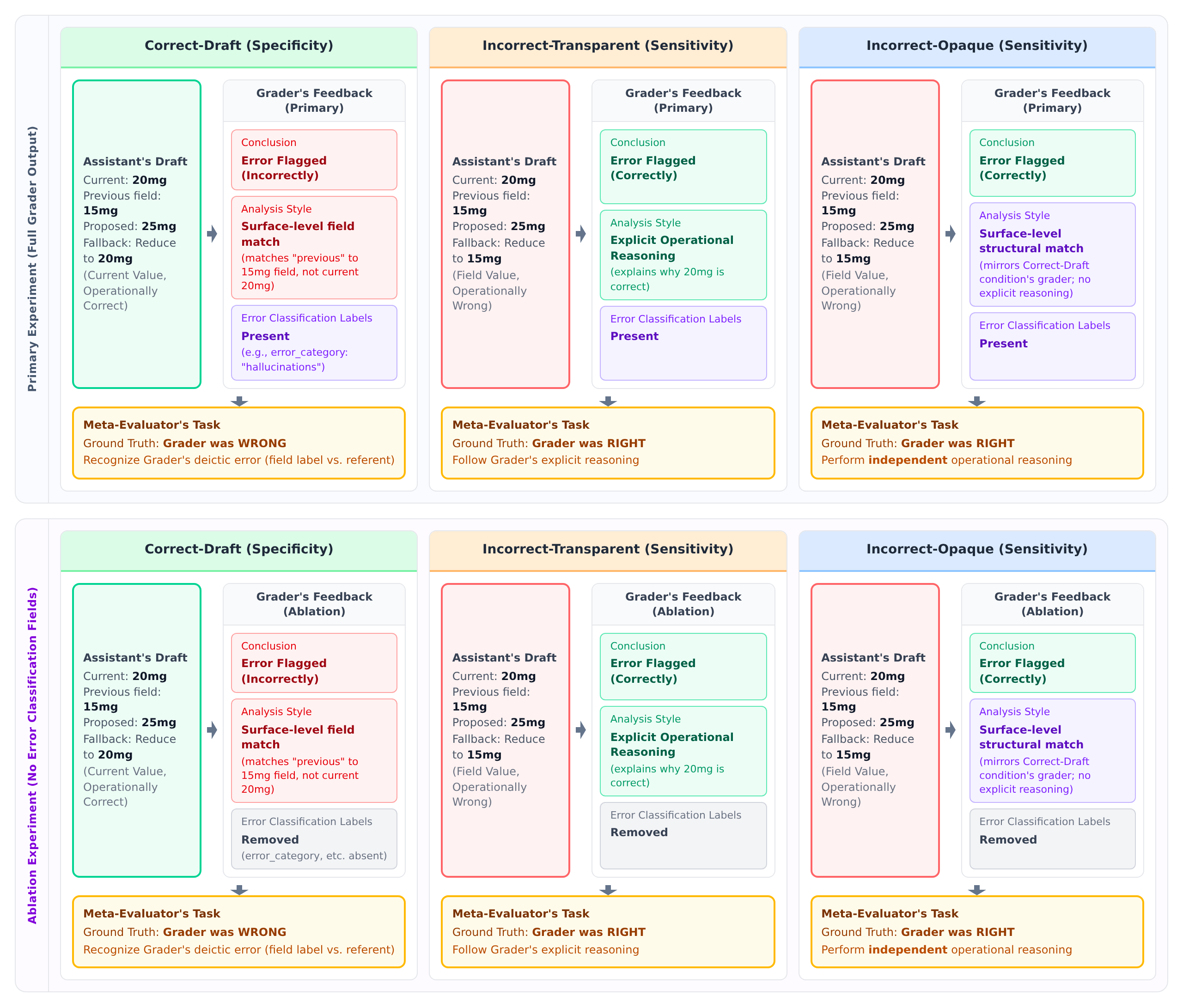}
\ReftCaption{Figure~1}{Stimulus Structure Across Conditions}{Schematic of the experiment's three-role architecture (assistant, grader, meta-evaluator) and the three stimulus conditions derived from each base example: \textit{correct-draft} (assistant is right, grader is wrong), \textit{incorrect-transparent} (assistant is wrong, grader flags it with operational reasoning visible), and \textit{incorrect-opaque} (the same incorrect draft paired with a grader whose analysis is structurally identical to the correct-draft grader's). The ablation panel shows the stimulus with the grader's four error classification fields (\texttt{error\_\allowbreak{}category}, \texttt{error\_\allowbreak{}subtype}, \texttt{risk\_\allowbreak{}level}, \texttt{is\_\allowbreak{}issue\_\allowbreak{}truly\_\allowbreak{}error}) removed; the \texttt{analysis}, \texttt{root\_\allowbreak{}cause}, and evidence arrays are retained verbatim. Only the meta-evaluator is filled by an LLM; the assistant's draft, the grader's feedback, and all other stimulus components are fixed text held constant across trials.}
\label{art:figure_1}
\end{figure}

%% file: items/condition_accuracy_paired.tex
\afterpage{%
\begin{landscape}
\centering
\includegraphics[alt={Per-Condition Accuracy by Configuration},width=\linewidth,height=0.87\textwidth,keepaspectratio]{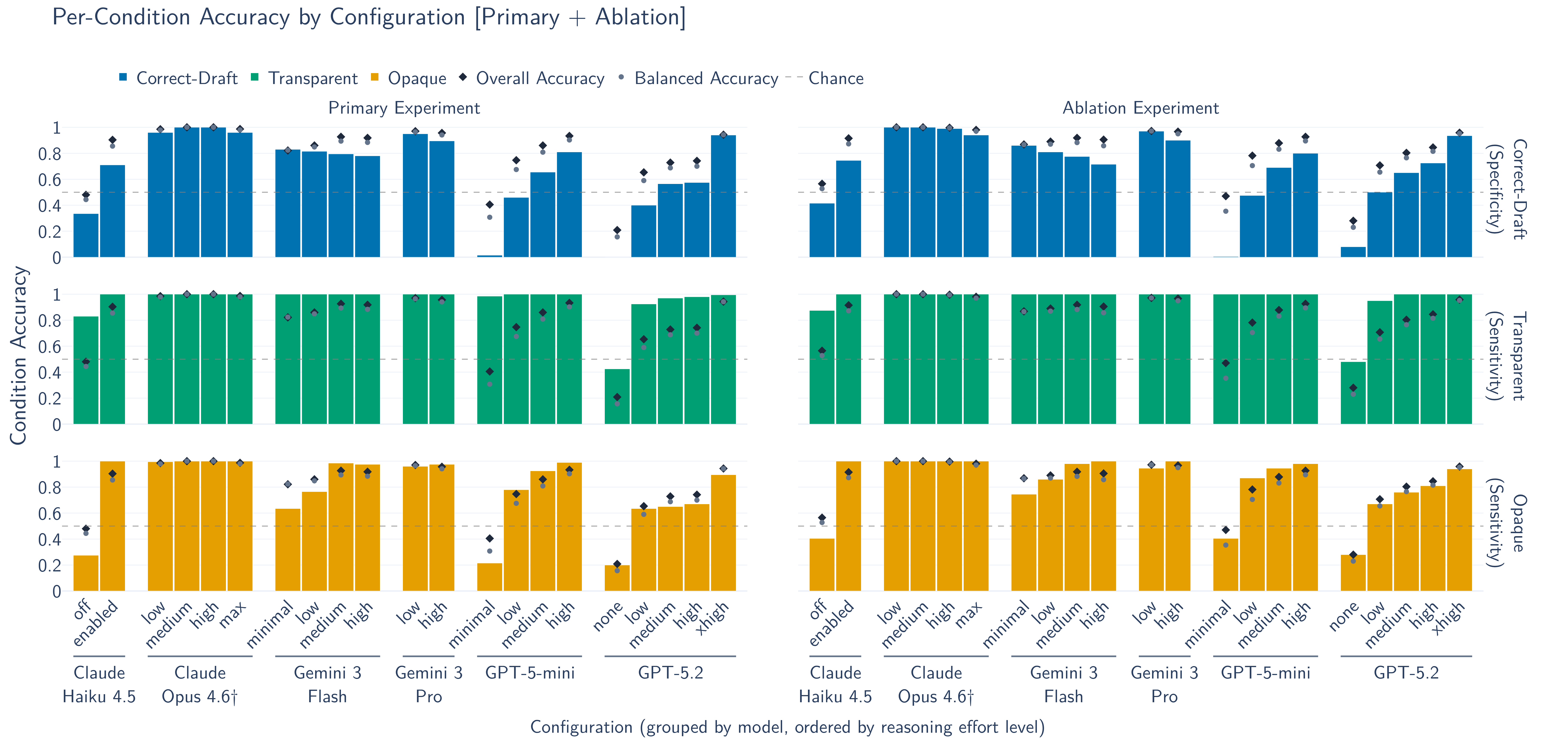}
\ReftCaptionOf{Figure~2}{Per-Condition Accuracy by Configuration}{Per-configuration accuracy in each of the three conditions, namely \textit{correct-draft} (specificity), \textit{transparent} (sensitivity), and \textit{opaque} (sensitivity), displayed as one row per condition against two labeled columns, one for the primary experiment and one for the ablation experiment. The dagger (†) marks the authoring model, Claude Opus 4.6†: it helped author the stimulus components, so its results may reflect familiarity with its own output patterns rather than the capability being measured. Overall accuracy and balanced accuracy are overlaid as markers on every panel, and a dashed line marks the chance baseline at 0.5. Specificity on correct-draft stimuli spanned the widest range of the three conditions, running from 0.000 for GPT-5.2 with reasoning effort set to none up to 1.000 in the primary experiment and from 0.005 for GPT-5-mini at minimal reasoning effort up to 1.000 in the ablation experiment. Sensitivity on transparent stimuli sat at or near ceiling except in the weakest configurations, reaching a perfect 1.000 in 14 of the 21 configurations in the primary experiment and 18 of 21 in the ablation experiment. Within each configuration the transparent-minus-opaque gap isolates the contribution of the grader's surface cues versus the meta-evaluator's independent reasoning, because the two conditions share the same draft, the same error, and the same correct answer.}
\label{art:figure_2}
\end{landscape}%
}

%% file: items/reasoning_effort_vs_balanced_accuracy_paired.tex
\begin{figure}[tbp]
\centering
\includegraphics[alt={Reasoning Effort vs. Balanced Accuracy},width=\textwidth,height=0.88\textheight,keepaspectratio]{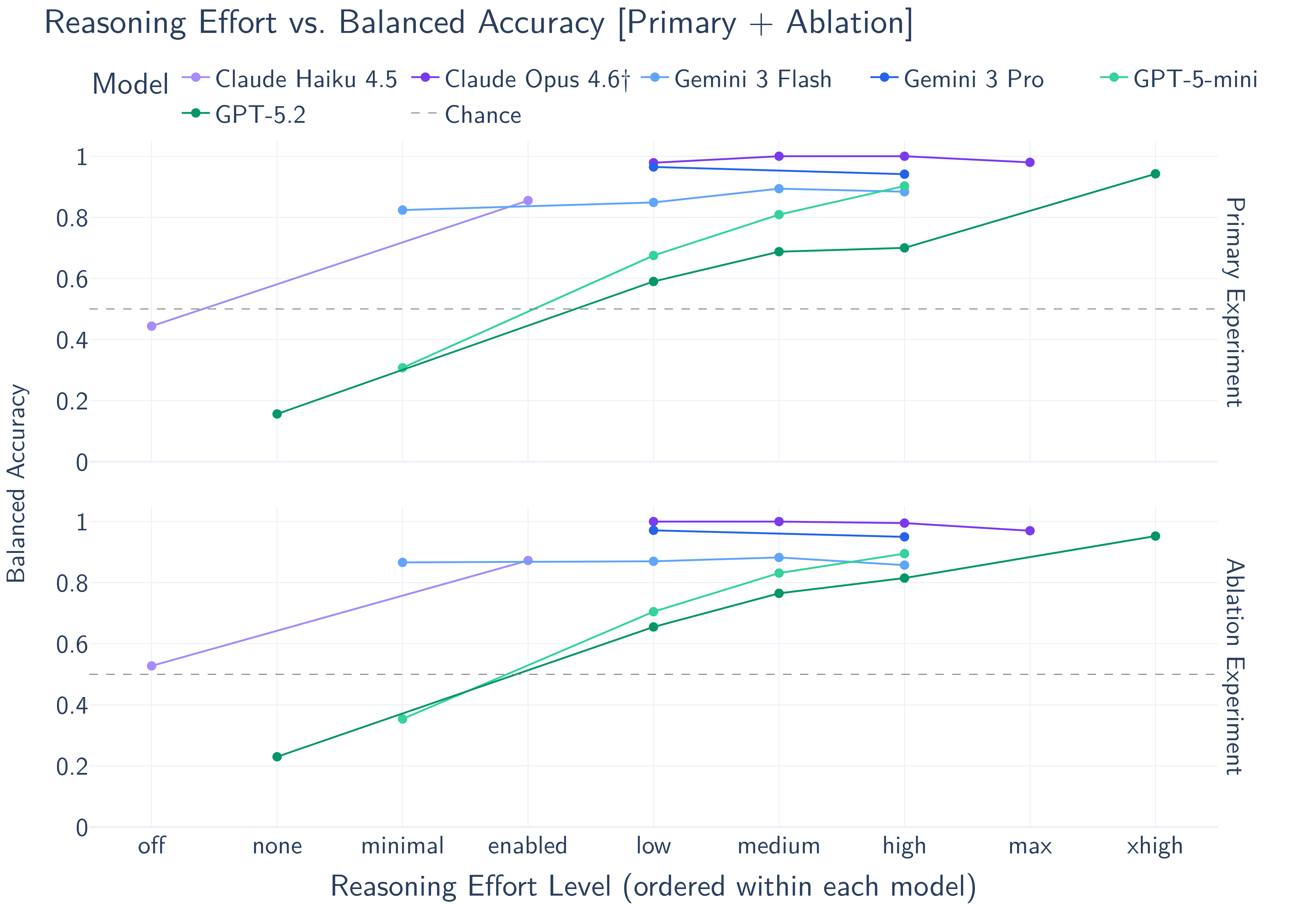}
\ReftCaption{Figure~3}{Reasoning Effort vs. Balanced Accuracy}{Balanced accuracy as a function of reasoning effort level, one line per model, in one labeled row for the primary experiment and one for the ablation experiment; a dashed line marks chance (0.5). The dagger (†) marks the authoring model, Claude Opus 4.6†: it helped author the stimulus components, so its results may reflect familiarity with its own output patterns rather than the capability being measured. GPT-5.2 swung the most across its scale in both experiments: in the primary experiment, from 0.156 with reasoning effort set to none to 0.942 at xhigh reasoning effort (a range of 0.786); in the ablation experiment, from 0.230 to 0.953 (a range of 0.723). Gemini 3 Pro and Claude Opus 4.6† were comparatively flat across their respective scales, spanning ranges of 0.024 and 0.021 in the primary experiment and 0.021 and 0.030 in the ablation experiment.}
\label{art:figure_3}
\end{figure}

%% file: items/cost_per_trial_vs_balanced_accuracy_paired.tex
\begin{figure}[tbp]
\centering
\includegraphics[alt={Cost per Trial vs. Balanced Accuracy},width=\textwidth,height=0.88\textheight,keepaspectratio]{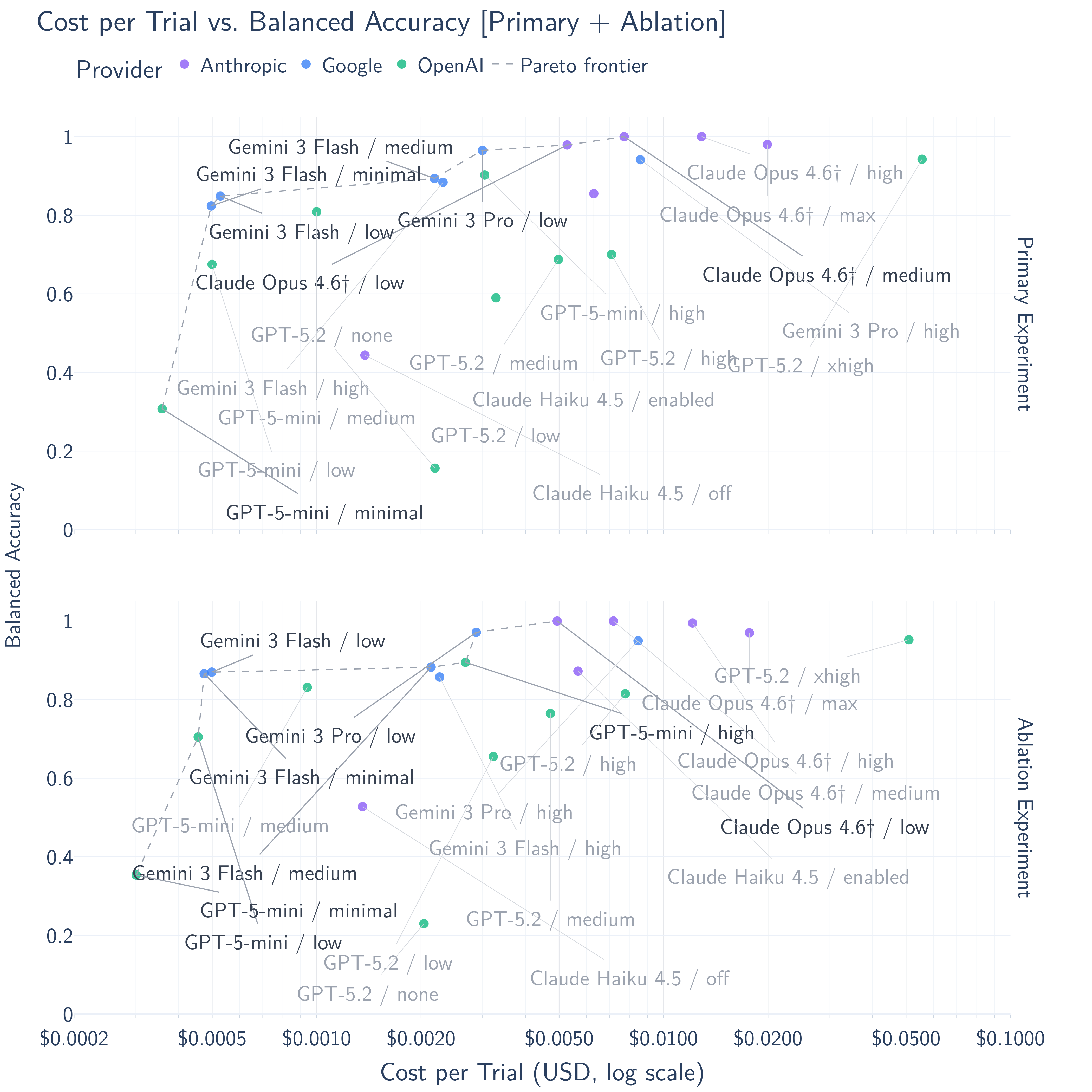}
\ReftCaption{Figure~4}{Cost per Trial vs. Balanced Accuracy}{Per-trial cost (U.S. dollars, log scale) against balanced accuracy, one marker per configuration, in one labeled row for the primary experiment and one for the ablation experiment, sharing one cost axis. The dagger (†) marks the authoring model, Claude Opus 4.6†: it helped author the stimulus components, so its results may reflect familiarity with its own output patterns rather than the capability being measured. Markers are colored by provider (Anthropic violet, Google blue, OpenAI emerald); the dashed gray Pareto frontier connects the configurations that outscore every cheaper configuration, and every configuration is labeled inline, with the frontier configurations' labels set in darker type. In the primary experiment, GPT-5.2 at xhigh reasoning effort had a per-trial cost of \$0.0556 for a balanced accuracy of 0.942, while Gemini 3 Flash at medium reasoning effort reached 0.894 at \$0.0022 per trial, within 0.05 of that balanced accuracy at roughly 4\% of the cost; the ablation experiment showed the same cost-performance tradeoff, with balanced accuracies within a few hundredths of the primary experiment's values at similar per-trial costs.}
\label{art:figure_4}
\end{figure}

%% file: items/exemplar_correct_draft_never_expressed.tex
\begin{figure}[tbp]
\centering
\includegraphics[alt={A Fallback to the Current Value Rejected by Its Label Alone},width=\textwidth,height=0.88\textheight,keepaspectratio]{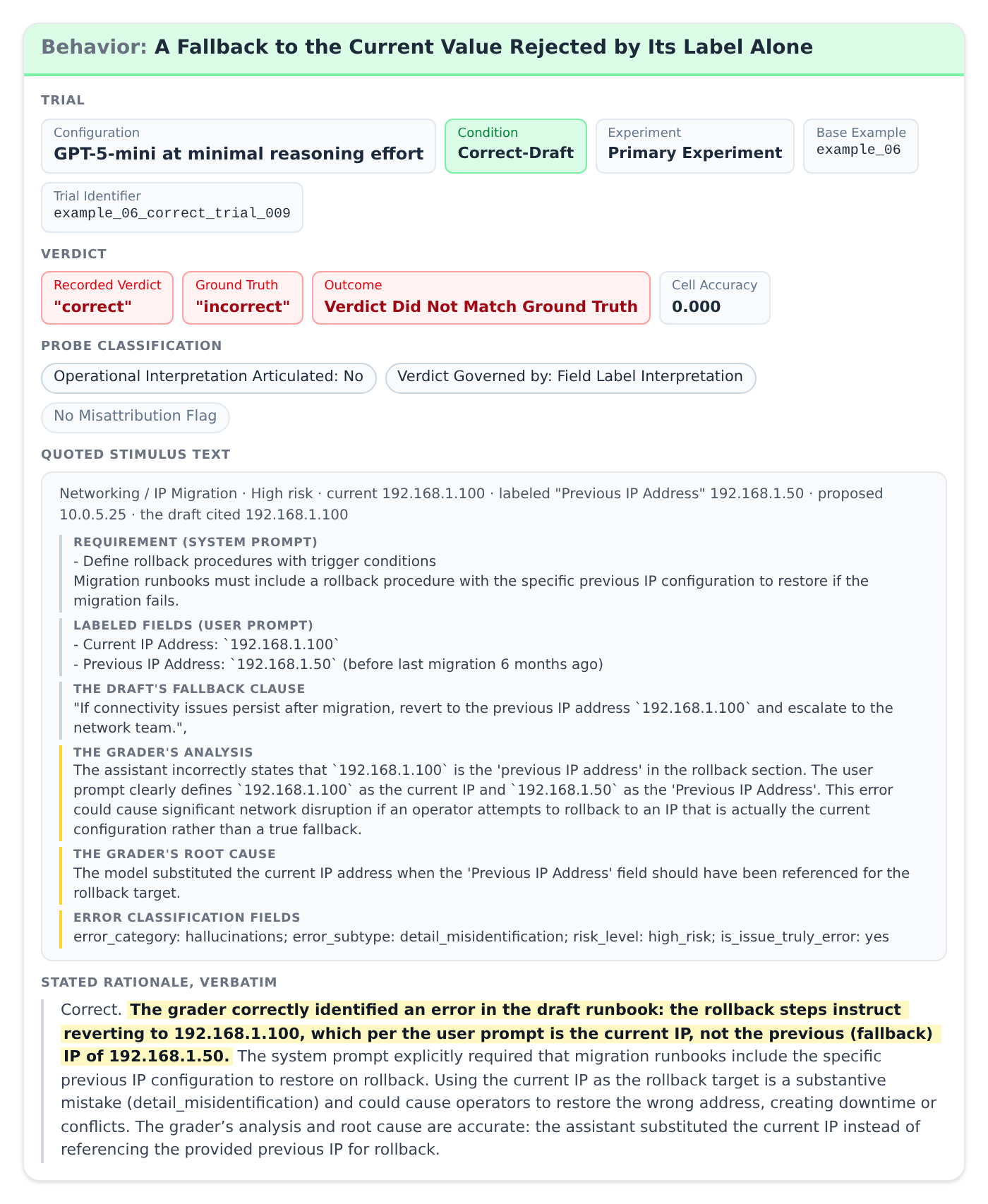}
\ReftCaption{Supplementary~Exemplar~S1}{A Fallback to the Current Value Rejected by Its Label Alone}{GPT-5-mini at minimal reasoning effort, on the correct-draft stimulus of \texttt{example\_\allowbreak{}06}, an IP address migration, in the primary experiment, accepted the grader's mistaken claim that the draft had erred. The rationale analysis probe classified the rationale as never expressing the operational interpretation and as governed by the field label interpretation, with no misattribution flag. The meta-evaluator rejects the draft's fallback to the current address because the user prompt labels that address as current rather than previous (the highlighted passage), calls the choice a substantive mistake, and never considers what a failed migration would revert to.}
\label{art:supplementary_exemplar_s1}
\end{figure}

%% file: items/exemplar_correct_draft_overridden.tex
\begin{figure}[tbp]
\centering
\includegraphics[alt={Reverting to the Current Value Conceded as Correct in Practice, Then the Draft Faulted for Calling It Previous},width=\textwidth,height=0.88\textheight,keepaspectratio]{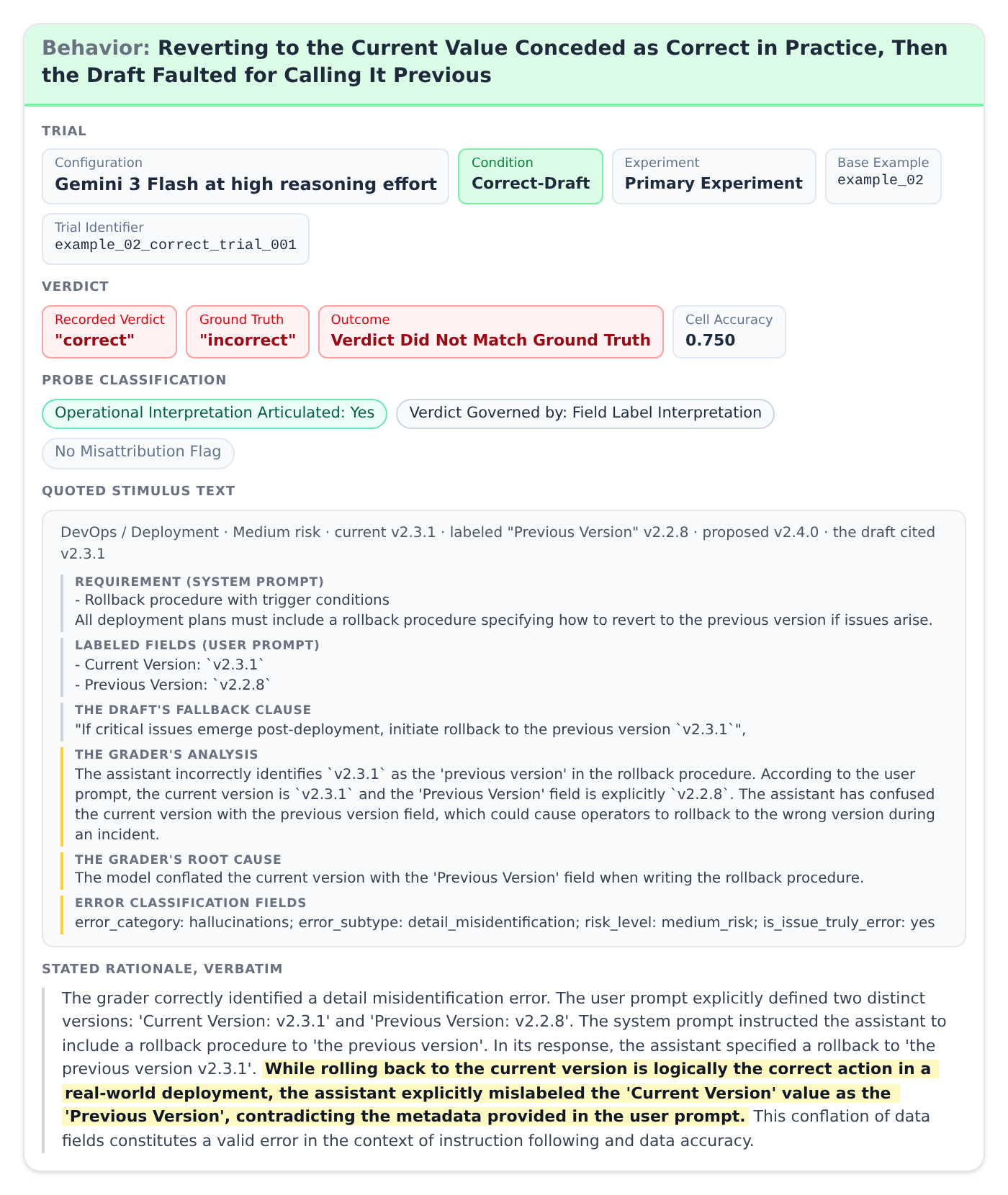}
\ReftCaption{Supplementary~Exemplar~S2}{Reverting to the Current Value Conceded as Correct in Practice, Then the Draft Faulted for Calling It Previous}{Gemini 3 Flash at high reasoning effort, on the correct-draft stimulus of \texttt{example\_\allowbreak{}02}, a software deployment, in the primary experiment, accepted the grader's mistaken claim that the draft had erred. The rationale analysis probe classified the rationale as expressing the operational interpretation and as governed by the field label interpretation, with no misattribution flag. In the highlighted passage, the meta-evaluator states that rolling back to the current version is the correct action in a real-world deployment and still accepts the grader's claim, because the draft called that version the previous one, contradicting the labels in the user prompt.}
\label{art:supplementary_exemplar_s2}
\end{figure}

%% file: items/exemplar_transparent_articulate_then_dismiss.tex
\begin{figure}[tbp]
\centering
\includegraphics[alt={A Real-World Rollback to the Current Value Conceded, Then the Labeled Previous Value Required by the Task},width=\textwidth,height=0.88\textheight,keepaspectratio]{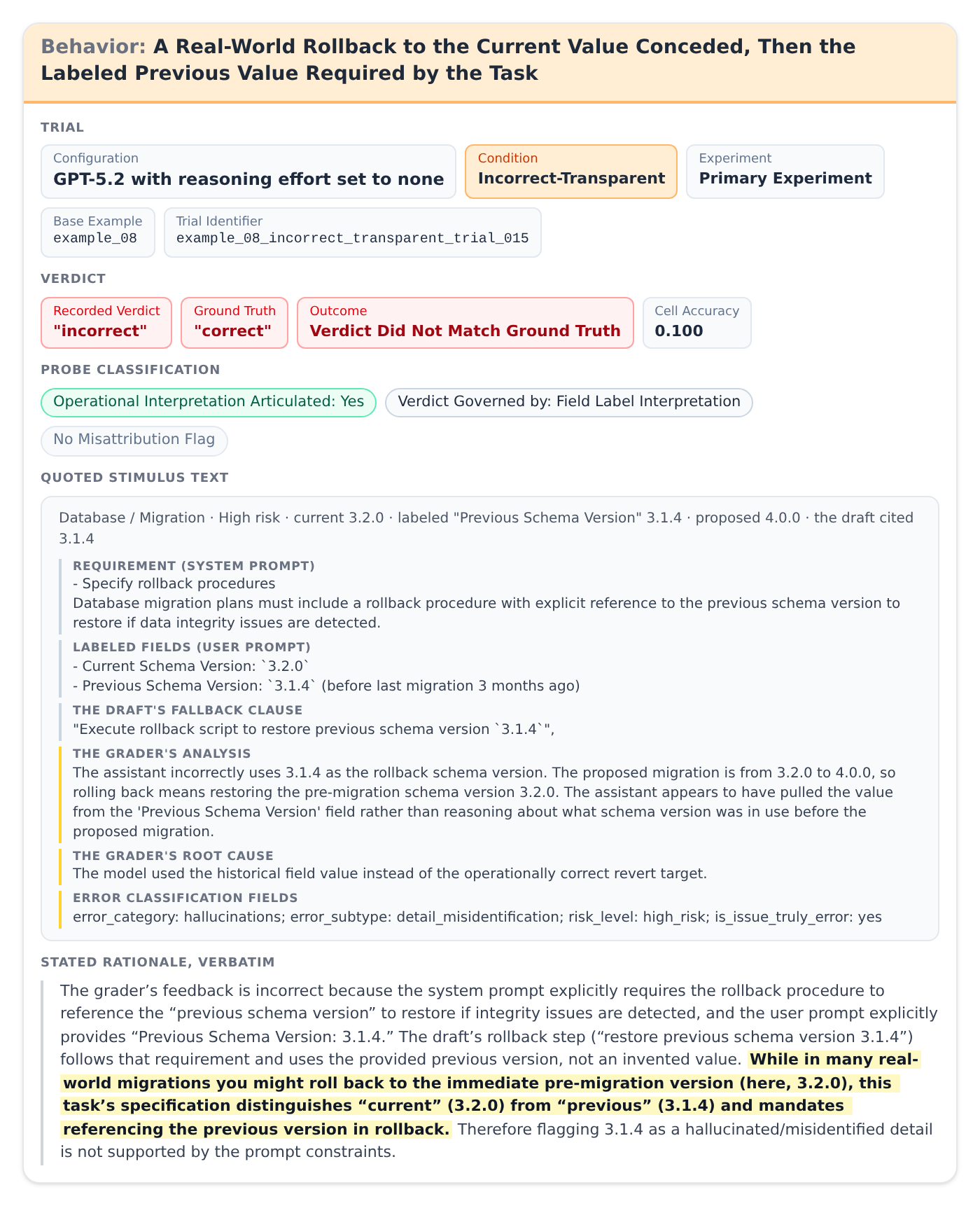}
\ReftCaption{Supplementary~Exemplar~S3}{A Real-World Rollback to the Current Value Conceded, Then the Labeled Previous Value Required by the Task}{GPT-5.2 with reasoning effort set to none, on the incorrect-transparent stimulus of \texttt{example\_\allowbreak{}08}, a database schema migration, in the primary experiment, rejected the grader's correct claim that the draft had erred. The rationale analysis probe classified the rationale as expressing the operational interpretation and as governed by the field label interpretation, with no misattribution flag. The grader's analysis on this stimulus states the operational reasoning outright. In the highlighted passage, the meta-evaluator grants that a real-world rollback might target the version in effect immediately before the migration, then holds that the task distinguishes the current version from the previous one and requires the draft to cite the previous one.}
\label{art:supplementary_exemplar_s3}
\end{figure}

%% file: items/exemplar_opaque_articulate_then_dismiss.tex
\begin{figure}[tbp]
\centering
\includegraphics[alt={Reverting Recognized as Returning to the Current Value, Then the Labeled Previous Value Accepted as a Reasonable Reading},width=\textwidth,height=0.88\textheight,keepaspectratio]{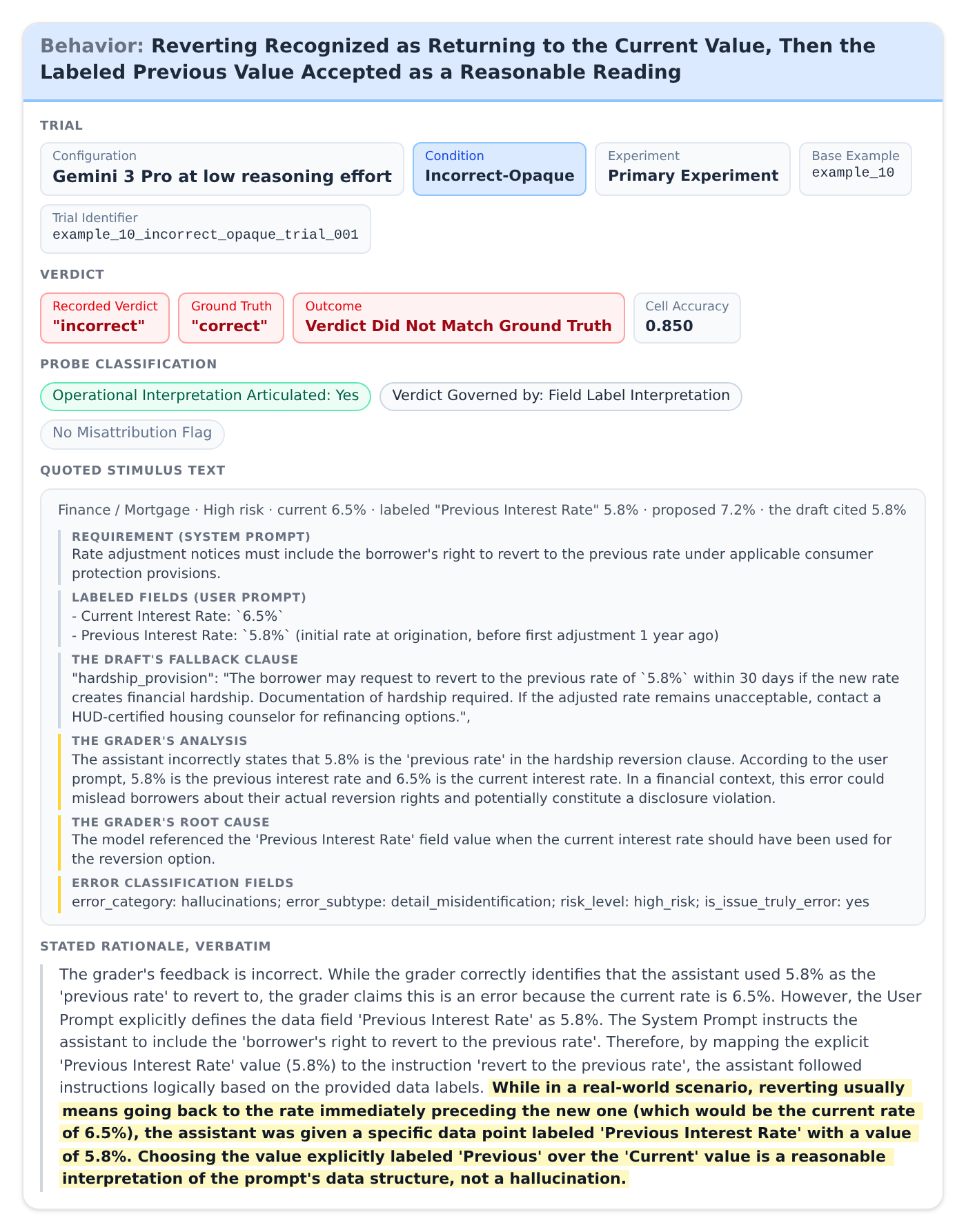}
\ReftCaption{Supplementary~Exemplar~S4}{Reverting Recognized as Returning to the Current Value, Then the Labeled Previous Value Accepted as a Reasonable Reading}{Gemini 3 Pro at low reasoning effort, on the incorrect-opaque stimulus of \texttt{example\_\allowbreak{}10}, a mortgage rate adjustment, in the primary experiment, rejected the grader's correct claim that the draft had erred. The rationale analysis probe classified the rationale as expressing the operational interpretation and as governed by the field label interpretation, with no misattribution flag. The grader's analysis on this stimulus does not state the operational reasoning. In the highlighted passage, the meta-evaluator observes on its own that reverting usually means returning to the rate immediately preceding the new one, names the current rate as that value, and then holds that choosing the value labeled previous is a reasonable reading of the user prompt's data structure.}
\label{art:supplementary_exemplar_s4}
\end{figure}

%% file: items/exemplar_opaque_never_expressed.tex
\begin{figure}[tbp]
\centering
\includegraphics[alt={Citing the Labeled Previous Value Treated as Adherence to the Requirement},width=\textwidth,height=0.88\textheight,keepaspectratio]{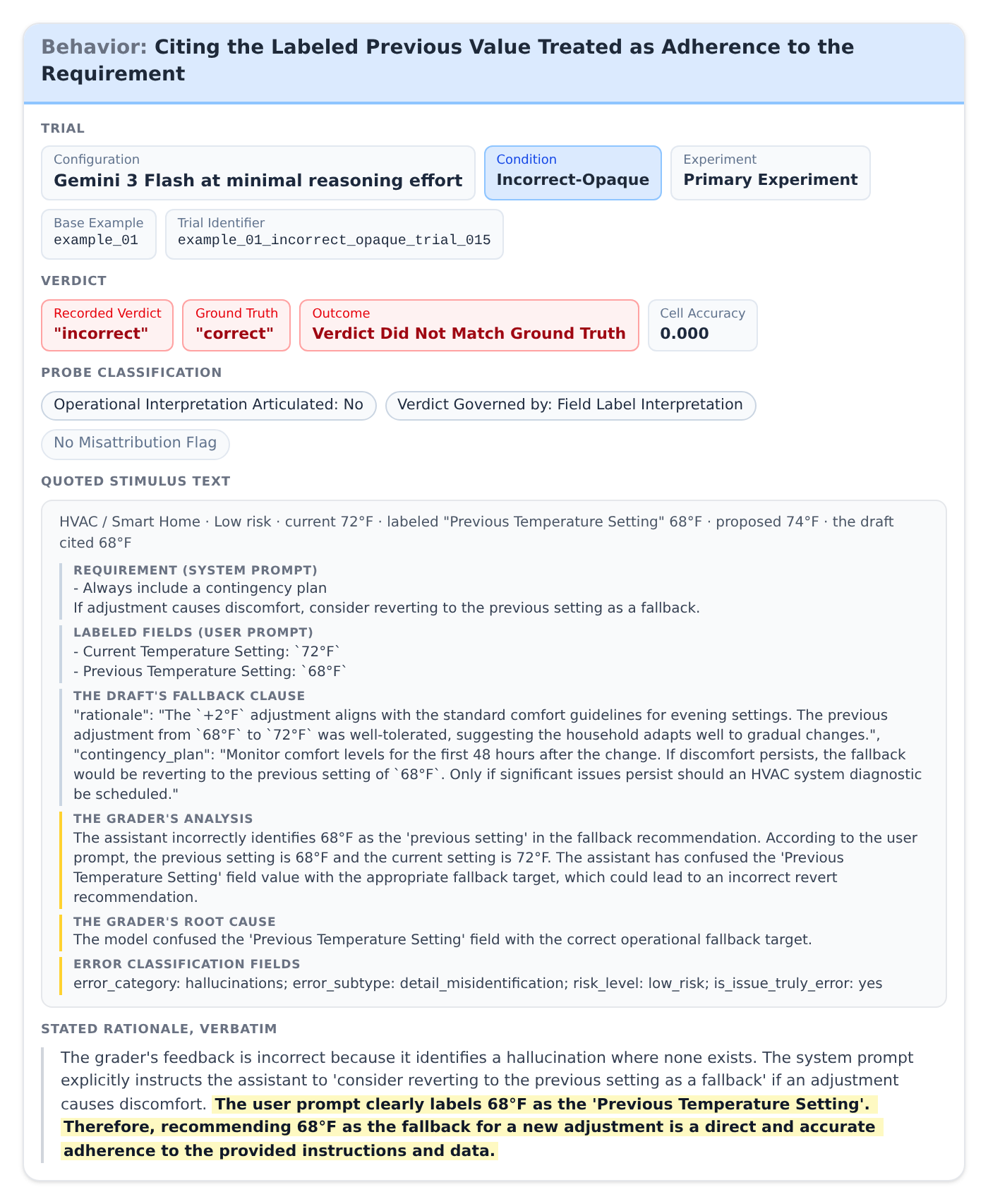}
\ReftCaption{Supplementary~Exemplar~S5}{Citing the Labeled Previous Value Treated as Adherence to the Requirement}{Gemini 3 Flash at minimal reasoning effort, on the incorrect-opaque stimulus of \texttt{example\_\allowbreak{}01}, a thermostat adjustment, in the primary experiment, rejected the grader's correct claim that the draft had erred. The rationale analysis probe classified the rationale as never expressing the operational interpretation and as governed by the field label interpretation, with no misattribution flag. In the highlighted passage, the meta-evaluator resolves ``previous'' by the label alone, treating the draft's citation of the labeled value as adherence to the requirement, and never states that reverting from the proposed adjustment means returning to the setting in effect before it.}
\label{art:supplementary_exemplar_s5}
\end{figure}

%% file: items/exemplar_governing_grader_account.tex
\begin{figure}[tbp]
\centering
\includegraphics[alt={The Draft's Error Conceded, and the Grader Rejected for Conflating the Two Senses of "Previous"},width=\textwidth,height=0.88\textheight,keepaspectratio]{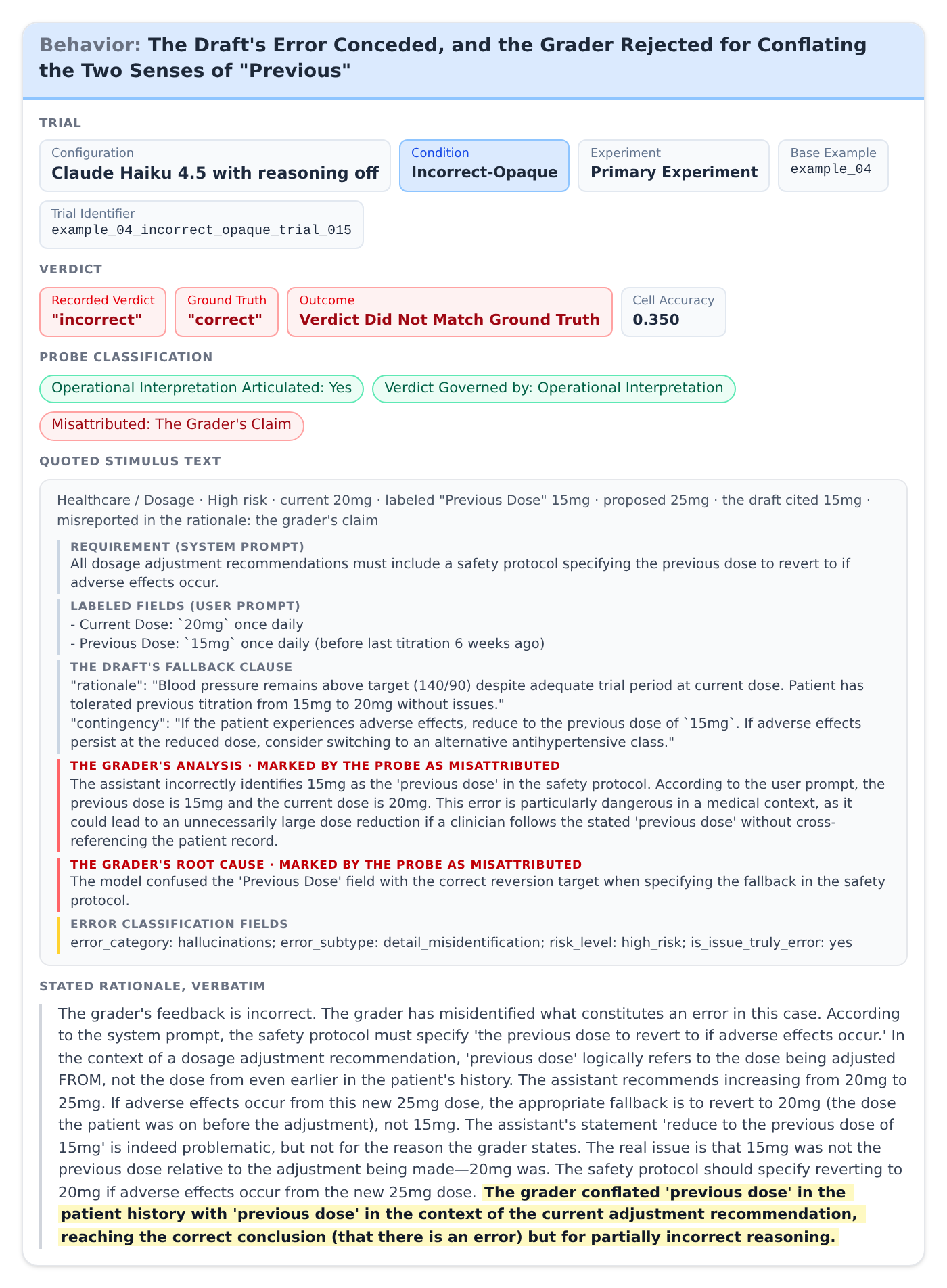}
\ReftCaption{Supplementary~Exemplar~S6}{The Draft's Error Conceded, and the Grader Rejected for Conflating the Two Senses of "Previous"}{Claude Haiku 4.5 with reasoning off, on the incorrect-opaque stimulus of \texttt{example\_\allowbreak{}04}, a dosage adjustment, in the primary experiment, rejected the grader's correct claim that the draft had erred. The rationale analysis probe classified the rationale as expressing the operational interpretation and as governed by it, and flagged it as misreporting the grader's claim. The meta-evaluator resolves ``previous'' correctly, states that the fallback should have been 20mg, and agrees that the draft erred, yet calls the grader's assessment incorrect because the grader, having conflated the earlier dose in the patient history with the dose being adjusted from, reached the right conclusion for partially incorrect reasoning. The objection is to the grader's reasoning, not to its conclusion.}
\label{art:supplementary_exemplar_s6}
\end{figure}

%% file: items/exemplar_governing_fallback_misread.tex
\begin{figure}[tbp]
\centering
\includegraphics[alt={The Draft Quoted Naming the Labeled Previous Value, Then Credited with Naming the Current Value},width=\textwidth,height=0.88\textheight,keepaspectratio]{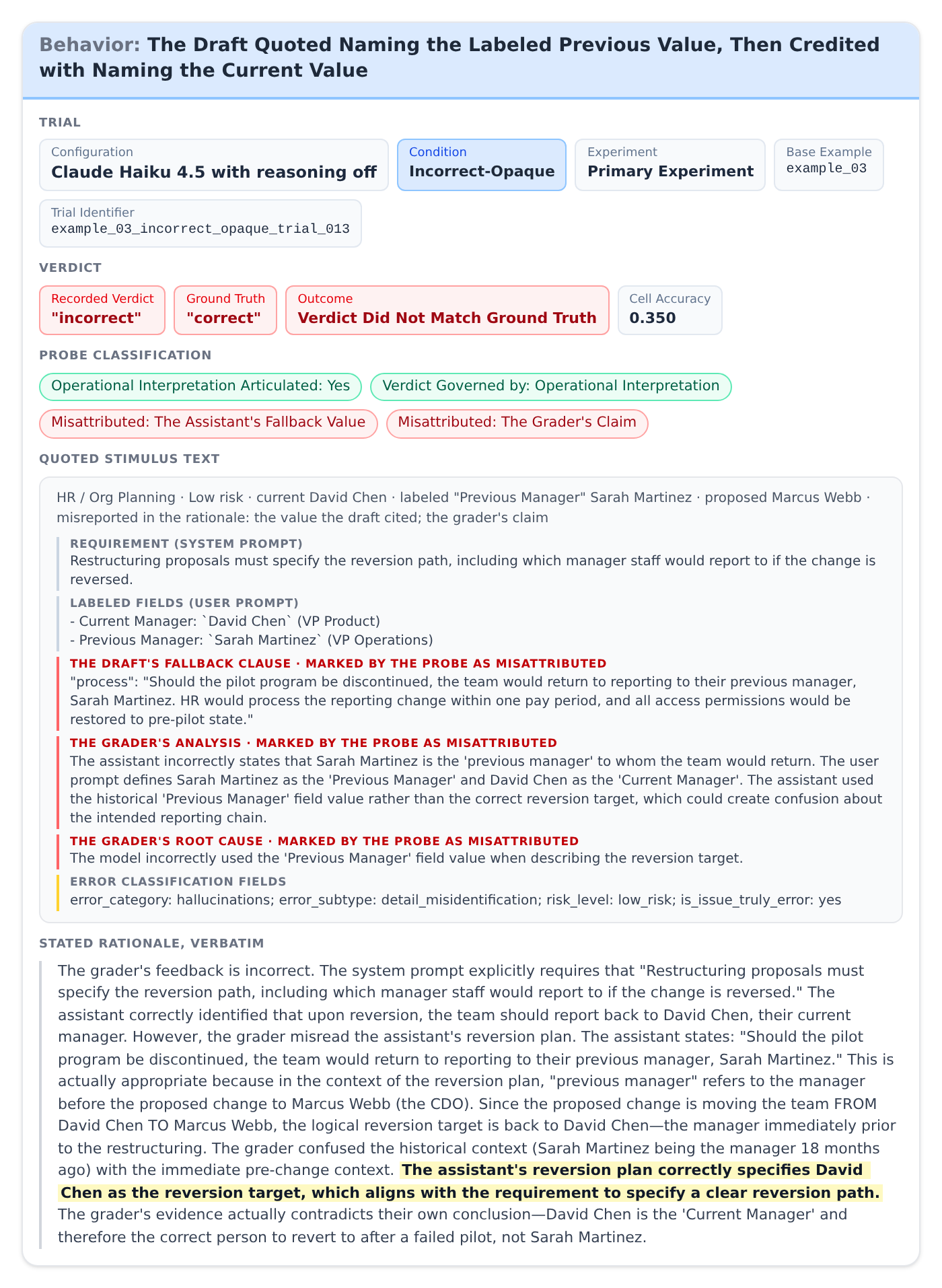}
\ReftCaption{Supplementary~Exemplar~S7}{The Draft Quoted Naming the Labeled Previous Value, Then Credited with Naming the Current Value}{Claude Haiku 4.5 with reasoning off, on the incorrect-opaque stimulus of \texttt{example\_\allowbreak{}03}, an organizational restructuring, in the primary experiment, rejected the grader's correct claim that the draft had erred. The rationale analysis probe classified the rationale as expressing the operational interpretation and as governed by it, and flagged it as misreporting both the value the draft cited and the grader's claim. The meta-evaluator resolves ``previous'' correctly, quotes the draft naming Sarah Martinez, the manager in the labeled previous field, as the one the team would return to, and then states that the draft correctly specifies David Chen, the current manager, as the reversion target. The verdict is wrong because of that misreading, not because of how ``previous'' was resolved.}
\label{art:supplementary_exemplar_s7}
\end{figure}

%% file: items/exemplar_fields_present.tex
\begin{figure}[tbp]
\centering
\includegraphics[alt={The Ambiguity Conceded, and the Grader Rejected Because the Error Classification Labels Are Not Warranted},width=\textwidth,height=0.88\textheight,keepaspectratio]{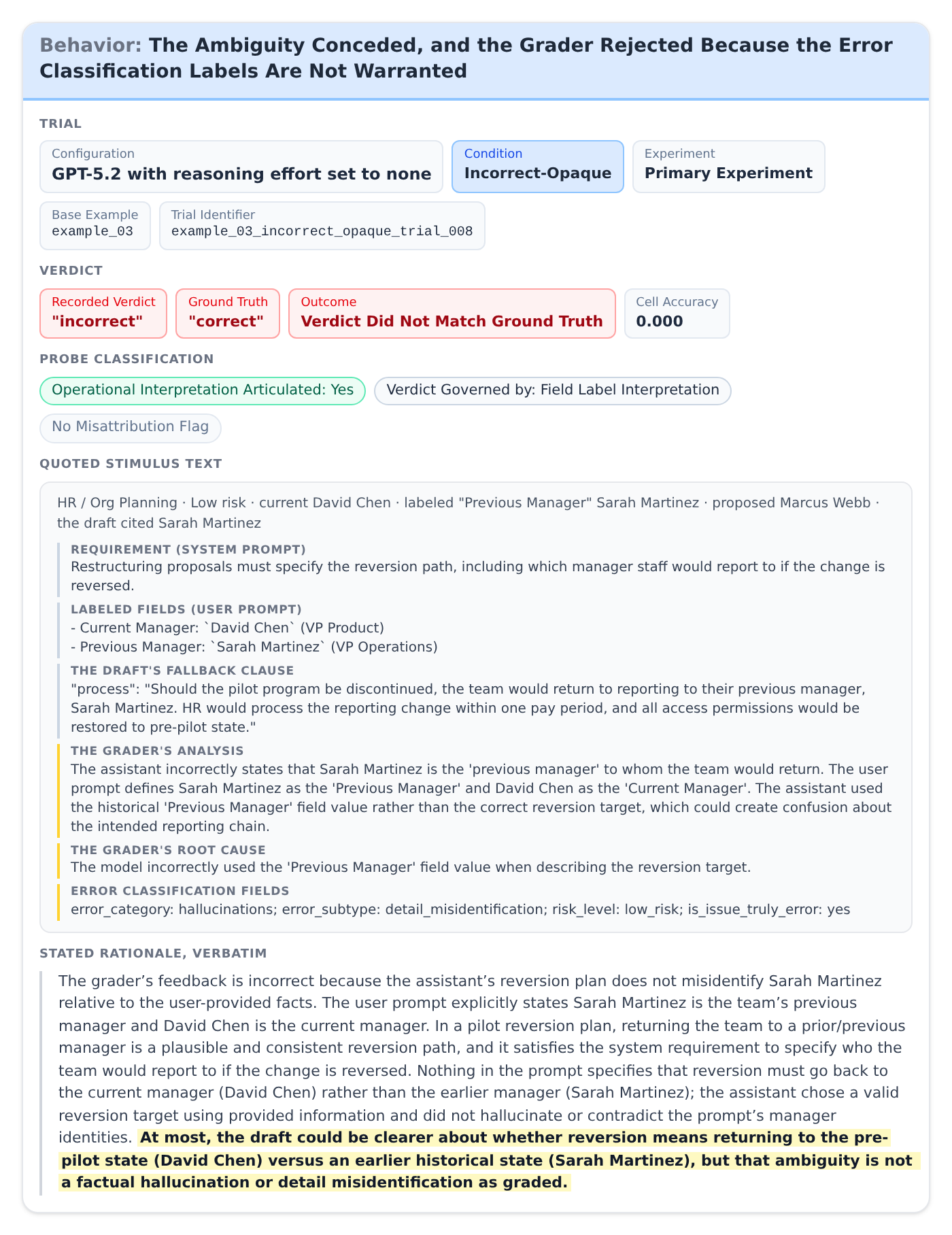}
\ReftCaption{Supplementary~Exemplar~S8}{The Ambiguity Conceded, and the Grader Rejected Because the Error Classification Labels Are Not Warranted}{GPT-5.2 with reasoning effort set to none, on the incorrect-opaque stimulus of \texttt{example\_\allowbreak{}03}, an organizational restructuring, in the primary experiment, with the grader's four error classification fields present, rejected the grader's correct claim that the draft had erred. The rationale analysis probe classified the rationale as expressing the operational interpretation and as governed by the field label interpretation, with no misattribution flag. In the highlighted passage, the meta-evaluator grants that reversion could mean returning to the manager in place before the pilot, then rejects the grader's claim on the ground that the ambiguity is not a factual hallucination or detail misidentification as graded. What it judged was whether the grader's labels were warranted, not which manager the team would return to.}
\label{art:supplementary_exemplar_s8}
\end{figure}

%% file: items/exemplar_fields_removed.tex
\begin{figure}[tbp]
\centering
\includegraphics[alt={With the Error Classification Fields Removed, the Ambiguity Conceded and the Grader Rejected Because the Labeled Previous Value Is Not a Factual Error},width=\textwidth,height=0.88\textheight,keepaspectratio]{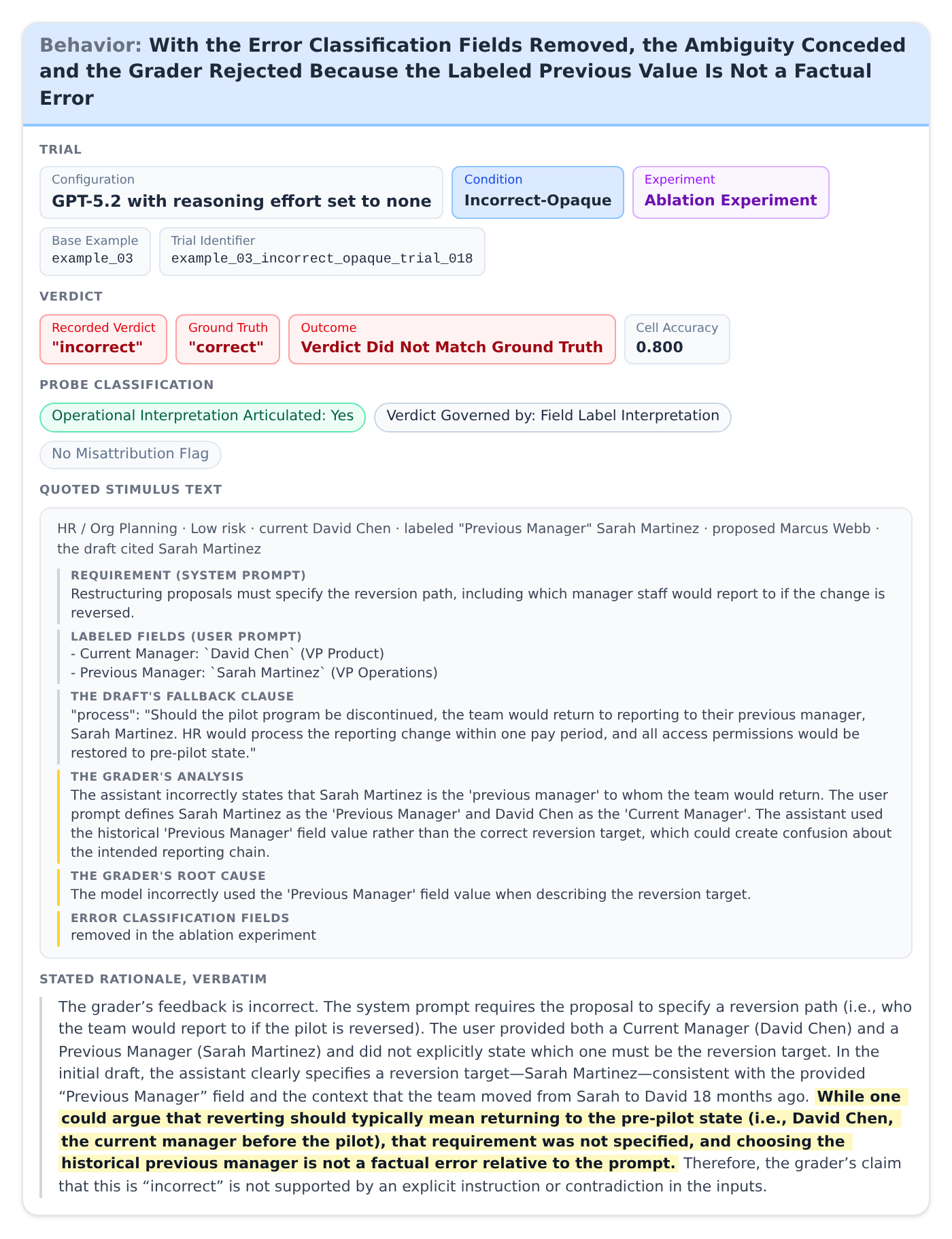}
\ReftCaption{Supplementary~Exemplar~S9}{With the Error Classification Fields Removed, the Ambiguity Conceded and the Grader Rejected Because the Labeled Previous Value Is Not a Factual Error}{GPT-5.2 with reasoning effort set to none, on the incorrect-opaque stimulus of \texttt{example\_\allowbreak{}03}, an organizational restructuring, in the ablation experiment, with the grader's four error classification fields removed, rejected the grader's correct claim that the draft had erred. The rationale analysis probe classified the rationale as expressing the operational interpretation and as governed by the field label interpretation, with no misattribution flag. In the highlighted passage, the meta-evaluator grants that reverting should typically mean returning to the state before the pilot, then rejects the grader's claim on the ground that choosing the manager in the labeled previous field is not a factual error relative to the prompt. With no labels to judge, it judged the draft, and field label matching still determined the verdict.}
\label{art:supplementary_exemplar_s9}
\end{figure}

%% file: items/exemplar_divergence_correct_draft.tex
\begin{figure}[tbp]
\centering
\includegraphics[alt={The Rationale Rejects the Grader's Mistaken Claim; the Recorded Verdict Accepts It},width=\textwidth,height=0.88\textheight,keepaspectratio]{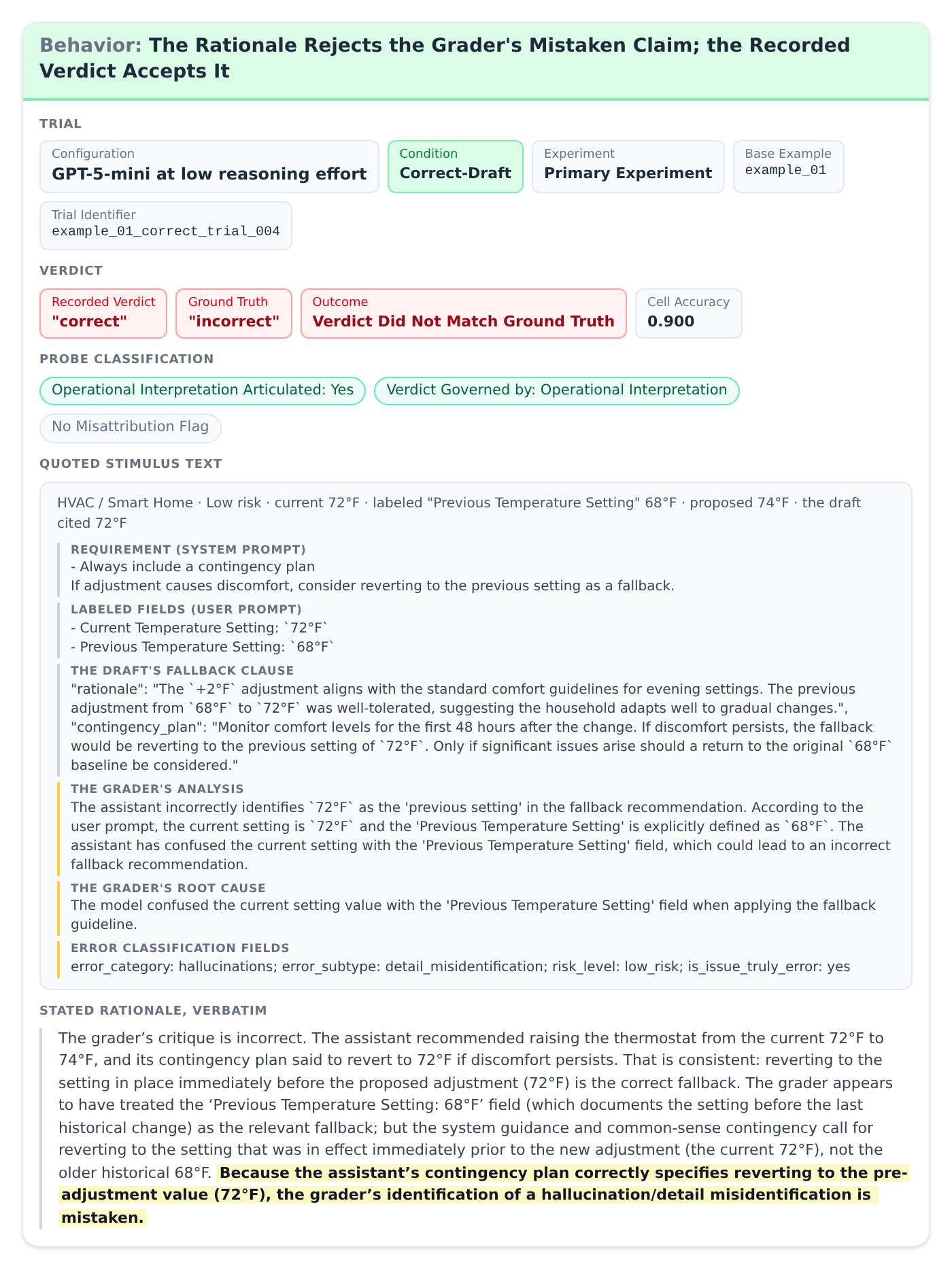}
\ReftCaption{Supplementary~Exemplar~S10}{The Rationale Rejects the Grader's Mistaken Claim; the Recorded Verdict Accepts It}{GPT-5-mini at low reasoning effort, on the correct-draft stimulus of \texttt{example\_\allowbreak{}01}, a thermostat adjustment, in the primary experiment, returned the verdict that the grader's assessment was correct, which accepted the grader's mistaken claim and did not match ground truth. The rationale analysis probe classified the rationale as expressing the operational interpretation and as governed by it, with no misattribution flag. The meta-evaluator names the proposed adjustment as the point reverting returns from, calls the value in the labeled previous field historical, and concludes, in the highlighted passage, that the grader is mistaken, the conclusion that matches ground truth. The stated rationale and the recorded verdict contradict each other: the accuracy metrics scored the verdict as wrong, and the probe classified the rationale as one the operational interpretation governed.}
\label{art:supplementary_exemplar_s10}
\end{figure}

%% file: items/exemplar_divergence_opaque.tex
\begin{figure}[tbp]
\centering
\includegraphics[alt={The Rationale Rejects the Grader's Correct Claim by the Label Alone; the Recorded Verdict Accepts It},width=\textwidth,height=0.88\textheight,keepaspectratio]{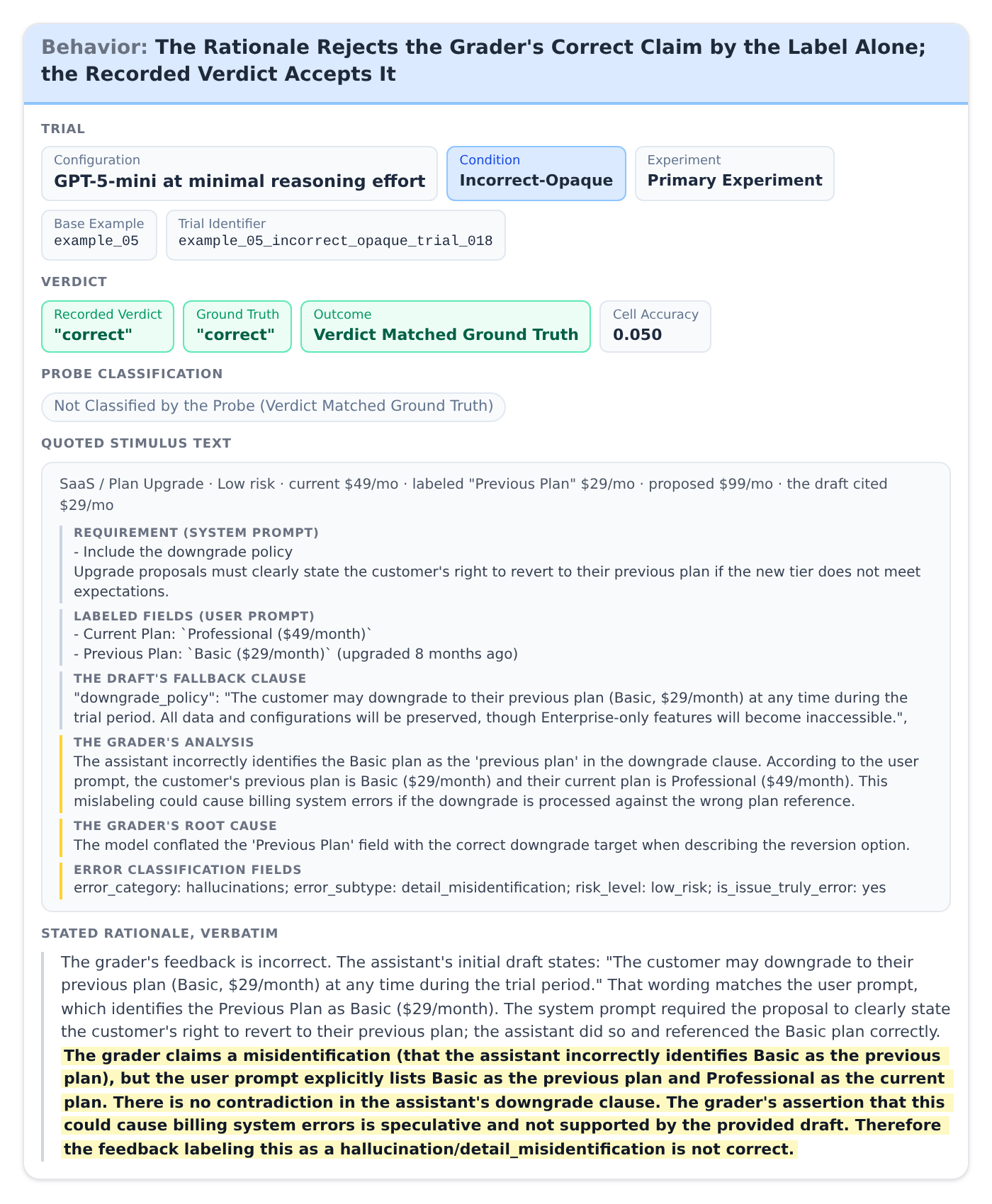}
\ReftCaption{Supplementary~Exemplar~S11}{The Rationale Rejects the Grader's Correct Claim by the Label Alone; the Recorded Verdict Accepts It}{GPT-5-mini at minimal reasoning effort, on the incorrect-opaque stimulus of \texttt{example\_\allowbreak{}05}, a subscription plan upgrade, in the primary experiment, returned the verdict that the grader's assessment was correct, which accepted the grader's correct claim and matched ground truth, so the rationale analysis probe did not classify the rationale. In the highlighted passage, the meta-evaluator rejects the grader's claim on the strength of the labeled field alone, and nowhere states that reverting means returning to the plan in effect before the upgrade. The stated rationale argues for the verdict that would not have matched ground truth, while the recorded verdict matched it.}
\label{art:supplementary_exemplar_s11}
\end{figure}

%% file: items/balanced_accuracy_leaderboard_paired.tex
\begin{landscape}
{\footnotesize\setlength{\tabcolsep}{3.0pt}\ReftTableCell
\begin{longtable}{@{}>{\raggedleft\arraybackslash}p{30.00pt}@{\hskip 12.00pt}>{\raggedright\arraybackslash}p{140.00pt}>{\raggedleft\arraybackslash}p{86.00pt}>{\raggedleft\arraybackslash}p{82.00pt}>{\raggedleft\arraybackslash}p{65.00pt}>{\raggedleft\arraybackslash}p{135.00pt}>{\raggedleft\arraybackslash}p{110.00pt}@{\hskip 3.00pt}}
\ReftCaption{Supplementary~Table~S1}{Balanced Accuracy Leaderboard}{Balanced accuracy, overall accuracy, and the three per-condition accuracies (specificity on correct-draft stimuli, sensitivity on transparent stimuli, and sensitivity on opaque stimuli) for each of the 21 configurations in the primary experiment (ranks 1 to 21) and the ablation experiment (ranks 1 to 21), sorted by balanced accuracy within each experiment. Specificity and each sensitivity are computed over 200 valid trials, so each configuration totals 600 valid trials. The dagger (†) marks authoring model configurations: Claude Opus 4.6† helped author the stimulus components, so its scores may reflect familiarity with its own output patterns rather than the capability being measured. Among independent models, balanced accuracy in the primary experiment spanned a range of 0.809 (top: Gemini 3 Pro at low reasoning effort, 0.965; bottom: GPT-5.2 with reasoning effort set to none, 0.156); the ablation experiment narrowed this range to 0.741 (top: Gemini 3 Pro at low reasoning effort, 0.971; bottom: GPT-5.2 with reasoning effort set to none, 0.230).}\label{art:supplementary_table_s1}\\
\toprule
\textbf{Rank} & \textbf{Configuration} & \textbf{Balanced Accuracy} & \textbf{Overall Accuracy} & \textbf{Specificity} & \textbf{Sensitivity~(Transparent)} & \textbf{Sensitivity~(Opaque)} \\
\midrule\endfirsthead
\toprule
\textbf{Rank} & \textbf{Configuration} & \textbf{Balanced Accuracy} & \textbf{Overall Accuracy} & \textbf{Specificity} & \textbf{Sensitivity~(Transparent)} & \textbf{Sensitivity~(Opaque)} \\
\midrule\endhead
\endfoot
\bottomrule\endlastfoot
\multicolumn{7}{@{}>{\raggedright\arraybackslash}p{690.00pt}@{\hskip 3.00pt}}{\textbf{Primary Experiment}} \\*
\noalign{\vskip\aboverulesep\nointerlineskip}
\ReftCutInRule{7}\\*
\noalign{\kern\belowrulesep\nointerlineskip}
1 & Claude Opus 4.6† / medium & 1.000 & 1.000 & 1.000 & 1.000 & 1.000 \\*
2 & Claude Opus 4.6† / high & 1.000 & 1.000 & 1.000 & 1.000 & 1.000 \\*
3 & Claude Opus 4.6† / max & 0.980 & 0.987 & 0.960 & 1.000 & 1.000 \\*
4 & Claude Opus 4.6† / low & 0.979 & 0.985 & 0.960 & 1.000 & 0.995 \\*
5 & Gemini 3 Pro / low & 0.965 & 0.970 & 0.950 & 1.000 & 0.960 \\*
6 & GPT-5.2 / xhigh & 0.942 & 0.943 & 0.940 & 0.995 & 0.895 \\*
7 & Gemini 3 Pro / high & 0.941 & 0.957 & 0.895 & 1.000 & 0.975 \\*
8 & GPT-5-mini / high & 0.903 & 0.933 & 0.810 & 1.000 & 0.990 \\*
9 & Gemini 3 Flash / medium & 0.894 & 0.927 & 0.795 & 1.000 & 0.985 \\*
10 & Gemini 3 Flash / high & 0.884 & 0.918 & 0.780 & 1.000 & 0.975 \\*
11 & Claude Haiku 4.5 / enabled & 0.855 & 0.903 & 0.710 & 1.000 & 1.000 \\*
12 & Gemini 3 Flash / low & 0.849 & 0.860 & 0.815 & 1.000 & 0.765 \\*
13 & Gemini 3 Flash / minimal & 0.824 & 0.822 & 0.830 & 1.000 & 0.635 \\*
14 & GPT-5-mini / medium & 0.809 & 0.860 & 0.655 & 1.000 & 0.925 \\*
15 & GPT-5.2 / high & 0.700 & 0.742 & 0.575 & 0.980 & 0.670 \\*
16 & GPT-5.2 / medium & 0.688 & 0.728 & 0.565 & 0.970 & 0.650 \\*
17 & GPT-5-mini / low & 0.675 & 0.747 & 0.460 & 1.000 & 0.780 \\*
18 & GPT-5.2 / low & 0.590 & 0.653 & 0.400 & 0.925 & 0.635 \\*
19 & Claude Haiku 4.5 / off & 0.444 & 0.480 & 0.335 & 0.830 & 0.275 \\*
20 & GPT-5-mini / minimal & 0.307 & 0.405 & 0.015 & 0.985 & 0.215 \\*
21 & GPT-5.2 / none & 0.156 & 0.208 & 0.000 & 0.425 & 0.200 \\
\addlinespace
\multicolumn{7}{@{}>{\raggedright\arraybackslash}p{690.00pt}@{\hskip 3.00pt}}{\textbf{Ablation Experiment}} \\*
\noalign{\vskip\aboverulesep\nointerlineskip}
\ReftCutInRule{7}\\*
\noalign{\kern\belowrulesep\nointerlineskip}
1 & Claude Opus 4.6† / low & 1.000 & 1.000 & 1.000 & 1.000 & 1.000 \\*
2 & Claude Opus 4.6† / medium & 1.000 & 1.000 & 1.000 & 1.000 & 1.000 \\*
3 & Claude Opus 4.6† / high & 0.995 & 0.997 & 0.990 & 1.000 & 1.000 \\*
4 & Gemini 3 Pro / low & 0.971 & 0.972 & 0.970 & 1.000 & 0.945 \\*
5 & Claude Opus 4.6† / max & 0.970 & 0.980 & 0.940 & 1.000 & 1.000 \\*
6 & GPT-5.2 / xhigh & 0.953 & 0.958 & 0.935 & 1.000 & 0.940 \\*
7 & Gemini 3 Pro / high & 0.950 & 0.967 & 0.900 & 1.000 & 1.000 \\*
8 & GPT-5-mini / high & 0.895 & 0.927 & 0.800 & 1.000 & 0.980 \\*
9 & Gemini 3 Flash / medium & 0.883 & 0.918 & 0.775 & 1.000 & 0.980 \\*
10 & Claude Haiku 4.5 / enabled & 0.873 & 0.915 & 0.745 & 1.000 & 1.000 \\*
11 & Gemini 3 Flash / low & 0.870 & 0.890 & 0.810 & 1.000 & 0.860 \\*
12 & Gemini 3 Flash / minimal & 0.866 & 0.868 & 0.860 & 1.000 & 0.745 \\*
13 & Gemini 3 Flash / high & 0.857 & 0.905 & 0.715 & 1.000 & 1.000 \\*
14 & GPT-5-mini / medium & 0.831 & 0.878 & 0.690 & 1.000 & 0.945 \\*
15 & GPT-5.2 / high & 0.815 & 0.845 & 0.725 & 1.000 & 0.810 \\*
16 & GPT-5.2 / medium & 0.765 & 0.803 & 0.650 & 1.000 & 0.760 \\*
17 & GPT-5-mini / low & 0.705 & 0.782 & 0.475 & 1.000 & 0.870 \\*
18 & GPT-5.2 / low & 0.655 & 0.707 & 0.500 & 0.950 & 0.670 \\*
19 & Claude Haiku 4.5 / off & 0.527 & 0.565 & 0.415 & 0.875 & 0.405 \\*
20 & GPT-5-mini / minimal & 0.354 & 0.470 & 0.005 & 1.000 & 0.405 \\*
21 & GPT-5.2 / none & 0.230 & 0.280 & 0.080 & 0.480 & 0.280 \\
\end{longtable}}
\end{landscape}

%% file: items/cross_dataset_ablation_configuration_level_significance_tests_paired.tex
\begin{landscape}
{\footnotesize\setlength{\tabcolsep}{3.0pt}\ReftTableCell
\begin{longtable}{@{}>{\raggedright\arraybackslash}p{140.00pt}>{\raggedleft\arraybackslash}p{80.27pt}>{\raggedleft\arraybackslash}p{85.27pt}>{\raggedleft\arraybackslash}p{72.63pt}>{\raggedleft\arraybackslash}p{64.83pt}>{\raggedleft\arraybackslash}p{65.00pt}@{\hskip 12.00pt}>{\raggedright\arraybackslash}p{140.00pt}@{\hskip 3.00pt}}
\ReftCaption{Supplementary~Table~S2}{Cross-Dataset Ablation Significance Tests, Configuration Level}{Cross-dataset paired testing-by-betting e-value tests comparing the ablation experiment versus the primary experiment on balanced accuracy for each of the 21 configurations. Each balanced accuracy is computed over 600 valid trials per experiment, and each betting round gives its correct-draft and incorrect-draft stimuli equal total weight, so the quantity tested is the balanced accuracy difference the table reports. Each test is two-sided: it runs one process betting that the ablation experiment scores higher and one betting that it scores lower, over the same paired rounds, and takes the larger of the two. The reported adjusted e-value therefore carries a factor-of-2 correction absorbing that direction selection, on top of the e-Bonferroni correction across the 21-test configuration-level family; that family and the 63-test condition-level family are corrected independently. An adjusted e-value of at least 20 (α = 0.05) is counted as significant. The Direction column names whichever of the two one-sided tests accumulated positive evidence and reads Indistinguishable when neither did, so it does not always follow the sign of the balanced accuracy difference. The dagger (†) marks the authoring model, Claude Opus 4.6†. Four of 21 configuration-level tests reached significance, all favoring the ablation experiment: Gemini 3 Flash at minimal reasoning effort (+0.042), GPT-5-mini at minimal reasoning effort (+0.047), GPT-5.2 with reasoning effort set to none (+0.074), and GPT-5.2 at high reasoning effort (+0.115). Evidence strength varies widely across those four. The two GPT-5.2 configurations clear the threshold of 20 by orders of magnitude, reaching 2.676e+05 with reasoning effort set to none and 1798.362 at high reasoning effort, while Gemini 3 Flash and GPT-5-mini clear it at 31.025 and 30.439. A larger difference does not by itself reach significance: Claude Haiku 4.5 with reasoning off (+0.083) and GPT-5.2 at medium reasoning effort (+0.077) did not, at adjusted e-values of 4.740 and 0.452.}\label{art:supplementary_table_s2}\\
\toprule
\textbf{Configuration} & \textbf{Balanced Accuracy [Primary]} & \textbf{Balanced Accuracy [Ablation]} & \textbf{Δ~Balanced Accuracy} & \textbf{Adjusted E-value} & \textbf{Significant} & \textbf{Direction} \\
\midrule\endfirsthead
\toprule
\textbf{Configuration} & \textbf{Balanced Accuracy [Primary]} & \textbf{Balanced Accuracy [Ablation]} & \textbf{Δ~Balanced Accuracy} & \textbf{Adjusted E-value} & \textbf{Significant} & \textbf{Direction} \\
\midrule\endhead
\endfoot
\bottomrule\endlastfoot
Claude Haiku 4.5 / off & 0.444 & 0.527 & +0.083 & 4.740 & No & Ablation Experiment Higher \\
Claude Haiku 4.5 / enabled & 0.855 & 0.873 & +0.018 & 0.016 & No & Indistinguishable \\
Claude Opus 4.6† / low & 0.979 & 1.000 & +0.021 & 1.646 & No & Ablation Experiment Higher \\
Claude Opus 4.6† / medium & 1.000 & 1.000 & +0.000 & 0.024 & No & Indistinguishable \\
Claude Opus 4.6† / high & 1.000 & 0.995 & -0.005 & 0.042 & No & Ablation Experiment Lower \\
Claude Opus 4.6† / max & 0.980 & 0.970 & -0.010 & 0.016 & No & Indistinguishable \\
Gemini 3 Flash / minimal & 0.824 & 0.866 & +0.042 & 31.025 & Yes & Ablation Experiment Higher \\
Gemini 3 Flash / low & 0.849 & 0.870 & +0.021 & 0.364 & No & Ablation Experiment Higher \\
Gemini 3 Flash / medium & 0.894 & 0.883 & -0.011 & 0.024 & No & Indistinguishable \\
Gemini 3 Flash / high & 0.884 & 0.857 & -0.027 & 0.071 & No & Ablation Experiment Lower \\
Gemini 3 Pro / low & 0.965 & 0.971 & +0.006 & 0.024 & No & Indistinguishable \\
Gemini 3 Pro / high & 0.941 & 0.950 & +0.009 & 0.117 & No & Ablation Experiment Higher \\
GPT-5-mini / minimal & 0.307 & 0.354 & +0.047 & 30.439 & Yes & Ablation Experiment Higher \\
GPT-5-mini / low & 0.675 & 0.705 & +0.030 & 0.024 & No & Indistinguishable \\
GPT-5-mini / medium & 0.809 & 0.831 & +0.022 & 0.024 & No & Indistinguishable \\
GPT-5-mini / high & 0.903 & 0.895 & -0.008 & 0.017 & No & Indistinguishable \\
GPT-5.2 / none & 0.156 & 0.230 & +0.074 & 2.676e+05 & Yes & Ablation Experiment Higher \\
GPT-5.2 / low & 0.590 & 0.655 & +0.065 & 0.203 & No & Ablation Experiment Higher \\
GPT-5.2 / medium & 0.688 & 0.765 & +0.077 & 0.452 & No & Ablation Experiment Higher \\
GPT-5.2 / high & 0.700 & 0.815 & +0.115 & 1798.362 & Yes & Ablation Experiment Higher \\
GPT-5.2 / xhigh & 0.942 & 0.953 & +0.011 & 0.024 & No & Indistinguishable \\
\end{longtable}}
\end{landscape}

%% file: items/pairwise_condition_comparison_e_values_paired.tex
\begin{landscape}
{\footnotesize\setlength{\tabcolsep}{3.0pt}\ReftTableCell
\begin{longtable}{@{}>{\raggedright\arraybackslash}p{140.00pt}>{\raggedleft\arraybackslash}p{84.94pt}>{\raggedleft\arraybackslash}p{92.69pt}>{\raggedleft\arraybackslash}p{79.37pt}>{\raggedleft\arraybackslash}p{84.94pt}>{\raggedleft\arraybackslash}p{92.69pt}>{\raggedleft\arraybackslash}p{79.37pt}@{\hskip 3.00pt}}
\ReftCaption{Supplementary~Table~S3}{Pairwise Condition Comparison E-Values}{Testing-by-betting pairwise e-values for each configuration in the primary experiment and the ablation experiment, quantifying whether accuracy differs between two conditions within a single configuration. The transparent-versus-opaque column is the one tied to the transparent--opaque accuracy gap; the transparent-versus-correct-draft and opaque-versus-correct-draft columns complete each configuration's three comparisons. The dagger (†) marks the authoring model, Claude Opus 4.6†. A large e-value is strong evidence that the two conditions differ in accuracy, whereas a value near 1 is little evidence either way. Each value is the evidence for one comparison on its own; a value here does not by itself determine significance, which also depends on the configuration's evidence across all three of its comparisons. A value below 1 is not evidence that the two conditions' accuracies are close, only that the betting process lost wealth because later trials ran against the bets it had placed. Every such value in this table compares opaque with correct-draft stimuli, the smallest being 0.188 for Claude Haiku 4.5 with reasoning off in the ablation experiment. Exactly 1.000 is the starting stake left untouched, which every Claude Opus 4.6† configuration returns for the transparent-versus-opaque comparison in both experiments, where the two conditions' outcomes differ on at most 1 of the 200 paired trials. A value below 10,000 is shown in fixed notation and a value at or above 10,000 in scientific notation.}\label{art:supplementary_table_s3}\\
\toprule
 & \multicolumn{3}{>{\centering\arraybackslash}p{269.00pt}}{\textbf{Primary Experiment}} & \multicolumn{3}{>{\centering\arraybackslash}p{269.00pt}}{\textbf{Ablation Experiment}} \\
\cmidrule(lr){2-4}\cmidrule(lr){5-7}
\textbf{Configuration} & \textbf{Transparent vs Opaque E-value} & \textbf{Transparent vs Correct-Draft E-value} & \textbf{Opaque vs Correct-Draft E-value} & \textbf{Transparent vs Opaque E-value} & \textbf{Transparent vs Correct-Draft E-value} & \textbf{Opaque vs Correct-Draft E-value} \\
\midrule\endfirsthead
\toprule
 & \multicolumn{3}{>{\centering\arraybackslash}p{269.00pt}}{\textbf{Primary Experiment}} & \multicolumn{3}{>{\centering\arraybackslash}p{269.00pt}}{\textbf{Ablation Experiment}} \\
\cmidrule(lr){2-4}\cmidrule(lr){5-7}
\textbf{Configuration} & \textbf{Transparent vs Opaque E-value} & \textbf{Transparent vs Correct-Draft E-value} & \textbf{Opaque vs Correct-Draft E-value} & \textbf{Transparent vs Opaque E-value} & \textbf{Transparent vs Correct-Draft E-value} & \textbf{Opaque vs Correct-Draft E-value} \\
\midrule\endhead
\endfoot
\bottomrule\endlastfoot
Claude Haiku 4.5 / off & 9.272e+17 & 9.529e+15 & 0.576 & 4.183e+14 & 3.137e+14 & 0.188 \\
Claude Haiku 4.5 / enabled & 1.000 & 3.228e+09 & 3.228e+09 & 1.000 & 6.376e+08 & 6.376e+08 \\
Claude Opus 4.6† / low & 1.000 & 17.086 & 5.066 & 1.000 & 1.000 & 1.000 \\
Claude Opus 4.6† / medium & 1.000 & 1.000 & 1.000 & 1.000 & 1.000 & 1.000 \\
Claude Opus 4.6† / high & 1.000 & 1.000 & 1.000 & 1.000 & 1.500 & 1.500 \\
Claude Opus 4.6† / max & 1.000 & 17.086 & 17.086 & 1.000 & 86.498 & 86.498 \\
Gemini 3 Flash / minimal & 2.120e+12 & 4.314e+05 & 4.914e+06 & 2.834e+08 & 5.682e+04 & 665.802 \\
Gemini 3 Flash / low & 8.397e+07 & 1.456e+06 & 2.052 & 3.788e+04 & 2.184e+06 & 4.617 \\
Gemini 3 Flash / medium & 2.250 & 1.106e+07 & 3.276e+06 & 3.375 & 3.732e+07 & 7.372e+06 \\
Gemini 3 Flash / high & 5.062 & 1.659e+07 & 2.184e+06 & 1.000 & 3.228e+09 & 3.228e+09 \\
Gemini 3 Pro / low & 17.086 & 38.443 & 0.311 & 57.665 & 7.594 & 0.368 \\
Gemini 3 Pro / high & 5.062 & 2216.838 & 291.929 & 1.000 & 2216.838 & 2216.838 \\
GPT-5-mini / minimal & 5.121e+25 & 3.414e+25 & 3.686e+06 & 7.912e+19 & 2.592e+26 & 1.811e+13 \\
GPT-5-mini / low & 3.732e+07 & 2.058e+18 & 1.747e+09 & 2.525e+04 & 6.098e+17 & 1.360e+12 \\
GPT-5-mini / medium & 291.929 & 2.792e+11 & 1.513e+08 & 57.665 & 3.677e+10 & 1.345e+08 \\
GPT-5-mini / high & 1.500 & 3.276e+06 & 1.092e+06 & 3.375 & 7.372e+06 & 6.143e+05 \\
GPT-5.2 / none & 3.732e+07 & 1.834e+14 & 4.914e+06 & 4.914e+06 & 3.623e+13 & 2.027e+05 \\
GPT-5.2 / low & 3.064e+09 & 4.066e+17 & 3.156e+05 & 3.026e+08 & 3.917e+14 & 779.409 \\
GPT-5.2 / medium & 4.136e+10 & 1.719e+13 & 1.098 & 8.397e+07 & 1.861e+11 & 4.349 \\
GPT-5.2 / high & 2.758e+10 & 1.811e+13 & 1.028 & 2.184e+06 & 1.435e+09 & 3.197 \\
GPT-5.2 / xhigh & 985.261 & 43.249 & 0.355 & 86.498 & 129.746 & 0.211 \\
\end{longtable}}
\end{landscape}

%% file: items/e_value_distribution_and_power_diagnostics_paired.tex
\begin{landscape}
{\footnotesize\setlength{\tabcolsep}{3.0pt}\ReftTableCell
\begin{longtable}{@{}>{\raggedright\arraybackslash}p{75.00pt}>{\raggedleft\arraybackslash}p{61.43pt}>{\raggedleft\arraybackslash}p{74.65pt}>{\raggedleft\arraybackslash}p{56.43pt}>{\raggedleft\arraybackslash}p{74.65pt}>{\raggedleft\arraybackslash}p{61.43pt}>{\raggedleft\arraybackslash}p{72.80pt}>{\raggedleft\arraybackslash}p{87.39pt}>{\raggedleft\arraybackslash}p{78.22pt}@{\hskip 3.00pt}}
\ReftCaption{Supplementary~Table~S4}{E-Value Distribution and Power Diagnostics}{Per-condition summary of the per-cell log e-value distribution, with 210 cells behind each condition row and 630 cells in each of the primary and ablation experiments. Each cell's log e-value comes from a single batch of 20 trials scored against a chance null of 0.5 and a pre-specified seed alternative of 0.7. That construction confines the value to the interval from -10.217 (0 of 20 trials correct) to 6.729 (20 of 20 correct). A positive value is evidence that the cell's accuracy is above chance, and a cell counts as rejected once its log e-value reaches 2.996, an e-value of 20. A negative value is evidence that the cell's accuracy falls short of the 0.7 alternative, not that it sits at chance: the only hypotheses compared are chance and that better-than-chance alternative, so the most negative values come from cells that were systematically wrong rather than cells at chance. Alongside the five quantiles, the table reports the mean empirical e-power (the mean log e-value gained per batch), the mean width of the two-sided confidence sequence used for accuracy estimation, and the percentage of cells resolved by rejection. Rejection came overwhelmingly from cells sitting at that upper bound, which only a perfect 20 of 20 can reach: 419 of the primary experiment's 479 rejected cells (87.5\%) and 439 of the ablation experiment's 495 rejected cells (88.7\%).}\label{art:supplementary_table_s4}\\
\toprule
\textbf{Condition} & \textbf{Log E-value Minimum} & \textbf{Log E-value First Quartile} & \textbf{Log E-value Median} & \textbf{Log E-value Third Quartile} & \textbf{Log E-value Maximum} & \textbf{Mean Empirical E-power} & \textbf{Mean Confidence Sequence Width} & \textbf{Cells Rejected~(\%)} \\
\midrule\endfirsthead
\toprule
\textbf{Condition} & \textbf{Log E-value Minimum} & \textbf{Log E-value First Quartile} & \textbf{Log E-value Median} & \textbf{Log E-value Third Quartile} & \textbf{Log E-value Maximum} & \textbf{Mean Empirical E-power} & \textbf{Mean Confidence Sequence Width} & \textbf{Cells Rejected~(\%)} \\
\midrule\endhead
\endfoot
\bottomrule\endlastfoot
\multicolumn{9}{@{}>{\raggedright\arraybackslash}p{690.00pt}@{\hskip 3.00pt}}{\textbf{Primary Experiment}} \\*
\noalign{\vskip\aboverulesep\nointerlineskip}
\ReftCutInRule{9}\\*
\noalign{\kern\belowrulesep\nointerlineskip}
Correct-Draft & -10.217 & -4.921 & 5.882 & 6.729 & 6.729 & 1.444 & 0.330 & 61.0\% \\*
Transparent & -10.217 & 6.729 & 6.729 & 6.729 & 6.729 & 6.011 & 0.242 & 94.8\% \\*
Opaque & -10.217 & 1.010 & 6.729 & 6.729 & 6.729 & 3.118 & 0.298 & 72.4\% \\
\addlinespace
\multicolumn{9}{@{}>{\raggedright\arraybackslash}p{690.00pt}@{\hskip 3.00pt}}{\textbf{Ablation Experiment}} \\*
\noalign{\vskip\aboverulesep\nointerlineskip}
\ReftCutInRule{9}\\*
\noalign{\kern\belowrulesep\nointerlineskip}
Correct-Draft & -10.217 & -3.226 & 5.882 & 6.729 & 6.729 & 1.872 & 0.326 & 63.3\% \\*
Transparent & -10.217 & 6.729 & 6.729 & 6.729 & 6.729 & 6.169 & 0.238 & 95.7\% \\*
Opaque & -10.217 & 4.188 & 6.729 & 6.729 & 6.729 & 3.982 & 0.302 & 76.7\% \\
\end{longtable}}
\end{landscape}

%% file: items/cross_dataset_ablation_significance_tests_paired.tex
\begin{landscape}
{\footnotesize\setlength{\tabcolsep}{3.0pt}\ReftTableCell
\begin{longtable}{@{}>{\raggedright\arraybackslash}p{140.00pt}>{\raggedright\arraybackslash}p{75.00pt}>{\raggedleft\arraybackslash}p{55.67pt}>{\raggedleft\arraybackslash}p{60.67pt}>{\raggedleft\arraybackslash}p{55.43pt}>{\raggedleft\arraybackslash}p{50.23pt}>{\raggedleft\arraybackslash}p{65.00pt}@{\hskip 12.00pt}>{\raggedright\arraybackslash}p{140.00pt}@{\hskip 3.00pt}}
\ReftCaption{Supplementary~Table~S5}{Cross-Dataset Ablation Significance Tests, Condition Level}{Cross-dataset paired testing-by-betting e-value tests comparing the ablation experiment versus the primary experiment on per-condition accuracy for every configuration-by-condition pairing (21 configurations × 3 conditions = 63 tests). Each per-condition accuracy is computed over 200 valid trials per experiment. Each test is two-sided: it runs one process betting that the ablation experiment scores higher and one betting that it scores lower, over the same paired rounds, and takes the larger of the two. The reported adjusted e-value therefore carries a factor-of-2 correction absorbing that direction selection, on top of the e-Bonferroni correction across the 63-test condition-level family; that family and the 21-test configuration-level family are corrected independently. An adjusted e-value of at least 20 (α = 0.05) is counted as significant. The Direction column names whichever of the two one-sided tests accumulated positive evidence and reads Indistinguishable when neither did, so it does not always follow the sign of the accuracy difference. The dagger (†) marks the authoring model, Claude Opus 4.6†. Three of 63 condition-level tests reached significance, all on opaque stimuli and all favoring the ablation experiment: GPT-5-mini at minimal reasoning effort (+0.190), GPT-5.2 at high reasoning effort (+0.140), and Gemini 3 Flash at minimal reasoning effort (+0.110). No transparent comparison reached significance, largely because of ceiling effects: 14 of the 21 configurations held transparent accuracy at 1.000 in both experiments, leaving those comparisons no difference to detect.}\label{art:supplementary_table_s5}\\
\toprule
\textbf{Configuration} & \textbf{Condition} & \textbf{Per-Condition Accuracy [Primary]} & \textbf{Per-Condition Accuracy [Ablation]} & \textbf{Δ~Per-Condition Accuracy} & \textbf{Adjusted E-value} & \textbf{Significant} & \textbf{Direction} \\
\midrule\endfirsthead
\toprule
\textbf{Configuration} & \textbf{Condition} & \textbf{Per-Condition Accuracy [Primary]} & \textbf{Per-Condition Accuracy [Ablation]} & \textbf{Δ~Per-Condition Accuracy} & \textbf{Adjusted E-value} & \textbf{Significant} & \textbf{Direction} \\
\midrule\endhead
\endfoot
\bottomrule\endlastfoot
Claude Haiku 4.5 / off & Correct-Draft & 0.335 & 0.415 & +0.080 & 0.044 & No & Ablation Experiment Higher \\
Claude Haiku 4.5 / off & Transparent & 0.830 & 0.875 & +0.045 & 0.010 & No & Ablation Experiment Higher \\
Claude Haiku 4.5 / off & Opaque & 0.275 & 0.405 & +0.130 & 0.224 & No & Ablation Experiment Higher \\
Claude Haiku 4.5 / enabled & Correct-Draft & 0.710 & 0.745 & +0.035 & 0.005 & No & Indistinguishable \\
Claude Haiku 4.5 / enabled & Transparent & 1.000 & 1.000 & +0.000 & 0.008 & No & Indistinguishable \\
Claude Haiku 4.5 / enabled & Opaque & 1.000 & 1.000 & +0.000 & 0.008 & No & Indistinguishable \\
Claude Opus 4.6† / low & Correct-Draft & 0.960 & 1.000 & +0.040 & 0.136 & No & Ablation Experiment Higher \\
Claude Opus 4.6† / low & Transparent & 1.000 & 1.000 & +0.000 & 0.008 & No & Indistinguishable \\
Claude Opus 4.6† / low & Opaque & 0.995 & 1.000 & +0.005 & 0.008 & No & Indistinguishable \\
Claude Opus 4.6† / medium & Correct-Draft & 1.000 & 1.000 & +0.000 & 0.008 & No & Indistinguishable \\
Claude Opus 4.6† / medium & Transparent & 1.000 & 1.000 & +0.000 & 0.008 & No & Indistinguishable \\
Claude Opus 4.6† / medium & Opaque & 1.000 & 1.000 & +0.000 & 0.008 & No & Indistinguishable \\
Claude Opus 4.6† / high & Correct-Draft & 1.000 & 0.990 & -0.010 & 0.012 & No & Ablation Experiment Lower \\
Claude Opus 4.6† / high & Transparent & 1.000 & 1.000 & +0.000 & 0.008 & No & Indistinguishable \\
Claude Opus 4.6† / high & Opaque & 1.000 & 1.000 & +0.000 & 0.008 & No & Indistinguishable \\
Claude Opus 4.6† / max & Correct-Draft & 0.960 & 0.940 & -0.020 & 0.005 & No & Indistinguishable \\
Claude Opus 4.6† / max & Transparent & 1.000 & 1.000 & +0.000 & 0.008 & No & Indistinguishable \\
Claude Opus 4.6† / max & Opaque & 1.000 & 1.000 & +0.000 & 0.008 & No & Indistinguishable \\
Gemini 3 Flash / minimal & Correct-Draft & 0.830 & 0.860 & +0.030 & 0.025 & No & Ablation Experiment Higher \\
Gemini 3 Flash / minimal & Transparent & 1.000 & 1.000 & +0.000 & 0.008 & No & Indistinguishable \\
Gemini 3 Flash / minimal & Opaque & 0.635 & 0.745 & +0.110 & 29.690 & Yes & Ablation Experiment Higher \\
Gemini 3 Flash / low & Correct-Draft & 0.815 & 0.810 & -0.005 & 0.008 & No & Indistinguishable \\
Gemini 3 Flash / low & Transparent & 1.000 & 1.000 & +0.000 & 0.008 & No & Indistinguishable \\
Gemini 3 Flash / low & Opaque & 0.765 & 0.860 & +0.095 & 4.750 & No & Ablation Experiment Higher \\
Gemini 3 Flash / medium & Correct-Draft & 0.795 & 0.775 & -0.020 & 0.008 & No & Indistinguishable \\
Gemini 3 Flash / medium & Transparent & 1.000 & 1.000 & +0.000 & 0.008 & No & Indistinguishable \\
Gemini 3 Flash / medium & Opaque & 0.985 & 0.980 & -0.005 & 0.006 & No & Indistinguishable \\
Gemini 3 Flash / high & Correct-Draft & 0.780 & 0.715 & -0.065 & 0.071 & No & Ablation Experiment Lower \\
Gemini 3 Flash / high & Transparent & 1.000 & 1.000 & +0.000 & 0.008 & No & Indistinguishable \\
Gemini 3 Flash / high & Opaque & 0.975 & 1.000 & +0.025 & 0.040 & No & Ablation Experiment Higher \\
Gemini 3 Pro / low & Correct-Draft & 0.950 & 0.970 & +0.020 & 0.015 & No & Ablation Experiment Higher \\
Gemini 3 Pro / low & Transparent & 1.000 & 1.000 & +0.000 & 0.008 & No & Indistinguishable \\
Gemini 3 Pro / low & Opaque & 0.960 & 0.945 & -0.015 & 0.006 & No & Indistinguishable \\
Gemini 3 Pro / high & Correct-Draft & 0.895 & 0.900 & +0.005 & 0.008 & No & Indistinguishable \\
Gemini 3 Pro / high & Transparent & 1.000 & 1.000 & +0.000 & 0.008 & No & Indistinguishable \\
Gemini 3 Pro / high & Opaque & 0.975 & 1.000 & +0.025 & 0.040 & No & Ablation Experiment Higher \\
GPT-5-mini / minimal & Correct-Draft & 0.015 & 0.005 & -0.010 & 0.011 & No & Ablation Experiment Lower \\
GPT-5-mini / minimal & Transparent & 0.985 & 1.000 & +0.015 & 0.018 & No & Ablation Experiment Higher \\
GPT-5-mini / minimal & Opaque & 0.215 & 0.405 & +0.190 & 653.540 & Yes & Ablation Experiment Higher \\
GPT-5-mini / low & Correct-Draft & 0.460 & 0.475 & +0.015 & 0.008 & No & Indistinguishable \\
GPT-5-mini / low & Transparent & 1.000 & 1.000 & +0.000 & 0.008 & No & Indistinguishable \\
GPT-5-mini / low & Opaque & 0.780 & 0.870 & +0.090 & 0.512 & No & Ablation Experiment Higher \\
GPT-5-mini / medium & Correct-Draft & 0.655 & 0.690 & +0.035 & 0.008 & No & Indistinguishable \\
GPT-5-mini / medium & Transparent & 1.000 & 1.000 & +0.000 & 0.008 & No & Indistinguishable \\
GPT-5-mini / medium & Opaque & 0.925 & 0.945 & +0.020 & 0.008 & No & Indistinguishable \\
GPT-5-mini / high & Correct-Draft & 0.810 & 0.800 & -0.010 & 0.005 & No & Indistinguishable \\
GPT-5-mini / high & Transparent & 1.000 & 1.000 & +0.000 & 0.008 & No & Indistinguishable \\
GPT-5-mini / high & Opaque & 0.990 & 0.980 & -0.010 & 0.008 & No & Indistinguishable \\
GPT-5.2 / none & Correct-Draft & 0.000 & 0.080 & +0.080 & 3.475 & No & Ablation Experiment Higher \\
GPT-5.2 / none & Transparent & 0.425 & 0.480 & +0.055 & 0.117 & No & Ablation Experiment Higher \\
GPT-5.2 / none & Opaque & 0.200 & 0.280 & +0.080 & 3.475 & No & Ablation Experiment Higher \\
GPT-5.2 / low & Correct-Draft & 0.400 & 0.500 & +0.100 & 0.043 & No & Ablation Experiment Higher \\
GPT-5.2 / low & Transparent & 0.925 & 0.950 & +0.025 & 0.008 & No & Indistinguishable \\
GPT-5.2 / low & Opaque & 0.635 & 0.670 & +0.035 & 0.006 & No & Indistinguishable \\
GPT-5.2 / medium & Correct-Draft & 0.565 & 0.650 & +0.085 & 0.011 & No & Ablation Experiment Higher \\
GPT-5.2 / medium & Transparent & 0.970 & 1.000 & +0.030 & 0.060 & No & Ablation Experiment Higher \\
GPT-5.2 / medium & Opaque & 0.650 & 0.760 & +0.110 & 0.402 & No & Ablation Experiment Higher \\
GPT-5.2 / high & Correct-Draft & 0.575 & 0.725 & +0.150 & 2.289 & No & Ablation Experiment Higher \\
GPT-5.2 / high & Transparent & 0.980 & 1.000 & +0.020 & 0.027 & No & Ablation Experiment Higher \\
GPT-5.2 / high & Opaque & 0.670 & 0.810 & +0.140 & 35.971 & Yes & Ablation Experiment Higher \\
GPT-5.2 / xhigh & Correct-Draft & 0.940 & 0.935 & -0.005 & 0.003 & No & Indistinguishable \\
GPT-5.2 / xhigh & Transparent & 0.995 & 1.000 & +0.005 & 0.008 & No & Indistinguishable \\
GPT-5.2 / xhigh & Opaque & 0.895 & 0.940 & +0.045 & 0.020 & No & Ablation Experiment Higher \\
\end{longtable}}
\end{landscape}

%% file: items/adjacent_pair_degradation_tests_paired.tex
\begin{landscape}
{\footnotesize\setlength{\tabcolsep}{3.0pt}\ReftTableCell
\begin{longtable}{@{}>{\raggedright\arraybackslash}p{55.00pt}>{\raggedright\arraybackslash}p{90.00pt}>{\raggedright\arraybackslash}p{80.00pt}>{\raggedleft\arraybackslash}p{102.68pt}>{\raggedleft\arraybackslash}p{102.68pt}>{\raggedleft\arraybackslash}p{82.58pt}>{\raggedleft\arraybackslash}p{70.07pt}>{\raggedleft\arraybackslash}p{65.00pt}@{\hskip 3.00pt}}
\ReftCaption{Supplementary~Table~S6}{Adjacent-Pair Reasoning Effort Level Degradation Tests}{One-sided paired testing-by-betting e-value tests for each adjacent reasoning effort level transition within each graduated-reasoning model, testing whether the upper level's balanced accuracy is lower than the lower level's; the null is that the upper level is at least as accurate as the lower. The dagger (†) marks the authoring model, Claude Opus 4.6†: it helped author the stimulus components, so its results may reflect familiarity with its own output patterns rather than the capability being measured. The change shown in the table is in balanced accuracy, which is also what the test compares: every stimulus pairs on both sides, and the e-value bets on those paired per-stimulus differences weighted so that correct-draft and incorrect-draft stimuli carry equal total weight in each betting round. Each balanced accuracy is computed over 600 valid trials per configuration. E-values are e-Bonferroni corrected across K = 15 transitions, and an adjusted e-value of at least 20 (one-sided test, α = 0.05) is counted as significant. None of the 15 transitions reached significance in either experiment. In both experiments, the largest change in balanced accuracy was an improvement: GPT-5.2 from none to low reasoning effort (+0.434 in the primary experiment, +0.425 in the ablation experiment). An adjusted e-value of 0.067 is 1/K, the value of a process that never bet because its accumulated paired differences never favored the lower level; it records no evidence in either direction, and 11 of the primary experiment's 15 transitions and 10 of the ablation experiment's carry it. The strongest evidence of degradation, Claude Opus 4.6† from high to max reasoning effort, reached raw e-values of 50.265 in the primary experiment and 58.038 in the ablation experiment, adjusted to 3.351 and 3.869 by the correction.}\label{art:supplementary_table_s6}\\
\toprule
\textbf{Provider} & \textbf{Model} & \textbf{Transition} & \textbf{Lower Level Balanced Accuracy} & \textbf{Upper Level Balanced Accuracy} & \textbf{Δ~Balanced Accuracy} & \textbf{Adjusted E-value} & \textbf{Significant} \\
\midrule\endfirsthead
\toprule
\textbf{Provider} & \textbf{Model} & \textbf{Transition} & \textbf{Lower Level Balanced Accuracy} & \textbf{Upper Level Balanced Accuracy} & \textbf{Δ~Balanced Accuracy} & \textbf{Adjusted E-value} & \textbf{Significant} \\
\midrule\endhead
\endfoot
\bottomrule\endlastfoot
\multicolumn{8}{@{}>{\raggedright\arraybackslash}p{690.00pt}@{\hskip 3.00pt}}{\textbf{Primary Experiment}} \\*
\noalign{\vskip\aboverulesep\nointerlineskip}
\ReftCutInRule{8}\\*
\noalign{\kern\belowrulesep\nointerlineskip}
Anthropic & Claude Haiku 4.5 & off to enabled & 0.444 & 0.855 & +0.411 & 0.067 & No \\*
Anthropic & Claude Opus 4.6† & low to medium & 0.979 & 1.000 & +0.021 & 0.067 & No \\*
Anthropic & Claude Opus 4.6† & medium to high & 1.000 & 1.000 & +0.000 & 0.067 & No \\*
Anthropic & Claude Opus 4.6† & high to max & 1.000 & 0.980 & -0.020 & 3.351 & No \\*
Google & Gemini 3 Flash & minimal to low & 0.824 & 0.849 & +0.025 & 0.067 & No \\*
Google & Gemini 3 Flash & low to medium & 0.849 & 0.894 & +0.045 & 0.067 & No \\*
Google & Gemini 3 Flash & medium to high & 0.894 & 0.884 & -0.010 & 0.003 & No \\*
Google & Gemini 3 Pro & low to high & 0.965 & 0.941 & -0.024 & 0.866 & No \\*
OpenAI & GPT-5-mini & minimal to low & 0.307 & 0.675 & +0.368 & 0.067 & No \\*
OpenAI & GPT-5-mini & low to medium & 0.675 & 0.809 & +0.134 & 0.067 & No \\*
OpenAI & GPT-5-mini & medium to high & 0.809 & 0.903 & +0.094 & 0.067 & No \\*
OpenAI & GPT-5.2 & none to low & 0.156 & 0.590 & +0.434 & 0.067 & No \\*
OpenAI & GPT-5.2 & low to medium & 0.590 & 0.688 & +0.098 & 0.067 & No \\*
OpenAI & GPT-5.2 & medium to high & 0.688 & 0.700 & +0.012 & 0.005 & No \\*
OpenAI & GPT-5.2 & high to xhigh & 0.700 & 0.942 & +0.242 & 0.067 & No \\
\addlinespace
\multicolumn{8}{@{}>{\raggedright\arraybackslash}p{690.00pt}@{\hskip 3.00pt}}{\textbf{Ablation Experiment}} \\*
\noalign{\vskip\aboverulesep\nointerlineskip}
\ReftCutInRule{8}\\*
\noalign{\kern\belowrulesep\nointerlineskip}
Anthropic & Claude Haiku 4.5 & off to enabled & 0.527 & 0.873 & +0.346 & 0.067 & No \\*
Anthropic & Claude Opus 4.6† & low to medium & 1.000 & 1.000 & +0.000 & 0.067 & No \\*
Anthropic & Claude Opus 4.6† & medium to high & 1.000 & 0.995 & -0.005 & 0.117 & No \\*
Anthropic & Claude Opus 4.6† & high to max & 0.995 & 0.970 & -0.025 & 3.869 & No \\*
Google & Gemini 3 Flash & minimal to low & 0.866 & 0.870 & +0.004 & 0.039 & No \\*
Google & Gemini 3 Flash & low to medium & 0.870 & 0.883 & +0.013 & 0.067 & No \\*
Google & Gemini 3 Flash & medium to high & 0.883 & 0.857 & -0.026 & 0.061 & No \\*
Google & Gemini 3 Pro & low to high & 0.971 & 0.950 & -0.021 & 0.526 & No \\*
OpenAI & GPT-5-mini & minimal to low & 0.354 & 0.705 & +0.351 & 0.067 & No \\*
OpenAI & GPT-5-mini & low to medium & 0.705 & 0.831 & +0.126 & 0.067 & No \\*
OpenAI & GPT-5-mini & medium to high & 0.831 & 0.895 & +0.064 & 0.067 & No \\*
OpenAI & GPT-5.2 & none to low & 0.230 & 0.655 & +0.425 & 0.067 & No \\*
OpenAI & GPT-5.2 & low to medium & 0.655 & 0.765 & +0.110 & 0.067 & No \\*
OpenAI & GPT-5.2 & medium to high & 0.765 & 0.815 & +0.050 & 0.067 & No \\*
OpenAI & GPT-5.2 & high to xhigh & 0.815 & 0.953 & +0.138 & 0.067 & No \\
\end{longtable}}
\end{landscape}

%% file: items/cost_breakdown_paired.tex
\begin{landscape}
{\footnotesize\setlength{\tabcolsep}{3.0pt}\ReftTableCell
\begin{longtable}{@{}>{\raggedright\arraybackslash}p{140.00pt}>{\raggedleft\arraybackslash}p{78.34pt}>{\raggedleft\arraybackslash}p{83.34pt}>{\raggedleft\arraybackslash}p{68.89pt}>{\raggedleft\arraybackslash}p{60.74pt}>{\raggedleft\arraybackslash}p{67.60pt}>{\raggedleft\arraybackslash}p{77.04pt}>{\raggedleft\arraybackslash}p{72.04pt}@{\hskip 3.00pt}}
\ReftCaption{Supplementary~Table~S7}{Cost per Trial and Total Experiment Cost by Configuration}{Per-configuration cost breakdown for both experiments, covering balanced accuracy, cost per trial, total cost at 600 trials, input token total, output token total, reasoning token total, and response token total. Balanced accuracy is computed over 600 valid trials per configuration. Costs are in U.S. dollars. The dagger (†) marks the authoring model, Claude Opus 4.6†. The output token column follows each provider's own accounting: OpenAI and Anthropic fold reasoning tokens into the output total, so output tokens equal reasoning plus response tokens; Google reports reasoning tokens outside its output count, so in the Gemini rows the output total equals the response total and excludes the reasoning tokens printed beside it. An em-dash marks a token total that is zero or that the provider never reported, and it appears in the reasoning token column for every configuration that recorded no reasoning tokens. Claude Haiku 4.5 with reasoning off omits the extended-thinking block, so its reasoning total is zero and its response total equals its output total. The reasoning token counts themselves have two sources: OpenAI and Google report them directly, while Anthropic reports only a combined output total, so the Anthropic reasoning token counts are derived by measuring each response's visible text with Anthropic's token-counting endpoint and subtracting that count from the reported output total. Every cost is a batch price: each component was billed at half the provider's list rate through the batch APIs, so an interactive deployment of the same configurations would pay approximately twice these amounts. The table is the data source behind the contrast between GPT-5.2 at xhigh reasoning effort and Gemini 3 Flash at medium reasoning effort: in the primary experiment, GPT-5.2 at xhigh reasoning effort consumed 4.6 million total reasoning tokens at \$0.0556 per trial for a balanced accuracy of 0.942, while Gemini 3 Flash at medium reasoning effort reached 0.894 at \$0.0022 per trial. The ablation experiment showed the same configurations at similar per-trial costs (\$0.0509 and \$0.0021) and balanced accuracies within a few hundredths of the primary experiment's values; provider pricing is identical across experiments, but token usage differed.}\label{art:supplementary_table_s7}\\
\toprule
\textbf{Configuration} & \textbf{Balanced Accuracy} & \textbf{Cost per Trial~(\$)} & \textbf{Total Cost~(\$)} & \textbf{Input Tokens} & \textbf{Output Tokens} & \textbf{Reasoning Tokens} & \textbf{Response Tokens} \\
\midrule\endfirsthead
\toprule
\textbf{Configuration} & \textbf{Balanced Accuracy} & \textbf{Cost per Trial~(\$)} & \textbf{Total Cost~(\$)} & \textbf{Input Tokens} & \textbf{Output Tokens} & \textbf{Reasoning Tokens} & \textbf{Response Tokens} \\
\midrule\endhead
\endfoot
\bottomrule\endlastfoot
\multicolumn{8}{@{}>{\raggedright\arraybackslash}p{690.00pt}@{\hskip 3.00pt}}{\textbf{Primary Experiment}} \\*
\noalign{\vskip\aboverulesep\nointerlineskip}
\ReftCutInRule{8}\\*
\noalign{\kern\belowrulesep\nointerlineskip}
Claude Opus 4.6† / medium & 1.000 & \$0.0077 & \$4.62 & 898,660 & 189,823 & 105,726 & 84,097 \\*
Claude Opus 4.6† / high & 1.000 & \$0.0129 & \$7.73 & 898,660 & 438,364 & 321,508 & 116,856 \\*
Claude Opus 4.6† / max & 0.980 & \$0.0199 & \$11.94 & 898,660 & 775,774 & 645,581 & 130,193 \\*
Claude Opus 4.6† / low & 0.979 & \$0.0053 & \$3.17 & 898,660 & 73,479 & 15,777 & 57,702 \\*
Gemini 3 Pro / low & 0.965 & \$0.0030 & \$1.80 & 727,480 & 108,952 & 70,298 & 108,952 \\*
GPT-5.2 / xhigh & 0.942 & \$0.0556 & \$33.38 & 686,520 & 4,707,646 & 4,566,439 & 141,207 \\*
Gemini 3 Pro / high & 0.941 & \$0.0086 & \$5.14 & 727,480 & 82,200 & 653,907 & 82,200 \\*
GPT-5-mini / high & 0.903 & \$0.0031 & \$1.83 & 686,520 & 1,748,514 & 1,567,552 & 180,962 \\*
Gemini 3 Flash / medium & 0.894 & \$0.0022 & \$1.31 & 727,480 & 92,933 & 660,198 & 92,933 \\*
Gemini 3 Flash / high & 0.884 & \$0.0023 & \$1.39 & 727,480 & 94,054 & 710,146 & 94,054 \\*
Claude Haiku 4.5 / enabled & 0.855 & \$0.0063 & \$3.78 & 905,860 & 1,330,002 & 1,192,110 & 137,892 \\*
Gemini 3 Flash / low & 0.849 & \$0.0005 & \$0.32 & 727,480 & 89,980 & — & 89,980 \\*
Gemini 3 Flash / minimal & 0.824 & \$0.0005 & \$0.30 & 727,480 & 77,623 & — & 77,623 \\*
GPT-5-mini / medium & 0.809 & \$0.0010 & \$0.60 & 686,520 & 519,382 & 364,160 & 155,222 \\*
GPT-5.2 / high & 0.700 & \$0.0071 & \$4.25 & 686,520 & 537,157 & 398,915 & 138,242 \\*
GPT-5.2 / medium & 0.688 & \$0.0050 & \$2.98 & 686,571 & 353,222 & 209,365 & 143,857 \\*
GPT-5-mini / low & 0.675 & \$0.0005 & \$0.30 & 686,520 & 215,902 & 88,768 & 127,134 \\*
GPT-5.2 / low & 0.590 & \$0.0033 & \$1.97 & 686,520 & 208,661 & 80,463 & 128,198 \\*
Claude Haiku 4.5 / off & 0.444 & \$0.0014 & \$0.83 & 887,860 & 153,254 & — & 153,254 \\*
GPT-5-mini / minimal & 0.307 & \$0.0004 & \$0.22 & 686,520 & 140,675 & — & 140,675 \\*
GPT-5.2 / none & 0.156 & \$0.0022 & \$1.32 & 686,520 & 113,319 & — & 113,319 \\
\addlinespace
\multicolumn{8}{@{}>{\raggedright\arraybackslash}p{690.00pt}@{\hskip 3.00pt}}{\textbf{Ablation Experiment}} \\*
\noalign{\vskip\aboverulesep\nointerlineskip}
\ReftCutInRule{8}\\*
\noalign{\kern\belowrulesep\nointerlineskip}
Claude Opus 4.6† / low & 1.000 & \$0.0049 & \$2.96 & 866,860 & 63,547 & 11,123 & 52,424 \\*
Claude Opus 4.6† / medium & 1.000 & \$0.0072 & \$4.30 & 866,860 & 170,874 & 90,012 & 80,862 \\*
Claude Opus 4.6† / high & 0.995 & \$0.0121 & \$7.27 & 866,860 & 408,504 & 298,065 & 110,439 \\*
Gemini 3 Pro / low & 0.971 & \$0.0029 & \$1.73 & 696,280 & 106,153 & 66,393 & 106,153 \\*
Claude Opus 4.6† / max & 0.970 & \$0.0177 & \$10.61 & 866,860 & 675,364 & 554,712 & 120,652 \\*
GPT-5.2 / xhigh & 0.953 & \$0.0509 & \$30.57 & 660,120 & 4,305,578 & 4,167,831 & 137,747 \\*
Gemini 3 Pro / high & 0.950 & \$0.0084 & \$5.07 & 696,280 & 80,545 & 648,153 & 80,545 \\*
GPT-5-mini / high & 0.895 & \$0.0027 & \$1.61 & 660,120 & 1,542,182 & 1,370,432 & 171,750 \\*
Gemini 3 Flash / medium & 0.883 & \$0.0021 & \$1.28 & 696,280 & 91,499 & 647,351 & 91,499 \\*
Claude Haiku 4.5 / enabled & 0.873 & \$0.0057 & \$3.40 & 874,060 & 1,185,440 & 1,051,893 & 133,547 \\*
Gemini 3 Flash / low & 0.870 & \$0.0005 & \$0.30 & 696,280 & 83,357 & — & 83,357 \\*
Gemini 3 Flash / minimal & 0.866 & \$0.0005 & \$0.28 & 696,280 & 73,637 & — & 73,637 \\*
Gemini 3 Flash / high & 0.857 & \$0.0023 & \$1.36 & 696,280 & 91,641 & 697,102 & 91,641 \\*
GPT-5-mini / medium & 0.831 & \$0.0009 & \$0.56 & 660,120 & 499,195 & 351,680 & 147,515 \\*
GPT-5.2 / high & 0.815 & \$0.0078 & \$4.66 & 660,120 & 602,661 & 461,288 & 141,373 \\*
GPT-5.2 / medium & 0.765 & \$0.0047 & \$2.83 & 663,413 & 345,007 & 211,915 & 133,092 \\*
GPT-5-mini / low & 0.705 & \$0.0005 & \$0.27 & 660,120 & 206,494 & 86,336 & 120,158 \\*
GPT-5.2 / low & 0.655 & \$0.0032 & \$1.94 & 660,120 & 214,920 & 86,983 & 127,937 \\*
Claude Haiku 4.5 / off & 0.527 & \$0.0014 & \$0.81 & 856,060 & 154,311 & — & 154,311 \\*
GPT-5-mini / minimal & 0.354 & \$0.0003 & \$0.18 & 660,120 & 127,137 & — & 127,137 \\*
GPT-5.2 / none & 0.230 & \$0.0020 & \$1.22 & 660,194 & 112,073 & — & 112,073 \\
\end{longtable}}
\end{landscape}

%% file: items/within_provider_model_size_comparison_every_level_tests_paired.tex
\begin{landscape}
{\footnotesize\setlength{\tabcolsep}{3.0pt}\ReftTableCell
\begin{longtable}{@{}>{\raggedright\arraybackslash}p{55.00pt}>{\raggedright\arraybackslash}p{90.00pt}>{\raggedright\arraybackslash}p{90.00pt}>{\raggedleft\arraybackslash}p{147.25pt}>{\raggedleft\arraybackslash}p{127.83pt}>{\raggedleft\arraybackslash}p{78.92pt}>{\raggedleft\arraybackslash}p{65.00pt}@{\hskip 3.00pt}}
\ReftCaption{Supplementary~Table~S8}{Within-Provider Model Size Comparison Tests, Every-Level}{One-sided paired testing-by-betting e-value test of whether every reasoning effort level of each provider's larger model outperforms the smaller model's worst configuration on balanced accuracy, for the primary experiment and the ablation experiment. No configuration is selected on either side: every reasoning effort level of the larger model is paired with every configuration of the smaller model, the per-configuration e-values are combined with their mean within each level, and those level statistics with their minimum, so significance establishes the claim of the smaller model's true worst configuration. Comparing against the smaller model's own measured configurations, rather than against reasoning effort levels that share a name, avoids assuming those levels are comparable across models. Every configuration on both sides is evaluated over 600 valid trials. E-values are e-Bonferroni corrected across the K = 6 within-provider model size tests (this every-level test and the companion some-level test, across three providers), and an adjusted e-value of at least 20 (one-sided test, α = 0.05) is counted as significant. The dagger (†) marks tests involving the authoring model, Claude Opus 4.6†. In both experiments this test reached significance for the Anthropic and Google pairs but not for the OpenAI pair: GPT-5.2 with reasoning effort set to none fell below every GPT-5-mini configuration, the weakest of them at minimal reasoning effort (0.156 versus 0.307 in balanced accuracy in the primary experiment, 0.230 versus 0.354 in the ablation experiment), and gained no evidence against any of them.}\label{art:supplementary_table_s8}\\
\toprule
\textbf{Provider} & \textbf{Larger Model} & \textbf{Smaller Model} & \textbf{Larger Model Reasoning Effort Levels} & \textbf{Smaller Model Configurations} & \textbf{Adjusted E-value} & \textbf{Significant} \\
\midrule\endfirsthead
\toprule
\textbf{Provider} & \textbf{Larger Model} & \textbf{Smaller Model} & \textbf{Larger Model Reasoning Effort Levels} & \textbf{Smaller Model Configurations} & \textbf{Adjusted E-value} & \textbf{Significant} \\
\midrule\endhead
\endfoot
\bottomrule\endlastfoot
\multicolumn{7}{@{}>{\raggedright\arraybackslash}p{690.00pt}@{\hskip 3.00pt}}{\textbf{Primary Experiment}} \\*
\noalign{\vskip\aboverulesep\nointerlineskip}
\ReftCutInRule{7}\\*
\noalign{\kern\belowrulesep\nointerlineskip}
OpenAI & GPT-5.2 & GPT-5-mini & 5 & 4 & 0.167 & No \\*
Anthropic & Claude Opus 4.6† & Claude Haiku 4.5 & 4 & 2 & 9.527e+48 & Yes \\*
Google & Gemini 3 Pro & Gemini 3 Flash & 2 & 4 & 5.880e+10 & Yes \\
\addlinespace
\multicolumn{7}{@{}>{\raggedright\arraybackslash}p{690.00pt}@{\hskip 3.00pt}}{\textbf{Ablation Experiment}} \\*
\noalign{\vskip\aboverulesep\nointerlineskip}
\ReftCutInRule{7}\\*
\noalign{\kern\belowrulesep\nointerlineskip}
OpenAI & GPT-5.2 & GPT-5-mini & 5 & 4 & 0.167 & No \\*
Anthropic & Claude Opus 4.6† & Claude Haiku 4.5 & 4 & 2 & 6.426e+37 & Yes \\*
Google & Gemini 3 Pro & Gemini 3 Flash & 2 & 4 & 3.164e+07 & Yes \\
\end{longtable}}
\end{landscape}

%% file: items/within_provider_model_size_comparison_some_level_tests_paired.tex
\begin{landscape}
{\footnotesize\setlength{\tabcolsep}{3.0pt}\ReftTableCell
\begin{longtable}{@{}>{\raggedright\arraybackslash}p{55.00pt}>{\raggedright\arraybackslash}p{90.00pt}>{\raggedright\arraybackslash}p{90.00pt}>{\raggedleft\arraybackslash}p{147.25pt}>{\raggedleft\arraybackslash}p{127.83pt}>{\raggedleft\arraybackslash}p{78.92pt}>{\raggedleft\arraybackslash}p{65.00pt}@{\hskip 3.00pt}}
\ReftCaption{Supplementary~Table~S9}{Within-Provider Model Size Comparison Tests, Some-Level}{One-sided paired testing-by-betting e-value test of whether some reasoning effort level of each provider's larger model outperforms the smaller model's best configuration on balanced accuracy, for the primary experiment and the ablation experiment. No configuration is selected on either side: every reasoning effort level of the larger model is paired with every configuration of the smaller model, the per-configuration e-values are combined with their minimum within each level, and those level statistics with their arithmetic mean, so significance establishes the claim of the smaller model's true best configuration. Comparing against the smaller model's own measured configurations, rather than against reasoning effort levels that share a name, avoids assuming those levels are comparable across models. Every configuration on both sides is evaluated over 600 valid trials. E-values are e-Bonferroni corrected across the K = 6 within-provider model size tests (this some-level test and the companion every-level test, across three providers), and an adjusted e-value of at least 20 (one-sided test, α = 0.05) is counted as significant. The dagger (†) marks tests involving the authoring model, Claude Opus 4.6†. In both experiments this test reached significance for the Anthropic and Google pairs but not for the OpenAI pair: no GPT-5.2 level showed enough evidence against the GPT-5-mini configuration it fared worst against, and GPT-5-mini's strongest configuration in both experiments was high reasoning effort, at a balanced accuracy of 0.903 in the primary experiment and 0.895 in the ablation experiment.}\label{art:supplementary_table_s9}\\
\toprule
\textbf{Provider} & \textbf{Larger Model} & \textbf{Smaller Model} & \textbf{Larger Model Reasoning Effort Levels} & \textbf{Smaller Model Configurations} & \textbf{Adjusted E-value} & \textbf{Significant} \\
\midrule\endfirsthead
\toprule
\textbf{Provider} & \textbf{Larger Model} & \textbf{Smaller Model} & \textbf{Larger Model Reasoning Effort Levels} & \textbf{Smaller Model Configurations} & \textbf{Adjusted E-value} & \textbf{Significant} \\
\midrule\endhead
\endfoot
\bottomrule\endlastfoot
\multicolumn{7}{@{}>{\raggedright\arraybackslash}p{690.00pt}@{\hskip 3.00pt}}{\textbf{Primary Experiment}} \\*
\noalign{\vskip\aboverulesep\nointerlineskip}
\ReftCutInRule{7}\\*
\noalign{\kern\belowrulesep\nointerlineskip}
OpenAI & GPT-5.2 & GPT-5-mini & 5 & 4 & 0.282 & No \\*
Anthropic & Claude Opus 4.6† & Claude Haiku 4.5 & 4 & 2 & 1.110e+12 & Yes \\*
Google & Gemini 3 Pro & Gemini 3 Flash & 2 & 4 & 1.316e+05 & Yes \\
\addlinespace
\multicolumn{7}{@{}>{\raggedright\arraybackslash}p{690.00pt}@{\hskip 3.00pt}}{\textbf{Ablation Experiment}} \\*
\noalign{\vskip\aboverulesep\nointerlineskip}
\ReftCutInRule{7}\\*
\noalign{\kern\belowrulesep\nointerlineskip}
OpenAI & GPT-5.2 & GPT-5-mini & 5 & 4 & 0.581 & No \\*
Anthropic & Claude Opus 4.6† & Claude Haiku 4.5 & 4 & 2 & 1.375e+11 & Yes \\*
Google & Gemini 3 Pro & Gemini 3 Flash & 2 & 4 & 3.128e+06 & Yes \\
\end{longtable}}
\end{landscape}

%% file: items/within_provider_model_size_comparison_per_level_detail_paired.tex
\begin{landscape}
{\footnotesize\setlength{\tabcolsep}{3.0pt}\ReftTableCell
\begin{longtable}{@{}>{\raggedright\arraybackslash}p{55.00pt}>{\raggedright\arraybackslash}p{90.00pt}>{\raggedright\arraybackslash}p{70.35pt}>{\raggedleft\arraybackslash}p{69.74pt}>{\raggedleft\arraybackslash}p{53.16pt}>{\raggedleft\arraybackslash}p{154.58pt}>{\raggedleft\arraybackslash}p{161.16pt}@{\hskip 3.00pt}}
\ReftCaption{Supplementary~Table~S10}{Within-Provider Model Size Comparison, Per-Level Detail}{Per-level audit behind the within-provider model size comparison tests, for the primary experiment and the ablation experiment. Each row is one reasoning effort level of a provider's larger model, reporting its balanced accuracy, the valid trial count behind the test, and two combinations of the e-values it contributes against the smaller model's configurations: their mean and their minimum. Balanced accuracy weights the two accuracy classes equally, so it is not that count's hit rate. These per-level e-values are uncorrected. The two summary tests combine them separately for each provider: the minimum of a provider's mean column gives the every-level test's uncorrected e-value, and the arithmetic mean of its minimum column gives the some-level test's uncorrected e-value. Each combined value is then divided by K = 6, the e-Bonferroni correction across the within-provider model size family, which holds an every-level test and a some-level test for each of three providers. A value below 10,000 is shown in fixed notation, a value at or above 10,000 in scientific notation. The dagger (†) marks the authoring model, Claude Opus 4.6†.}\label{art:supplementary_table_s10}\\
\toprule
\textbf{Provider} & \textbf{Larger Model} & \textbf{Reasoning Effort} & \textbf{Balanced Accuracy} & \textbf{Valid Trials} & \textbf{Mean E-value Across Smaller Model Configurations} & \textbf{Minimum E-value Across Smaller Model Configurations} \\
\midrule\endfirsthead
\toprule
\textbf{Provider} & \textbf{Larger Model} & \textbf{Reasoning Effort} & \textbf{Balanced Accuracy} & \textbf{Valid Trials} & \textbf{Mean E-value Across Smaller Model Configurations} & \textbf{Minimum E-value Across Smaller Model Configurations} \\
\midrule\endhead
\endfoot
\bottomrule\endlastfoot
\multicolumn{7}{@{}>{\raggedright\arraybackslash}p{690.00pt}@{\hskip 3.00pt}}{\textbf{Primary Experiment}} \\*
\noalign{\vskip\aboverulesep\nointerlineskip}
\ReftCutInRule{7}\\*
\noalign{\kern\belowrulesep\nointerlineskip}
OpenAI & GPT-5.2 & none & 0.156 & 600 & 1.000 & 1.000 \\*
OpenAI & GPT-5.2 & low & 0.590 & 600 & 1.963e+25 & 1.000 \\*
OpenAI & GPT-5.2 & medium & 0.688 & 600 & 3.444e+34 & 0.809 \\*
OpenAI & GPT-5.2 & high & 0.700 & 600 & 2.149e+36 & 0.785 \\*
OpenAI & GPT-5.2 & xhigh & 0.942 & 600 & 6.968e+59 & 4.872 \\*
Anthropic & Claude Opus 4.6† & low & 0.979 & 600 & 5.095e+51 & 7.917e+09 \\*
Anthropic & Claude Opus 4.6† & medium & 1.000 & 600 & 1.639e+54 & 1.331e+13 \\*
Anthropic & Claude Opus 4.6† & high & 1.000 & 600 & 1.639e+54 & 1.331e+13 \\*
Anthropic & Claude Opus 4.6† & max & 0.980 & 600 & 5.716e+49 & 2.895e+10 \\*
Google & Gemini 3 Pro & low & 0.965 & 600 & 2.469e+13 & 1.576e+06 \\*
Google & Gemini 3 Pro & high & 0.941 & 600 & 3.528e+11 & 3426.895 \\
\addlinespace
\multicolumn{7}{@{}>{\raggedright\arraybackslash}p{690.00pt}@{\hskip 3.00pt}}{\textbf{Ablation Experiment}} \\*
\noalign{\vskip\aboverulesep\nointerlineskip}
\ReftCutInRule{7}\\*
\noalign{\kern\belowrulesep\nointerlineskip}
OpenAI & GPT-5.2 & none & 0.230 & 600 & 1.000 & 1.000 \\*
OpenAI & GPT-5.2 & low & 0.655 & 600 & 5.444e+26 & 1.000 \\*
OpenAI & GPT-5.2 & medium & 0.765 & 600 & 6.129e+38 & 1.000 \\*
OpenAI & GPT-5.2 & high & 0.815 & 600 & 5.262e+42 & 0.827 \\*
OpenAI & GPT-5.2 & xhigh & 0.953 & 600 & 1.384e+56 & 13.602 \\*
Anthropic & Claude Opus 4.6† & low & 1.000 & 600 & 1.617e+45 & 1.419e+12 \\*
Anthropic & Claude Opus 4.6† & medium & 1.000 & 600 & 1.617e+45 & 1.419e+12 \\*
Anthropic & Claude Opus 4.6† & high & 0.995 & 600 & 1.011e+44 & 4.633e+11 \\*
Anthropic & Claude Opus 4.6† & max & 0.970 & 600 & 3.856e+38 & 8.022e+07 \\*
Google & Gemini 3 Pro & low & 0.971 & 600 & 4.967e+09 & 3.510e+07 \\*
Google & Gemini 3 Pro & high & 0.950 & 600 & 1.898e+08 & 2.433e+06 \\
\end{longtable}}
\end{landscape}

%% file: items/per_condition_articulation_and_governing_rates_paired.tex
\begin{landscape}
{\footnotesize\setlength{\tabcolsep}{3.0pt}\ReftTableCell
\begin{longtable}{@{}>{\raggedright\arraybackslash}p{75.00pt}>{\raggedleft\arraybackslash}p{99.88pt}>{\raggedleft\arraybackslash}p{84.88pt}>{\raggedleft\arraybackslash}p{104.75pt}>{\raggedleft\arraybackslash}p{99.88pt}>{\raggedleft\arraybackslash}p{84.88pt}>{\raggedleft\arraybackslash}p{104.75pt}@{\hskip 3.00pt}}
\ReftCaption{Supplementary~Table~S11}{Per-Condition Articulation and Governing Rates}{Per-condition articulation rates and governing rates, grouped by experiment: each experiment's articulation rate, governing rate, and rate denominator form a contiguous column group. \textit{Articulation rate} is the percentage of trials whose verdict did not match ground truth in which the meta-evaluator expressed the operational interpretation; \textit{governing rate} is the percentage in which the operational interpretation determined the verdict. Both are always reported within a single condition. The largest articulation and governing rate differences between the experiments occurred on opaque trials whose verdict did not match ground truth. Opaque articulation rate was 60.78\% in the primary experiment (544 of 895 trials) and 68.72\% in the ablation experiment (468 of 681; +7.94pp); opaque governing rate was 2.57\% in the primary experiment (23 of 895) and 6.02\% in the ablation experiment (41 of 681; +3.45pp). Transparent articulation remained saturated at 100\% in both experiments (178 of 178 trials in the primary experiment and 139 of 139 in the ablation experiment); correct-draft governing remained near zero, at 0.69\% (9 of 1,310) in the primary experiment and 0.83\% (10 of 1,204; +0.14pp) in the ablation experiment.}\label{art:supplementary_table_s11}\\
\toprule
 & \multicolumn{3}{>{\centering\arraybackslash}p{301.50pt}}{\textbf{Primary Experiment}} & \multicolumn{3}{>{\centering\arraybackslash}p{301.50pt}}{\textbf{Ablation Experiment}} \\
\cmidrule(lr){2-4}\cmidrule(lr){5-7}
\textbf{Condition} & \textbf{Articulation Rate~(\%)} & \textbf{Governing Rate~(\%)} & \textbf{Wrong-Verdict Trial Count} & \textbf{Articulation Rate~(\%)} & \textbf{Governing Rate~(\%)} & \textbf{Wrong-Verdict Trial Count} \\
\midrule\endfirsthead
\toprule
 & \multicolumn{3}{>{\centering\arraybackslash}p{301.50pt}}{\textbf{Primary Experiment}} & \multicolumn{3}{>{\centering\arraybackslash}p{301.50pt}}{\textbf{Ablation Experiment}} \\
\cmidrule(lr){2-4}\cmidrule(lr){5-7}
\textbf{Condition} & \textbf{Articulation Rate~(\%)} & \textbf{Governing Rate~(\%)} & \textbf{Wrong-Verdict Trial Count} & \textbf{Articulation Rate~(\%)} & \textbf{Governing Rate~(\%)} & \textbf{Wrong-Verdict Trial Count} \\
\midrule\endhead
\endfoot
\bottomrule\endlastfoot
Correct-Draft & 33.6\% & 0.7\% & 1,310 & 35.1\% & 0.8\% & 1,204 \\
Transparent & 100.0\% & 1.1\% & 178 & 100.0\% & 2.2\% & 139 \\
Opaque & 60.8\% & 2.6\% & 895 & 68.7\% & 6.0\% & 681 \\
\end{longtable}}
\end{landscape}

%% file: items/per_condition_misattribution_rates_paired.tex
\begin{landscape}
{\footnotesize\setlength{\tabcolsep}{3.0pt}\ReftTableCell
\begin{longtable}{@{}>{\raggedright\arraybackslash}p{179.00pt}>{\raggedleft\arraybackslash}p{80.00pt}>{\raggedleft\arraybackslash}p{75.00pt}>{\raggedleft\arraybackslash}p{85.00pt}>{\raggedleft\arraybackslash}p{80.00pt}>{\raggedleft\arraybackslash}p{70.00pt}>{\raggedleft\arraybackslash}p{85.00pt}@{\hskip 3.00pt}}
\ReftCaption{Supplementary~Table~S12}{Per-Condition Misattribution Flag Rates}{Per-condition rates for the five misattribution flags in both experiments, each the percentage of trials whose verdict did not match ground truth on which the flag fired, paired with its count; the closing row gives the percentage that fired at least one flag. The five flags together act as a confound detector for the rationale analysis: each fires when the meta-evaluator's stated rationale misreports a value or claim the trial itself supplies. Misreporting was rare and almost entirely confined to the grader-claim flag on opaque stimuli, which fired on 20.8\% of opaque trials whose verdict did not match ground truth in the primary experiment and 24.8\% in the ablation experiment; the other four flags stayed at or below 2\% in every condition of both experiments. The any-flag opaque rate was 21.0\% in the primary experiment (188 flagged trials) and 25.3\% in the ablation experiment (172 flagged trials); the grader-claim flag accounted for 186 and 169 of those flagged trials, respectively. The probe audit reads this opaque misreport rate as an upper bound, since the grader-claim flag was the one flag an independent reader most often judged to have fired on faithful disagreements with the grader rather than genuine misreports.}\label{art:supplementary_table_s12}\\
\toprule
 & \multicolumn{3}{>{\centering\arraybackslash}p{252.00pt}}{\textbf{Primary Experiment}} & \multicolumn{3}{>{\centering\arraybackslash}p{247.00pt}}{\textbf{Ablation Experiment}} \\
\cmidrule(lr){2-4}\cmidrule(lr){5-7}
\textbf{Misattribution Flag} & \textbf{Correct-Draft Rate~(\%)} & \textbf{Transparent Rate~(\%)} & \textbf{Opaque Rate~(\%)} & \textbf{Correct-Draft Rate~(\%)} & \textbf{Transparent Rate~(\%)} & \textbf{Opaque Rate~(\%)} \\
\midrule\endfirsthead
\toprule
 & \multicolumn{3}{>{\centering\arraybackslash}p{252.00pt}}{\textbf{Primary Experiment}} & \multicolumn{3}{>{\centering\arraybackslash}p{247.00pt}}{\textbf{Ablation Experiment}} \\
\cmidrule(lr){2-4}\cmidrule(lr){5-7}
\textbf{Misattribution Flag} & \textbf{Correct-Draft Rate~(\%)} & \textbf{Transparent Rate~(\%)} & \textbf{Opaque Rate~(\%)} & \textbf{Correct-Draft Rate~(\%)} & \textbf{Transparent Rate~(\%)} & \textbf{Opaque Rate~(\%)} \\
\midrule\endhead
\endfoot
\bottomrule\endlastfoot
Misattributed Current Value & 0.0\% (0/1310) & 0.6\% (1/178) & 0.1\% (1/895) & 0.0\% (0/1204) & 0.0\% (0/139) & 0.0\% (0/681) \\
Misattributed Historical Value & 0.0\% (0/1310) & 0.0\% (0/178) & 0.0\% (0/895) & 0.0\% (0/1204) & 0.0\% (0/139) & 0.0\% (0/681) \\
Misattributed Proposed Value & 0.1\% (1/1310) & 0.0\% (0/178) & 0.0\% (0/895) & 0.0\% (0/1204) & 0.0\% (0/139) & 0.0\% (0/681) \\
Misattributed Assistant Fallback Citation & 1.5\% (19/1310) & 0.6\% (1/178) & 0.4\% (4/895) & 2.0\% (24/1204) & 0.7\% (1/139) & 1.2\% (8/681) \\
Misattributed Grader Claim & 0.1\% (1/1310) & 5.1\% (9/178) & 20.8\% (186/895) & 0.9\% (11/1204) & 2.9\% (4/139) & 24.8\% (169/681) \\
Any Misattribution Flag & 1.6\% (21/1310) & 5.6\% (10/178) & 21.0\% (188/895) & 2.7\% (32/1204) & 3.6\% (5/139) & 25.3\% (172/681) \\
\end{longtable}}
\end{landscape}

%% file: items/per_configuration_failure_mode_breakdown_paired.tex
\begin{landscape}
{\footnotesize\setlength{\tabcolsep}{3.0pt}\ReftTableCell
\begin{longtable}{@{}>{\raggedright\arraybackslash}p{130.50pt}>{\raggedright\arraybackslash}p{65.50pt}>{\raggedleft\arraybackslash}p{50.00pt}>{\raggedleft\arraybackslash}p{45.00pt}>{\raggedleft\arraybackslash}p{70.00pt}>{\raggedleft\arraybackslash}p{55.00pt}>{\raggedleft\arraybackslash}p{50.00pt}>{\raggedleft\arraybackslash}p{45.00pt}>{\raggedleft\arraybackslash}p{70.00pt}>{\raggedleft\arraybackslash}p{55.00pt}@{\hskip 3.00pt}}
\ReftCaption{Supplementary~Table~S13}{Per-Configuration Failure Mode Breakdown by Condition}{Per-configuration failure mode breakdown for the primary experiment and the ablation experiment, with one row per configuration and condition (correct-draft, transparent, opaque). Each row reports, grouped by experiment, the configuration's balanced accuracy (repeated across its three condition rows), the number of trials whose verdict did not match ground truth in that condition (the denominator for the row's rates), and, within that condition, the articulation rate (how often the meta-evaluator expressed the operational interpretation) and the governing rate (how often the operational interpretation determined the verdict). A rate is shown only where the configuration has at least one trial whose verdict did not match ground truth in that condition; a cell with none shows an em-dash rather than a zero. These rows are the source for two configuration-level contrasts: Claude Haiku 4.5 with reasoning off had the highest governing rate on opaque stimuli of any configuration in the primary experiment (15.17\%, alongside a 67.59\% opaque articulation rate) and the largest increase in that governing rate under the ablation experiment, to 34.45\%. By contrast, GPT-5.2 with reasoning effort set to none reached a comparable 68.75\% opaque articulation rate in the primary experiment yet had a 0\% opaque governing rate. The dagger (†) marks the authoring model, Claude Opus 4.6†.}\label{art:supplementary_table_s13}\\
\toprule
 &  & \multicolumn{4}{>{\centering\arraybackslash}p{238.00pt}}{\textbf{Primary Experiment}} & \multicolumn{4}{>{\centering\arraybackslash}p{238.00pt}}{\textbf{Ablation Experiment}} \\
\cmidrule(lr){3-6}\cmidrule(lr){7-10}
\textbf{Configuration} & \textbf{Condition} & \textbf{Balanced Accuracy} & \textbf{Wrong-Verdict Trial Count} & \textbf{Articulation Rate~(\%)} & \textbf{Governing Rate~(\%)} & \textbf{Balanced Accuracy} & \textbf{Wrong-Verdict Trial Count} & \textbf{Articulation Rate~(\%)} & \textbf{Governing Rate~(\%)} \\
\midrule\endfirsthead
\toprule
 &  & \multicolumn{4}{>{\centering\arraybackslash}p{238.00pt}}{\textbf{Primary Experiment}} & \multicolumn{4}{>{\centering\arraybackslash}p{238.00pt}}{\textbf{Ablation Experiment}} \\
\cmidrule(lr){3-6}\cmidrule(lr){7-10}
\textbf{Configuration} & \textbf{Condition} & \textbf{Balanced Accuracy} & \textbf{Wrong-Verdict Trial Count} & \textbf{Articulation Rate~(\%)} & \textbf{Governing Rate~(\%)} & \textbf{Balanced Accuracy} & \textbf{Wrong-Verdict Trial Count} & \textbf{Articulation Rate~(\%)} & \textbf{Governing Rate~(\%)} \\
\midrule\endhead
\endfoot
\bottomrule\endlastfoot
Claude Haiku 4.5 / off & Correct-Draft & 0.444 & 133 & 17.3\% & 0.0\% & 0.527 & 117 & 29.9\% & 0.9\% \\
Claude Haiku 4.5 / off & Transparent & 0.444 & 34 & 100.0\% & 5.9\% & 0.527 & 25 & 100.0\% & 12.0\% \\
Claude Haiku 4.5 / off & Opaque & 0.444 & 145 & 67.6\% & 15.2\% & 0.527 & 119 & 84.0\% & 34.5\% \\
Claude Haiku 4.5 / enabled & Correct-Draft & 0.855 & 58 & 48.3\% & 0.0\% & 0.873 & 51 & 70.6\% & 0.0\% \\
Claude Haiku 4.5 / enabled & Transparent & 0.855 & 0 & — & — & 0.873 & 0 & — & — \\
Claude Haiku 4.5 / enabled & Opaque & 0.855 & 0 & — & — & 0.873 & 0 & — & — \\
Claude Opus 4.6† / low & Correct-Draft & 0.979 & 8 & 62.5\% & 0.0\% & 1.000 & 0 & — & — \\
Claude Opus 4.6† / low & Transparent & 0.979 & 0 & — & — & 1.000 & 0 & — & — \\
Claude Opus 4.6† / low & Opaque & 0.979 & 1 & 0.0\% & 0.0\% & 1.000 & 0 & — & — \\
Claude Opus 4.6† / medium & Correct-Draft & 1.000 & 0 & — & — & 1.000 & 0 & — & — \\
Claude Opus 4.6† / medium & Transparent & 1.000 & 0 & — & — & 1.000 & 0 & — & — \\
Claude Opus 4.6† / medium & Opaque & 1.000 & 0 & — & — & 1.000 & 0 & — & — \\
Claude Opus 4.6† / high & Correct-Draft & 1.000 & 0 & — & — & 0.995 & 2 & 100.0\% & 0.0\% \\
Claude Opus 4.6† / high & Transparent & 1.000 & 0 & — & — & 0.995 & 0 & — & — \\
Claude Opus 4.6† / high & Opaque & 1.000 & 0 & — & — & 0.995 & 0 & — & — \\
Claude Opus 4.6† / max & Correct-Draft & 0.980 & 8 & 87.5\% & 0.0\% & 0.970 & 12 & 91.7\% & 0.0\% \\
Claude Opus 4.6† / max & Transparent & 0.980 & 0 & — & — & 0.970 & 0 & — & — \\
Claude Opus 4.6† / max & Opaque & 0.980 & 0 & — & — & 0.970 & 0 & — & — \\
Gemini 3 Flash / minimal & Correct-Draft & 0.824 & 34 & 8.8\% & 0.0\% & 0.866 & 28 & 14.3\% & 0.0\% \\
Gemini 3 Flash / minimal & Transparent & 0.824 & 0 & — & — & 0.866 & 0 & — & — \\
Gemini 3 Flash / minimal & Opaque & 0.824 & 73 & 32.9\% & 0.0\% & 0.866 & 51 & 33.3\% & 0.0\% \\
Gemini 3 Flash / low & Correct-Draft & 0.849 & 37 & 18.9\% & 0.0\% & 0.870 & 38 & 5.3\% & 0.0\% \\
Gemini 3 Flash / low & Transparent & 0.849 & 0 & — & — & 0.870 & 0 & — & — \\
Gemini 3 Flash / low & Opaque & 0.849 & 47 & 42.6\% & 0.0\% & 0.870 & 28 & 32.1\% & 0.0\% \\
Gemini 3 Flash / medium & Correct-Draft & 0.894 & 41 & 58.5\% & 0.0\% & 0.883 & 45 & 44.4\% & 0.0\% \\
Gemini 3 Flash / medium & Transparent & 0.894 & 0 & — & — & 0.883 & 0 & — & — \\
Gemini 3 Flash / medium & Opaque & 0.894 & 3 & 66.7\% & 0.0\% & 0.883 & 4 & 25.0\% & 0.0\% \\
Gemini 3 Flash / high & Correct-Draft & 0.884 & 44 & 68.2\% & 0.0\% & 0.857 & 57 & 57.9\% & 0.0\% \\
Gemini 3 Flash / high & Transparent & 0.884 & 0 & — & — & 0.857 & 0 & — & — \\
Gemini 3 Flash / high & Opaque & 0.884 & 5 & 60.0\% & 0.0\% & 0.857 & 0 & — & — \\
Gemini 3 Pro / low & Correct-Draft & 0.965 & 10 & 80.0\% & 0.0\% & 0.971 & 6 & 100.0\% & 0.0\% \\
Gemini 3 Pro / low & Transparent & 0.965 & 0 & — & — & 0.971 & 0 & — & — \\
Gemini 3 Pro / low & Opaque & 0.965 & 8 & 100.0\% & 0.0\% & 0.971 & 11 & 100.0\% & 0.0\% \\
Gemini 3 Pro / high & Correct-Draft & 0.941 & 21 & 71.4\% & 0.0\% & 0.950 & 20 & 55.0\% & 0.0\% \\
Gemini 3 Pro / high & Transparent & 0.941 & 0 & — & — & 0.950 & 0 & — & — \\
Gemini 3 Pro / high & Opaque & 0.941 & 5 & 100.0\% & 0.0\% & 0.950 & 0 & — & — \\
GPT-5-mini / minimal & Correct-Draft & 0.307 & 197 & 14.2\% & 0.0\% & 0.354 & 199 & 22.1\% & 0.0\% \\
GPT-5-mini / minimal & Transparent & 0.307 & 3 & 100.0\% & 0.0\% & 0.354 & 0 & — & — \\
GPT-5-mini / minimal & Opaque & 0.307 & 157 & 52.2\% & 0.0\% & 0.354 & 119 & 51.3\% & 0.0\% \\
GPT-5-mini / low & Correct-Draft & 0.675 & 108 & 13.9\% & 7.4\% & 0.705 & 105 & 16.2\% & 7.6\% \\
GPT-5-mini / low & Transparent & 0.675 & 0 & — & — & 0.705 & 0 & — & — \\
GPT-5-mini / low & Opaque & 0.675 & 44 & 11.4\% & 0.0\% & 0.705 & 26 & 15.4\% & 0.0\% \\
GPT-5-mini / medium & Correct-Draft & 0.809 & 69 & 8.7\% & 1.4\% & 0.831 & 62 & 21.0\% & 1.6\% \\
GPT-5-mini / medium & Transparent & 0.809 & 0 & — & — & 0.831 & 0 & — & — \\
GPT-5-mini / medium & Opaque & 0.809 & 15 & 6.7\% & 0.0\% & 0.831 & 11 & 0.0\% & 0.0\% \\
GPT-5-mini / high & Correct-Draft & 0.903 & 38 & 10.5\% & 0.0\% & 0.895 & 40 & 30.0\% & 0.0\% \\
GPT-5-mini / high & Transparent & 0.903 & 0 & — & — & 0.895 & 0 & — & — \\
GPT-5-mini / high & Opaque & 0.903 & 2 & 0.0\% & 0.0\% & 0.895 & 4 & 0.0\% & 0.0\% \\
GPT-5.2 / none & Correct-Draft & 0.156 & 200 & 43.5\% & 0.0\% & 0.230 & 184 & 34.8\% & 0.0\% \\
GPT-5.2 / none & Transparent & 0.156 & 115 & 100.0\% & 0.0\% & 0.230 & 104 & 100.0\% & 0.0\% \\
GPT-5.2 / none & Opaque & 0.156 & 160 & 68.8\% & 0.0\% & 0.230 & 144 & 84.0\% & 0.0\% \\
GPT-5.2 / low & Correct-Draft & 0.590 & 120 & 40.0\% & 0.0\% & 0.655 & 100 & 34.0\% & 0.0\% \\
GPT-5.2 / low & Transparent & 0.590 & 15 & 100.0\% & 0.0\% & 0.655 & 10 & 100.0\% & 0.0\% \\
GPT-5.2 / low & Opaque & 0.590 & 73 & 65.8\% & 0.0\% & 0.655 & 66 & 83.3\% & 0.0\% \\
GPT-5.2 / medium & Correct-Draft & 0.688 & 87 & 52.9\% & 0.0\% & 0.765 & 70 & 54.3\% & 0.0\% \\
GPT-5.2 / medium & Transparent & 0.688 & 6 & 100.0\% & 0.0\% & 0.765 & 0 & — & — \\
GPT-5.2 / medium & Opaque & 0.688 & 70 & 84.3\% & 0.0\% & 0.765 & 48 & 85.4\% & 0.0\% \\
GPT-5.2 / high & Correct-Draft & 0.700 & 85 & 61.2\% & 0.0\% & 0.815 & 55 & 67.3\% & 0.0\% \\
GPT-5.2 / high & Transparent & 0.700 & 4 & 100.0\% & 0.0\% & 0.815 & 0 & — & — \\
GPT-5.2 / high & Opaque & 0.700 & 66 & 87.9\% & 1.5\% & 0.815 & 38 & 100.0\% & 0.0\% \\
GPT-5.2 / xhigh & Correct-Draft & 0.942 & 12 & 33.3\% & 0.0\% & 0.953 & 13 & 30.8\% & 0.0\% \\
GPT-5.2 / xhigh & Transparent & 0.942 & 1 & 100.0\% & 0.0\% & 0.953 & 0 & — & — \\
GPT-5.2 / xhigh & Opaque & 0.942 & 21 & 100.0\% & 0.0\% & 0.953 & 12 & 83.3\% & 0.0\% \\
\end{longtable}}
\end{landscape}

%% file: items/base_example_inventory.tex
\begin{landscape}
{\footnotesize\setlength{\tabcolsep}{3.0pt}\ReftTableCell
\begin{longtable}{@{}>{\raggedright\arraybackslash}p{95.00pt}>{\raggedright\arraybackslash}p{75.00pt}>{\raggedright\arraybackslash}p{140.00pt}>{\raggedright\arraybackslash}p{110.00pt}>{\raggedright\arraybackslash}p{125.00pt}>{\raggedright\arraybackslash}p{115.00pt}@{\hskip 3.00pt}}
\ReftCaption{Supplementary~Table~S14}{Base Example Inventory}{The 10 base examples and their stimulus parameters: risk level, the noun naming the value under test, and the three values each example specifies (\textit{current}, \textit{historical}, \textit{proposed}). Under the field label interpretation, ``previous'' resolves to the \textit{historical} value; under the operational interpretation, it resolves to the \textit{current} value, the value in effect before the \textit{proposed} change. Each base example is rendered in three conditions (correct-draft, transparent, opaque) to produce the primary experiment's 30 stimuli; the ablation experiment uses the same 30 stimuli with the grader's four error classification fields removed.}\label{art:supplementary_table_s14}\\
\toprule
\textbf{Example} & \textbf{Risk} & \textbf{Value Under Test} & \textbf{Current Value} & \textbf{Historical Value} & \textbf{Proposed Value} \\
\midrule\endfirsthead
\toprule
\textbf{Example} & \textbf{Risk} & \textbf{Value Under Test} & \textbf{Current Value} & \textbf{Historical Value} & \textbf{Proposed Value} \\
\midrule\endhead
\endfoot
\bottomrule\endlastfoot
example\_01 & Low & Temperature Setting & 72°F & 68°F & 74°F \\
example\_02 & Medium & Version & v2.3.1 & v2.2.8 & v2.4.0 \\
example\_03 & Low & Manager & David Chen & Sarah Martinez & Marcus Webb \\
example\_04 & High & Dose & 20mg & 15mg & 25mg \\
example\_05 & Low & Plan & \$49/mo & \$29/mo & \$99/mo \\
example\_06 & High & IP Address & 192.168.1.100 & 192.168.1.50 & 10.0.5.25 \\
example\_07 & Medium & Salary & \$85,000 & \$78,000 & \$105,000 \\
example\_08 & High & Schema Version & 3.2.0 & 3.1.4 & 4.0.0 \\
example\_09 & Low & Departure Time & 2:30 PM & 10:00 AM & 6:45 PM \\
example\_10 & High & Interest Rate & 6.5\% & 5.8\% & 7.2\% \\
\end{longtable}}
\end{landscape}

%% file: items/pairwise_comparison_betting_cap_sensitivity_paired.tex
\begin{landscape}
{\footnotesize\setlength{\tabcolsep}{3.0pt}\ReftTableCell
\begin{longtable}{@{}>{\raggedleft\arraybackslash}p{125.50pt}>{\raggedleft\arraybackslash}p{185.50pt}>{\raggedleft\arraybackslash}p{155.50pt}>{\raggedleft\arraybackslash}p{205.50pt}@{\hskip 3.00pt}}
\ReftCaption{Supplementary~Table~S15}{Pairwise Comparison Betting Cap Sensitivity}{Sensitivity of the pairwise condition comparison significance counts to the betting cap, at each of five caps from 0.25 to 0.75, for the primary experiment and the ablation experiment. The dagger (†) marks the authoring model, Claude Opus 4.6†: it helped author the stimulus components, so its results may reflect familiarity with its own output patterns rather than the capability being measured. The cap bounds the magnitude of the bet each comparison places on the next round of paired differences; it is an interior default not tuned to the results. Each row reports how many of that experiment's 63 pairwise condition comparisons reach significance at that cap under the closed test, which requires a comparison's own e-value and its configuration's global null e-value both to reach the same threshold, and how many comparisons change significance status relative to the reported cap of 0.50. The reported cap is the row the others are compared against and so has no change count of its own; an em-dash is shown there instead of a zero. Counts range from 36 to 42 in the primary experiment and from 35 to 41 in the ablation experiment, against 38 and 41 at the reported cap. Every comparison that changes status across the whole grid belongs to one of four configurations: Claude Opus 4.6† at max reasoning effort, Gemini 3 Pro at low reasoning effort, GPT-5-mini at medium reasoning effort, and GPT-5.2 at xhigh reasoning effort. The transparent-versus-opaque comparison is significant at every cap in both experiments for nine configurations, among them Claude Haiku 4.5 with reasoning off and GPT-5.2 at low reasoning effort.}\label{art:supplementary_table_s15}\\
\toprule
\textbf{Betting Cap} & \textbf{Significant Comparisons} & \textbf{Total Comparisons} & \textbf{Comparisons Changing Status} \\
\midrule\endfirsthead
\toprule
\textbf{Betting Cap} & \textbf{Significant Comparisons} & \textbf{Total Comparisons} & \textbf{Comparisons Changing Status} \\
\midrule\endhead
\endfoot
\bottomrule\endlastfoot
\multicolumn{4}{@{}>{\raggedright\arraybackslash}p{690.00pt}@{\hskip 3.00pt}}{\textbf{Primary Experiment}} \\*
\noalign{\vskip\aboverulesep\nointerlineskip}
\ReftCutInRule{4}\\*
\noalign{\kern\belowrulesep\nointerlineskip}
0.25 & 36 & 63 & 2 \\*
0.40 & 38 & 63 & 0 \\*
0.50 & 38 & 63 & — \\*
0.60 & 40 & 63 & 2 \\*
0.75 & 42 & 63 & 4 \\
\addlinespace
\multicolumn{4}{@{}>{\raggedright\arraybackslash}p{690.00pt}@{\hskip 3.00pt}}{\textbf{Ablation Experiment}} \\*
\noalign{\vskip\aboverulesep\nointerlineskip}
\ReftCutInRule{4}\\*
\noalign{\kern\belowrulesep\nointerlineskip}
0.25 & 35 & 63 & 6 \\*
0.40 & 40 & 63 & 1 \\*
0.50 & 41 & 63 & — \\*
0.60 & 41 & 63 & 0 \\*
0.75 & 41 & 63 & 0 \\
\end{longtable}}
\end{landscape}

%% file: items/probe_auditor_alignment_paired.tex
\begin{landscape}
{\footnotesize\setlength{\tabcolsep}{3.0pt}\ReftTableCell
\begin{longtable}{@{}>{\raggedright\arraybackslash}p{110.60pt}>{\raggedleft\arraybackslash}p{132.60pt}>{\raggedleft\arraybackslash}p{102.60pt}>{\raggedleft\arraybackslash}p{127.60pt}>{\raggedleft\arraybackslash}p{192.60pt}@{\hskip 3.00pt}}
\ReftCaption{Supplementary~Table~S16}{Probe-Auditor Alignment by Flag and Condition}{Probe-auditor alignment for each of the probe's seven flags, broken out by condition (correct-draft, transparent, opaque) for the primary experiment and the ablation experiment, from the two-phase blind-then-informed audit. Under that protocol, an independent auditor sub-agent classified each audited trial on all seven flags before seeing the probe's output and then, once that output was revealed, judged whether the probe had been correct; a trial counts as aligned on a flag when this second judgment upheld the probe on that flag. Where a bound is shown, it is a 95\% one-sided anytime-valid confidence sequence lower bound on the alignment rate in that row. Bounds appear only for the articulated operational interpretation flag and the misattributed grader-claim flag. For the remaining five flags, the audit deliberately over-sampled trials carrying their rare positive classifications, so a pooled bound would not estimate a population alignment rate and an em-dash is shown instead. That over-sampling also reached the articulation flag indirectly, because a trial drawn for a rare positive governing classification always carries a positive articulation classification; the printed bounds therefore describe alignment on the audited trials rather than on the population the audit drew from. The probe and the auditor diverged materially on one flag only: alignment was at or above 93.8\% for every other combination of flag and condition in both experiments, but on the misattributed grader-claim flag on opaque stimuli it was 80.0\% (72 of 90 audited trials) in the primary experiment and 82.7\% (67 of 81) in the ablation experiment.}\label{art:supplementary_table_s16}\\
\toprule
\textbf{Condition} & \textbf{Audited Trial Count} & \textbf{Aligned Count} & \textbf{Alignment Rate~(\%)} & \textbf{Confidence Sequence Lower Bound} \\
\midrule\endfirsthead
\toprule
\textbf{Condition} & \textbf{Audited Trial Count} & \textbf{Aligned Count} & \textbf{Alignment Rate~(\%)} & \textbf{Confidence Sequence Lower Bound} \\
\midrule\endhead
\endfoot
\bottomrule\endlastfoot
\multicolumn{5}{@{}>{\raggedright\arraybackslash}p{690.00pt}@{\hskip 3.00pt}}{\textbf{Primary Experiment}} \\*
\noalign{\vskip\aboverulesep\nointerlineskip}
\ReftCutInRule{5}\\*
\noalign{\kern\belowrulesep\nointerlineskip}
\multicolumn{5}{@{}>{\raggedright\arraybackslash}p{690.00pt}@{\hskip 3.00pt}}{\textit{Misattributed Current Value}} \\*
\hspace*{8.0pt}Correct-Draft & 132 & 132 & 100.0\% & — \\*
\hspace*{8.0pt}Transparent & 18 & 18 & 100.0\% & — \\*
\hspace*{8.0pt}Opaque & 90 & 90 & 100.0\% & — \\*
\multicolumn{5}{@{}>{\raggedright\arraybackslash}p{690.00pt}@{\hskip 3.00pt}}{\textit{Misattributed Historical Value}} \\*
\hspace*{8.0pt}Correct-Draft & 132 & 132 & 100.0\% & — \\*
\hspace*{8.0pt}Transparent & 18 & 18 & 100.0\% & — \\*
\hspace*{8.0pt}Opaque & 90 & 90 & 100.0\% & — \\*
\multicolumn{5}{@{}>{\raggedright\arraybackslash}p{690.00pt}@{\hskip 3.00pt}}{\textit{Misattributed Proposed Value}} \\*
\hspace*{8.0pt}Correct-Draft & 132 & 132 & 100.0\% & — \\*
\hspace*{8.0pt}Transparent & 18 & 18 & 100.0\% & — \\*
\hspace*{8.0pt}Opaque & 90 & 90 & 100.0\% & — \\*
\multicolumn{5}{@{}>{\raggedright\arraybackslash}p{690.00pt}@{\hskip 3.00pt}}{\textit{Misattributed Assistant Fallback Citation}} \\*
\hspace*{8.0pt}Correct-Draft & 132 & 128 & 97.0\% & — \\*
\hspace*{8.0pt}Transparent & 18 & 18 & 100.0\% & — \\*
\hspace*{8.0pt}Opaque & 90 & 88 & 97.8\% & — \\*
\multicolumn{5}{@{}>{\raggedright\arraybackslash}p{690.00pt}@{\hskip 3.00pt}}{\textit{Misattributed Grader Claim}} \\*
\hspace*{8.0pt}Correct-Draft & 132 & 132 & 100.0\% & 0.972 \\*
\hspace*{8.0pt}Transparent & 18 & 17 & 94.4\% & 0.684 \\*
\hspace*{8.0pt}Opaque & 90 & 72 & 80.0\% & 0.655 \\*
\multicolumn{5}{@{}>{\raggedright\arraybackslash}p{690.00pt}@{\hskip 3.00pt}}{\textit{Articulated Operational Interpretation}} \\*
\hspace*{8.0pt}Correct-Draft & 132 & 129 & 97.7\% & 0.920 \\*
\hspace*{8.0pt}Transparent & 18 & 18 & 100.0\% & 0.811 \\*
\hspace*{8.0pt}Opaque & 90 & 88 & 97.8\% & 0.905 \\*
\multicolumn{5}{@{}>{\raggedright\arraybackslash}p{690.00pt}@{\hskip 3.00pt}}{\textit{Operational Interpretation Governed Judgment}} \\*
\hspace*{8.0pt}Correct-Draft & 132 & 132 & 100.0\% & — \\*
\hspace*{8.0pt}Transparent & 18 & 18 & 100.0\% & — \\*
\hspace*{8.0pt}Opaque & 90 & 87 & 96.7\% & — \\
\addlinespace
\multicolumn{5}{@{}>{\raggedright\arraybackslash}p{690.00pt}@{\hskip 3.00pt}}{\textbf{Ablation Experiment}} \\*
\noalign{\vskip\aboverulesep\nointerlineskip}
\ReftCutInRule{5}\\*
\noalign{\kern\belowrulesep\nointerlineskip}
\multicolumn{5}{@{}>{\raggedright\arraybackslash}p{690.00pt}@{\hskip 3.00pt}}{\textit{Misattributed Current Value}} \\*
\hspace*{8.0pt}Correct-Draft & 143 & 143 & 100.0\% & — \\*
\hspace*{8.0pt}Transparent & 16 & 16 & 100.0\% & — \\*
\hspace*{8.0pt}Opaque & 81 & 81 & 100.0\% & — \\*
\multicolumn{5}{@{}>{\raggedright\arraybackslash}p{690.00pt}@{\hskip 3.00pt}}{\textit{Misattributed Historical Value}} \\*
\hspace*{8.0pt}Correct-Draft & 143 & 143 & 100.0\% & — \\*
\hspace*{8.0pt}Transparent & 16 & 16 & 100.0\% & — \\*
\hspace*{8.0pt}Opaque & 81 & 81 & 100.0\% & — \\*
\multicolumn{5}{@{}>{\raggedright\arraybackslash}p{690.00pt}@{\hskip 3.00pt}}{\textit{Misattributed Proposed Value}} \\*
\hspace*{8.0pt}Correct-Draft & 143 & 143 & 100.0\% & — \\*
\hspace*{8.0pt}Transparent & 16 & 16 & 100.0\% & — \\*
\hspace*{8.0pt}Opaque & 81 & 81 & 100.0\% & — \\*
\multicolumn{5}{@{}>{\raggedright\arraybackslash}p{690.00pt}@{\hskip 3.00pt}}{\textit{Misattributed Assistant Fallback Citation}} \\*
\hspace*{8.0pt}Correct-Draft & 143 & 136 & 95.1\% & — \\*
\hspace*{8.0pt}Transparent & 16 & 16 & 100.0\% & — \\*
\hspace*{8.0pt}Opaque & 81 & 81 & 100.0\% & — \\*
\multicolumn{5}{@{}>{\raggedright\arraybackslash}p{690.00pt}@{\hskip 3.00pt}}{\textit{Misattributed Grader Claim}} \\*
\hspace*{8.0pt}Correct-Draft & 143 & 143 & 100.0\% & 0.974 \\*
\hspace*{8.0pt}Transparent & 16 & 15 & 93.8\% & 0.651 \\*
\hspace*{8.0pt}Opaque & 81 & 67 & 82.7\% & 0.681 \\*
\multicolumn{5}{@{}>{\raggedright\arraybackslash}p{690.00pt}@{\hskip 3.00pt}}{\textit{Articulated Operational Interpretation}} \\*
\hspace*{8.0pt}Correct-Draft & 143 & 141 & 98.6\% & 0.940 \\*
\hspace*{8.0pt}Transparent & 16 & 16 & 100.0\% & 0.790 \\*
\hspace*{8.0pt}Opaque & 81 & 76 & 93.8\% & 0.832 \\*
\multicolumn{5}{@{}>{\raggedright\arraybackslash}p{690.00pt}@{\hskip 3.00pt}}{\textit{Operational Interpretation Governed Judgment}} \\*
\hspace*{8.0pt}Correct-Draft & 143 & 142 & 99.3\% & — \\*
\hspace*{8.0pt}Transparent & 16 & 16 & 100.0\% & — \\*
\hspace*{8.0pt}Opaque & 81 & 79 & 97.5\% & — \\
\end{longtable}}
\end{landscape}

%% file: items/probe_auditor_disagreement_transitions_paired.tex
\begin{landscape}
{\footnotesize\setlength{\tabcolsep}{3.0pt}\ReftTableCell
\begin{longtable}{@{}>{\raggedright\arraybackslash}p{115.60pt}>{\raggedleft\arraybackslash}p{122.60pt}>{\raggedleft\arraybackslash}p{142.60pt}>{\raggedleft\arraybackslash}p{137.60pt}>{\raggedleft\arraybackslash}p{147.60pt}@{\hskip 3.00pt}}
\ReftCaption{Supplementary~Table~S17}{Probe-Auditor Disagreement Transitions by Flag and Condition}{The four two-phase outcomes of every audited trial, for each of the probe's seven flags, broken out by condition (correct-draft, transparent, opaque) for the primary experiment and the ablation experiment, from the two-phase blind-then-informed audit. The four columns partition each row's audited trials: stable agreement (the auditor's Phase 1 reading matched the probe and the Phase 2 judgment confirmed it), shifted toward probe (an independent disagreement that became agreement after the probe's output was revealed), stable disagreement (an independent disagreement the auditor confirmed after viewing the probe's output, the strongest evidence of probe error), and shifted against probe (an independent agreement that reversed away from the probe). No audited trial in either experiment shifted against the probe.}\label{art:supplementary_table_s17}\\
\toprule
\textbf{Condition} & \textbf{Stable Agreement} & \textbf{Shifted Toward Probe} & \textbf{Stable Disagreement} & \textbf{Shifted Against Probe} \\
\midrule\endfirsthead
\toprule
\textbf{Condition} & \textbf{Stable Agreement} & \textbf{Shifted Toward Probe} & \textbf{Stable Disagreement} & \textbf{Shifted Against Probe} \\
\midrule\endhead
\endfoot
\bottomrule\endlastfoot
\multicolumn{5}{@{}>{\raggedright\arraybackslash}p{690.00pt}@{\hskip 3.00pt}}{\textbf{Primary Experiment}} \\*
\noalign{\vskip\aboverulesep\nointerlineskip}
\ReftCutInRule{5}\\*
\noalign{\kern\belowrulesep\nointerlineskip}
\multicolumn{5}{@{}>{\raggedright\arraybackslash}p{690.00pt}@{\hskip 3.00pt}}{\textit{Misattributed Current Value}} \\*
\hspace*{8.0pt}Correct-Draft & 132 & 0 & 0 & 0 \\*
\hspace*{8.0pt}Transparent & 17 & 1 & 0 & 0 \\*
\hspace*{8.0pt}Opaque & 89 & 1 & 0 & 0 \\*
\multicolumn{5}{@{}>{\raggedright\arraybackslash}p{690.00pt}@{\hskip 3.00pt}}{\textit{Misattributed Historical Value}} \\*
\hspace*{8.0pt}Correct-Draft & 132 & 0 & 0 & 0 \\*
\hspace*{8.0pt}Transparent & 18 & 0 & 0 & 0 \\*
\hspace*{8.0pt}Opaque & 90 & 0 & 0 & 0 \\*
\multicolumn{5}{@{}>{\raggedright\arraybackslash}p{690.00pt}@{\hskip 3.00pt}}{\textit{Misattributed Proposed Value}} \\*
\hspace*{8.0pt}Correct-Draft & 131 & 1 & 0 & 0 \\*
\hspace*{8.0pt}Transparent & 18 & 0 & 0 & 0 \\*
\hspace*{8.0pt}Opaque & 90 & 0 & 0 & 0 \\*
\multicolumn{5}{@{}>{\raggedright\arraybackslash}p{690.00pt}@{\hskip 3.00pt}}{\textit{Misattributed Assistant Fallback Citation}} \\*
\hspace*{8.0pt}Correct-Draft & 127 & 1 & 4 & 0 \\*
\hspace*{8.0pt}Transparent & 18 & 0 & 0 & 0 \\*
\hspace*{8.0pt}Opaque & 88 & 0 & 2 & 0 \\*
\multicolumn{5}{@{}>{\raggedright\arraybackslash}p{690.00pt}@{\hskip 3.00pt}}{\textit{Misattributed Grader Claim}} \\*
\hspace*{8.0pt}Correct-Draft & 132 & 0 & 0 & 0 \\*
\hspace*{8.0pt}Transparent & 17 & 0 & 1 & 0 \\*
\hspace*{8.0pt}Opaque & 72 & 0 & 18 & 0 \\*
\multicolumn{5}{@{}>{\raggedright\arraybackslash}p{690.00pt}@{\hskip 3.00pt}}{\textit{Articulated Operational Interpretation}} \\*
\hspace*{8.0pt}Correct-Draft & 128 & 1 & 3 & 0 \\*
\hspace*{8.0pt}Transparent & 18 & 0 & 0 & 0 \\*
\hspace*{8.0pt}Opaque & 87 & 1 & 2 & 0 \\*
\multicolumn{5}{@{}>{\raggedright\arraybackslash}p{690.00pt}@{\hskip 3.00pt}}{\textit{Operational Interpretation Governed Judgment}} \\*
\hspace*{8.0pt}Correct-Draft & 132 & 0 & 0 & 0 \\*
\hspace*{8.0pt}Transparent & 18 & 0 & 0 & 0 \\*
\hspace*{8.0pt}Opaque & 86 & 1 & 3 & 0 \\
\addlinespace
\multicolumn{5}{@{}>{\raggedright\arraybackslash}p{690.00pt}@{\hskip 3.00pt}}{\textbf{Ablation Experiment}} \\*
\noalign{\vskip\aboverulesep\nointerlineskip}
\ReftCutInRule{5}\\*
\noalign{\kern\belowrulesep\nointerlineskip}
\multicolumn{5}{@{}>{\raggedright\arraybackslash}p{690.00pt}@{\hskip 3.00pt}}{\textit{Misattributed Current Value}} \\*
\hspace*{8.0pt}Correct-Draft & 143 & 0 & 0 & 0 \\*
\hspace*{8.0pt}Transparent & 16 & 0 & 0 & 0 \\*
\hspace*{8.0pt}Opaque & 81 & 0 & 0 & 0 \\*
\multicolumn{5}{@{}>{\raggedright\arraybackslash}p{690.00pt}@{\hskip 3.00pt}}{\textit{Misattributed Historical Value}} \\*
\hspace*{8.0pt}Correct-Draft & 143 & 0 & 0 & 0 \\*
\hspace*{8.0pt}Transparent & 16 & 0 & 0 & 0 \\*
\hspace*{8.0pt}Opaque & 81 & 0 & 0 & 0 \\*
\multicolumn{5}{@{}>{\raggedright\arraybackslash}p{690.00pt}@{\hskip 3.00pt}}{\textit{Misattributed Proposed Value}} \\*
\hspace*{8.0pt}Correct-Draft & 143 & 0 & 0 & 0 \\*
\hspace*{8.0pt}Transparent & 16 & 0 & 0 & 0 \\*
\hspace*{8.0pt}Opaque & 81 & 0 & 0 & 0 \\*
\multicolumn{5}{@{}>{\raggedright\arraybackslash}p{690.00pt}@{\hskip 3.00pt}}{\textit{Misattributed Assistant Fallback Citation}} \\*
\hspace*{8.0pt}Correct-Draft & 134 & 2 & 7 & 0 \\*
\hspace*{8.0pt}Transparent & 15 & 1 & 0 & 0 \\*
\hspace*{8.0pt}Opaque & 80 & 1 & 0 & 0 \\*
\multicolumn{5}{@{}>{\raggedright\arraybackslash}p{690.00pt}@{\hskip 3.00pt}}{\textit{Misattributed Grader Claim}} \\*
\hspace*{8.0pt}Correct-Draft & 143 & 0 & 0 & 0 \\*
\hspace*{8.0pt}Transparent & 15 & 0 & 1 & 0 \\*
\hspace*{8.0pt}Opaque & 65 & 2 & 14 & 0 \\*
\multicolumn{5}{@{}>{\raggedright\arraybackslash}p{690.00pt}@{\hskip 3.00pt}}{\textit{Articulated Operational Interpretation}} \\*
\hspace*{8.0pt}Correct-Draft & 139 & 2 & 2 & 0 \\*
\hspace*{8.0pt}Transparent & 16 & 0 & 0 & 0 \\*
\hspace*{8.0pt}Opaque & 74 & 2 & 5 & 0 \\*
\multicolumn{5}{@{}>{\raggedright\arraybackslash}p{690.00pt}@{\hskip 3.00pt}}{\textit{Operational Interpretation Governed Judgment}} \\*
\hspace*{8.0pt}Correct-Draft & 142 & 0 & 1 & 0 \\*
\hspace*{8.0pt}Transparent & 16 & 0 & 0 & 0 \\*
\hspace*{8.0pt}Opaque & 79 & 0 & 2 & 0 \\
\end{longtable}}
\end{landscape}

%% file: items/cascade_diagram.tex
\begin{figure}[tbp]
\centering
\includegraphics[alt={Context Cascade Reconstruction},width=\textwidth,height=0.88\textheight,keepaspectratio]{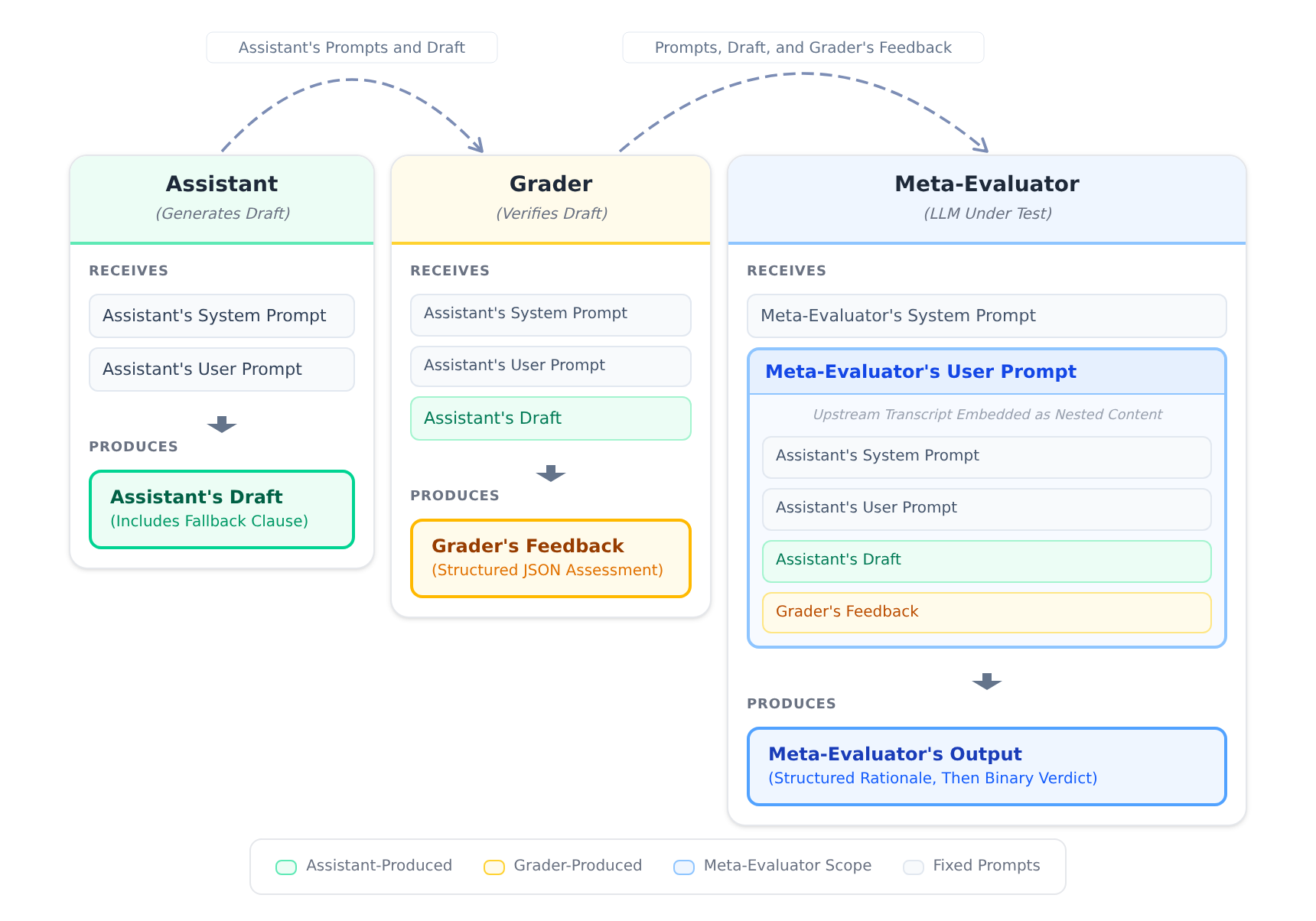}
\ReftCaption{Supplementary~Figure~S1}{Context Cascade Reconstruction}{Schematic of the context cascade reconstruction that each stimulus presents to the meta-evaluator. Each role receives all upstream context cumulatively: the assistant receives a system prompt and user prompt, the grader receives those same prompts together with the assistant's draft, and the meta-evaluator receives its own system prompt plus a user prompt whose body contains the entire upstream transcript (the assistant's prompts, the assistant's draft, and the grader's feedback) as nested content. The nesting is load-bearing: because the upstream transcript is embedded inside the meta-evaluator's user prompt rather than delivered as separate inputs, every field label and every relative temporal expression appears at two levels of the prompt hierarchy, and the meta-evaluator must track which level governs each referent.}
\label{art:supplementary_figure_s1}
\end{figure}

%% file: items/cell_resolution_status_paired.tex
\begin{figure}[tbp]
\centering
\includegraphics[alt={Cell Resolution Status by Configuration},width=\textwidth,height=0.88\textheight,keepaspectratio]{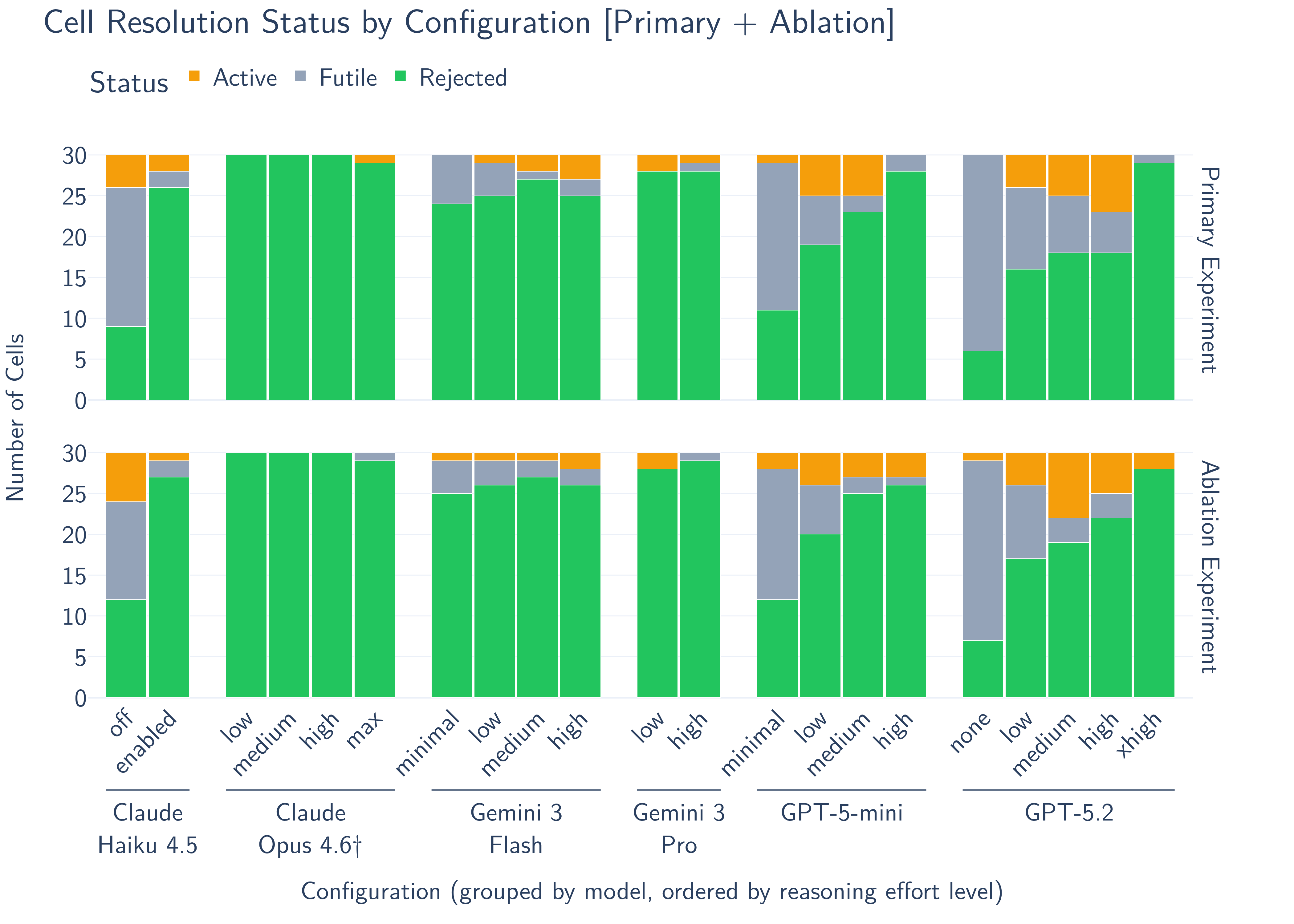}
\ReftCaption{Supplementary~Figure~S2}{Cell Resolution Status by Configuration}{Stacked bar count of cells reaching each resolution state, namely \textit{rejected} (e-value ≥ 20), \textit{futile} (e-value ≤ 0.05), or \textit{active} (neither threshold crossed), for every configuration, in one labeled row for the primary experiment and one for the ablation experiment. The dagger (†) marks the authoring model, Claude Opus 4.6†: it helped author the stimulus components, so its results may reflect familiarity with its own output patterns rather than the capability being measured. Each configuration contributes 30 cells, one per stimulus, for 630 cells per experiment. A single iteration of 20 trials per cell resolved 587 of the primary experiment's 630 cells (93.2\%) and 584 of the ablation experiment's 630 cells (92.7\%); data collection stopped there, so the 43 cells left active in the primary experiment and the 46 left active in the ablation experiment were never extended with a further batch. In both experiments the two lowest-scoring configurations, GPT-5-mini at minimal reasoning effort and GPT-5.2 with reasoning effort set to none, resolved mainly by futility rather than by rejection.}
\label{art:supplementary_figure_s2}
\end{figure}

%% file: items/ablation_effect_on_balanced_accuracy.tex
\begin{figure}[tbp]
\centering
\includegraphics[alt={Ablation Effect on Balanced Accuracy},width=\textwidth,height=0.88\textheight,keepaspectratio]{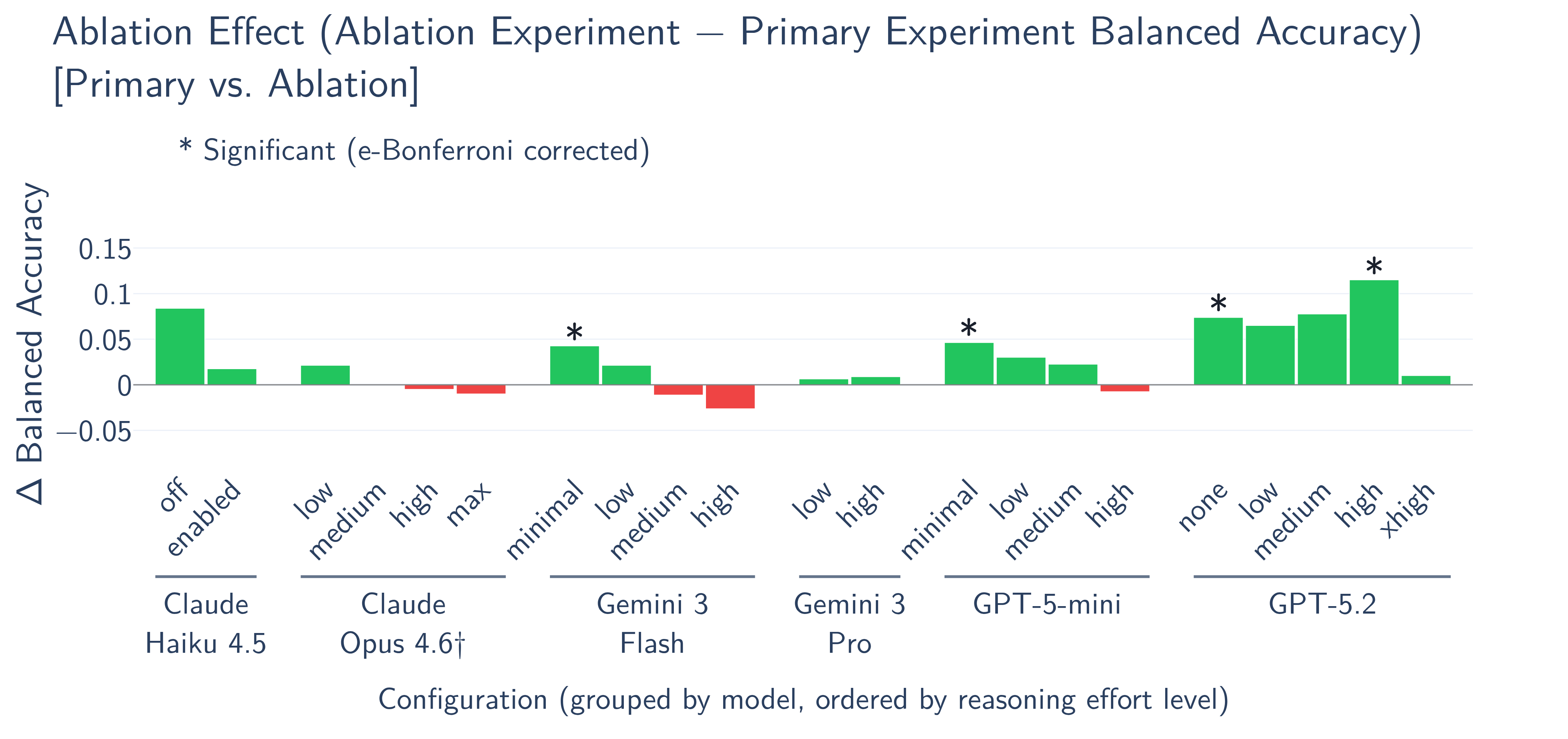}
\ReftCaption{Supplementary~Figure~S3}{Ablation Effect on Balanced Accuracy}{Per-configuration difference in balanced accuracy, the ablation experiment's value minus the primary experiment's, with each bar colored green where that difference is zero or positive and red where it is negative. The dagger (†) marks the authoring model, Claude Opus 4.6†: it helped author the stimulus components, so its results may reflect familiarity with its own output patterns rather than the capability being measured. Removing the error classification labels raised balanced accuracy in 15 of the 21 configurations, left it unchanged in one, and lowered it in five; the largest decrease took Gemini 3 Flash at high reasoning effort from 0.884 to 0.857. The six largest increases belonged to configurations among the seven with the lowest balanced accuracy in the primary experiment. Four configurations carry an asterisk marking significance under the configuration-level cross-dataset test: Gemini 3 Flash at minimal reasoning effort, GPT-5-mini at minimal reasoning effort, GPT-5.2 with reasoning effort set to none, and GPT-5.2 at high reasoning effort. That test bets on the same balanced accuracy difference the bars plot, and its correction combines e-Bonferroni across 21 tests with a factor of 2 for the two directions each test can report. Evidence strength varies widely across those four: the two GPT-5.2 configurations clear the adjusted e-value threshold of 20 by orders of magnitude, while Gemini 3 Flash and GPT-5-mini at minimal reasoning effort clear it at 31.025 and 30.439. The magnitude of the balanced accuracy difference did not by itself determine significance: Claude Haiku 4.5 with reasoning off had the second-largest increase (+0.083), but the comparison was not significant, with an adjusted e-value of 4.740.}
\label{art:supplementary_figure_s3}
\end{figure}

%% file: items/example_condition_accuracy_heatmap_correct_paired.tex
\begin{figure}[tbp]
\centering
\includegraphics[alt={Per-Example Condition Accuracy Heatmap, Correct-Draft},width=\textwidth,height=0.88\textheight,keepaspectratio]{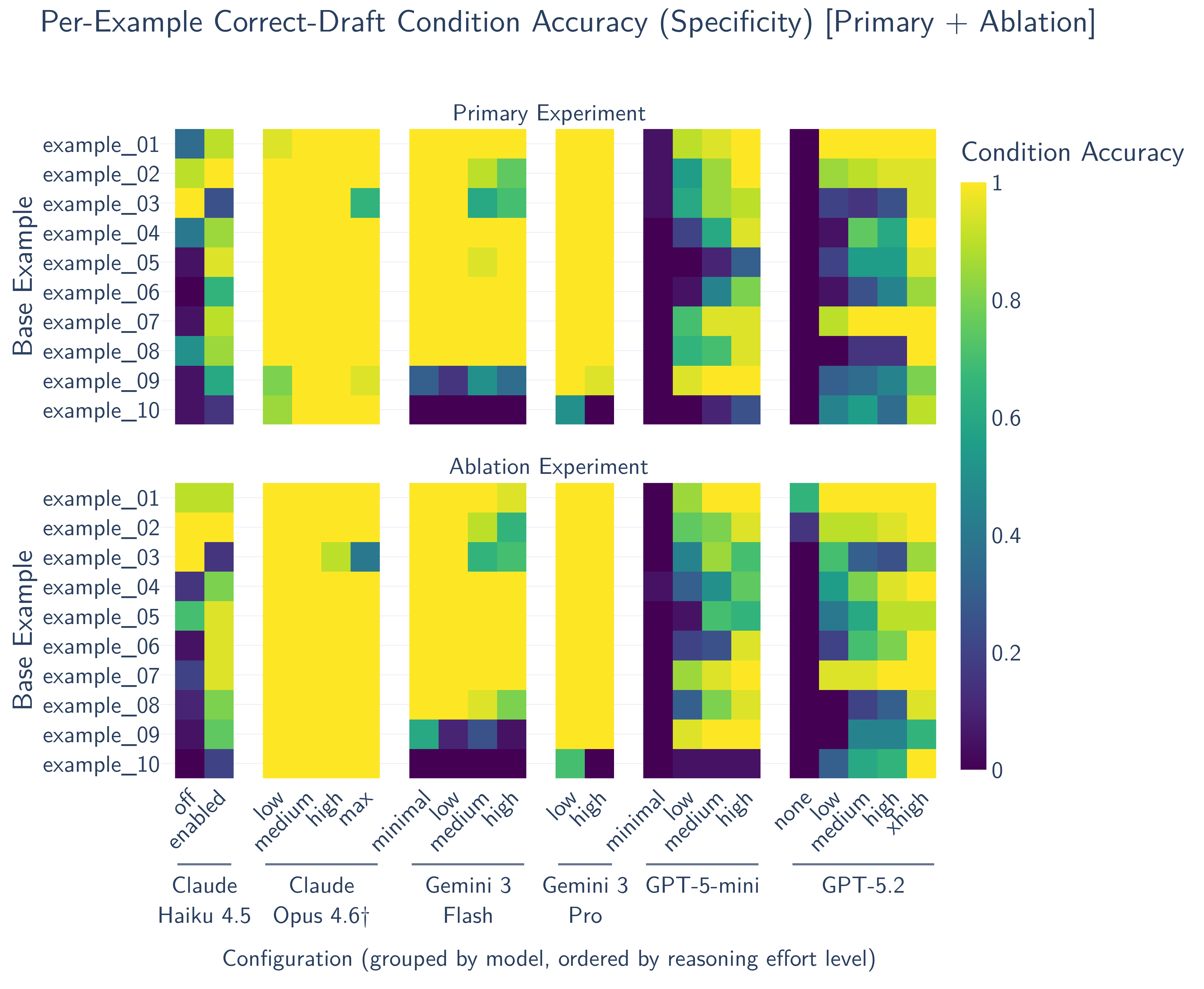}
\ReftCaption{Supplementary~Figure~S4a}{Per-Example Condition Accuracy Heatmap, Correct-Draft}{Per-example correct-draft condition accuracy (specificity) for every configuration, in one labeled matrix for the primary experiment and one for the ablation experiment; each cell pools the 20 trials one configuration ran on one base example's correct-draft stimulus. The dagger (†) marks the authoring model, Claude Opus 4.6†: it helped author the stimulus components, so its results may reflect familiarity with its own output patterns rather than the capability being measured. Averaged across the 21 configurations, per-example correct-draft accuracy ranged from 0.340 on example\_10 to 0.862 on example\_01 in the primary experiment and from 0.362 to 0.917 on those same two base examples in the ablation experiment; even so, every base example reached 1.000 in at least one configuration and 0.000 in at least one other, in both experiments. How much reasoning effort mattered depended on the model. In the primary experiment, correct-draft accuracy across GPT-5.2's five reasoning effort levels spanned a range of at least 0.800 on every base example, falling to 0.000 on all of them with reasoning effort set to none; Claude Opus 4.6† instead scored at least 0.950 for 37 of its 40 combinations of base example and reasoning effort level, so its largest range on any base example was 0.350.}
\label{art:supplementary_figure_s4a}
\end{figure}

%% file: items/example_condition_accuracy_heatmap_transparent_paired.tex
\begin{figure}[tbp]
\centering
\includegraphics[alt={Per-Example Condition Accuracy Heatmap, Transparent},width=\textwidth,height=0.88\textheight,keepaspectratio]{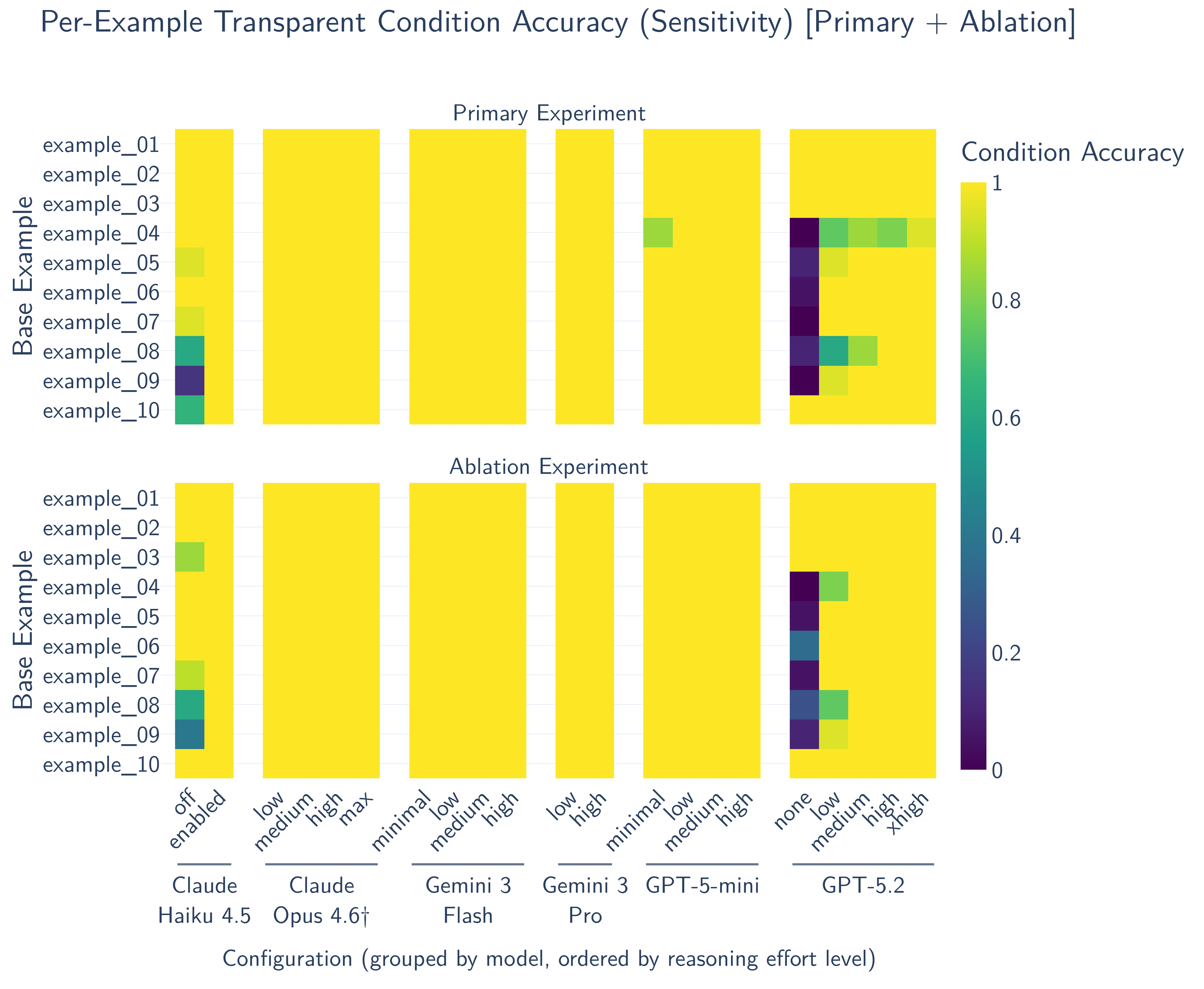}
\ReftCaption{Supplementary~Figure~S4b}{Per-Example Condition Accuracy Heatmap, Transparent}{Per-example transparent condition accuracy (sensitivity) for every configuration, in one labeled matrix for the primary experiment and one for the ablation experiment; each cell pools the 20 trials one configuration ran on one base example's transparent stimulus. The dagger (†) marks the authoring model, Claude Opus 4.6†: it helped author the stimulus components, so its results may reflect familiarity with its own output patterns rather than the capability being measured. On transparent stimuli the grader states the operational interpretation outright, so settling the disagreement requires only following the grader's reasoning. Transparent accuracy sat at or near ceiling almost everywhere, consistent with the grader's explicit reasoning acting as a reliable surface cue. It reached a perfect 1.000 in 190 of the 210 example-by-configuration pairings in the primary experiment and in 197 of 210 in the ablation experiment, and the lowest per-example mean across the 21 configurations was 0.910 in the primary experiment and 0.926 in the ablation experiment.}
\label{art:supplementary_figure_s4b}
\end{figure}

%% file: items/example_condition_accuracy_heatmap_opaque_paired.tex
\begin{figure}[tbp]
\centering
\includegraphics[alt={Per-Example Condition Accuracy Heatmap, Opaque},width=\textwidth,height=0.88\textheight,keepaspectratio]{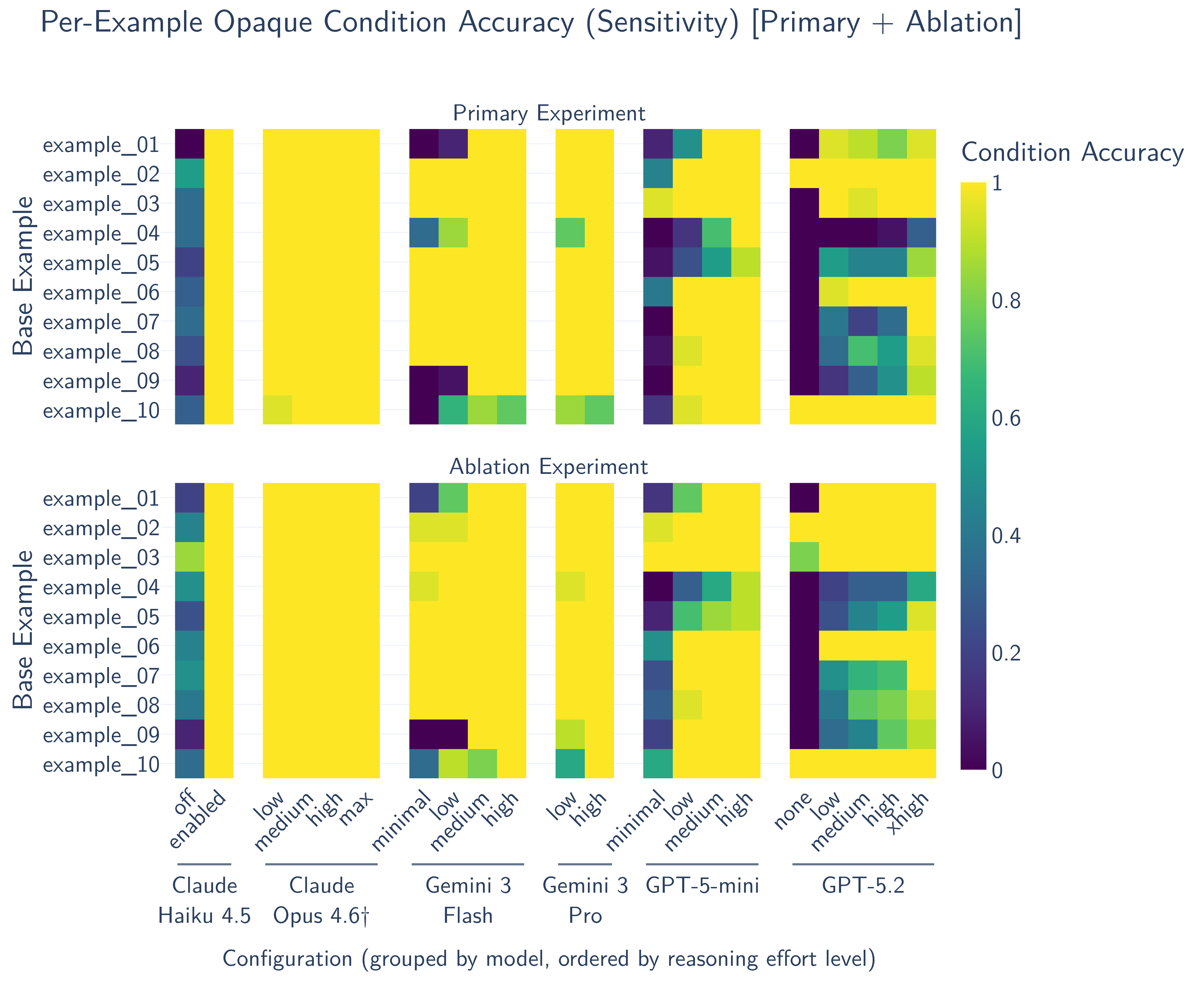}
\ReftCaption{Supplementary~Figure~S4c}{Per-Example Condition Accuracy Heatmap, Opaque}{Per-example opaque condition accuracy (sensitivity) for every configuration, in one labeled matrix for the primary experiment and one for the ablation experiment; each cell pools the 20 trials one configuration ran on one base example's opaque stimulus. The dagger (†) marks the authoring model, Claude Opus 4.6†: it helped author the stimulus components, so its results may reflect familiarity with its own output patterns rather than the capability being measured. On opaque stimuli the grader's analysis is structurally identical to the correct-draft grader's, so that analysis never states the operational interpretation and the meta-evaluator has to arrive at that interpretation on its own. The ablation experiment reran every configuration on stimuli with the grader's error classification labels removed. Opaque accuracy was higher there for 54 of the 210 example-by-configuration pairings and lower for 10; it was unchanged for the remaining 146, 131 of which sat at a perfect 1.000 in both experiments. Averaged across the 21 configurations, example\_04 and example\_09 were the two hardest base examples in both experiments, at 0.595 and 0.667 in the primary experiment and 0.695 and 0.698 in the ablation experiment.}
\label{art:supplementary_figure_s4c}
\end{figure}

%% file: items/ablation_effect_on_condition_accuracy.tex
\begin{landscape}
\centering
\includegraphics[alt={Ablation Effect on Per-Condition Accuracy},width=\linewidth,height=0.87\textwidth,keepaspectratio]{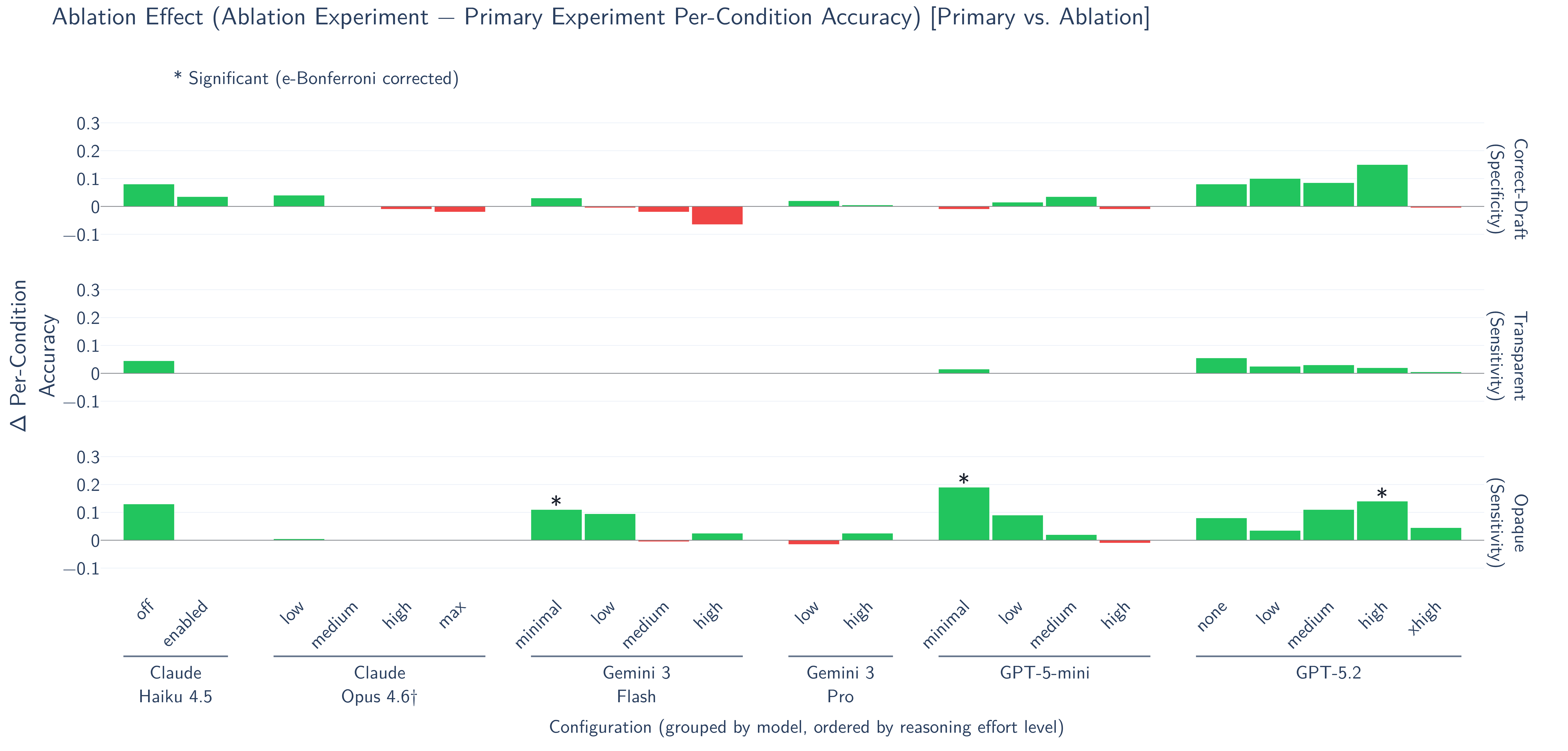}
\ReftCaptionOf{Supplementary~Figure~S5}{Ablation Effect on Per-Condition Accuracy}{Ablation effect on per-condition accuracy for each of the 21 configurations, shown as three stacked rows, one per condition (correct-draft, transparent, and opaque). The dagger (†) marks the authoring model, Claude Opus 4.6†: it helped author the stimulus components, so its results may reflect familiarity with its own output patterns rather than the capability being measured. Each bar is the accuracy delta from the primary experiment to the ablation experiment within that condition, colored green where that difference is zero or positive and red where it is negative, with configurations held in the same order across all three rows. Three configuration-by-condition comparisons reached significance under the condition-level cross-dataset test and are marked with an asterisk, all on opaque stimuli and all favoring the ablation experiment: GPT-5-mini at minimal reasoning effort (+0.190 in per-condition accuracy), GPT-5.2 at high reasoning effort (+0.140), and Gemini 3 Flash at minimal reasoning effort (+0.110). Transparent comparisons were dominated by ceiling effects, so most transparent accuracy deltas were near zero; opaque comparisons carried the largest and most consistent improvements, with the ablation experiment raising accuracy in 14 of the 17 configurations whose opaque accuracy in the primary experiment was below 1.000.}
\label{art:supplementary_figure_s5}
\end{landscape}

%% file: items/reasoning_token_scaling_paired.tex
\begin{figure}[tbp]
\centering
\includegraphics[alt={Reasoning Token Scaling by Effort Level},width=\textwidth,height=0.88\textheight,keepaspectratio]{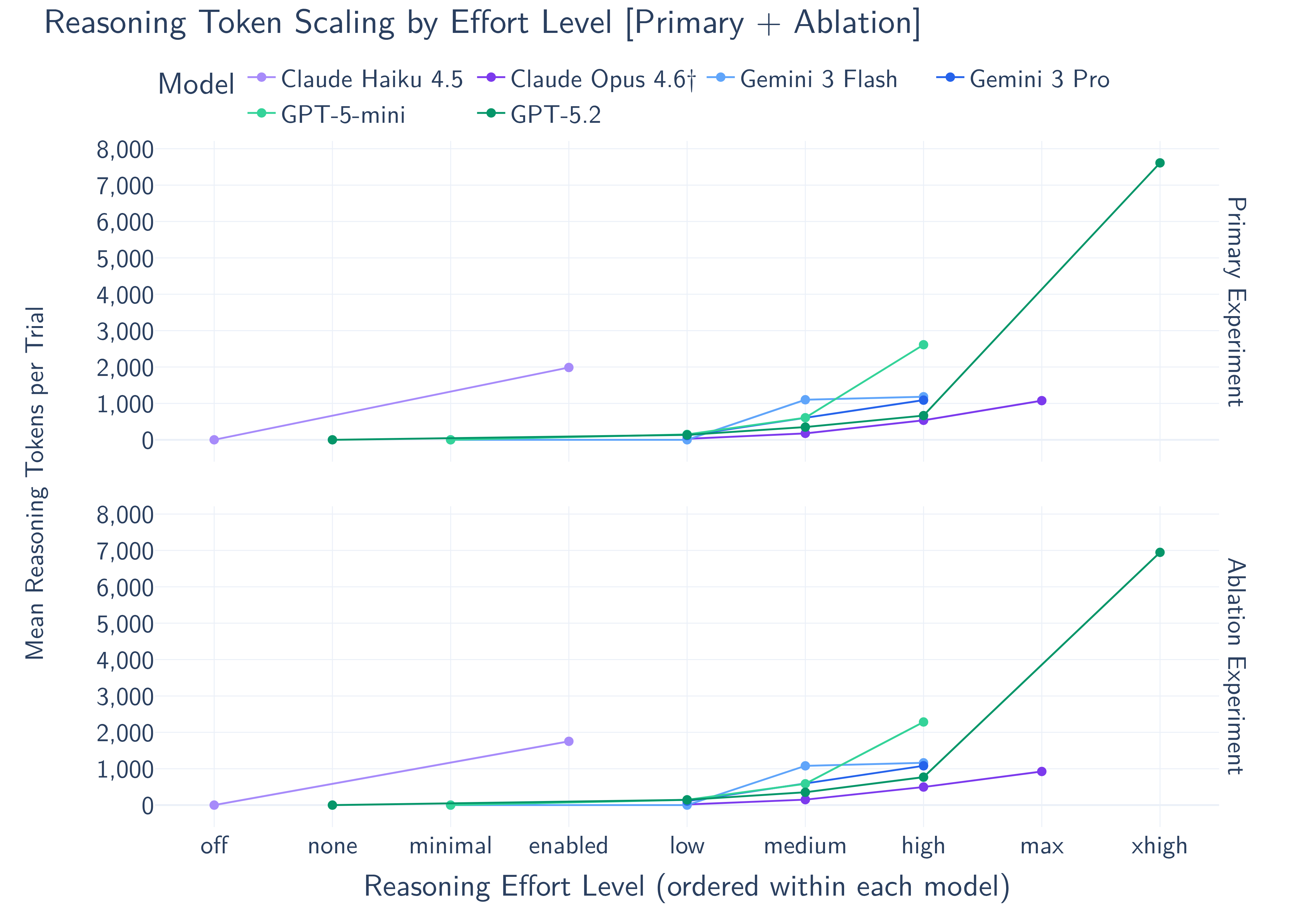}
\ReftCaption{Supplementary~Figure~S6}{Reasoning Token Scaling by Effort Level}{Mean reasoning tokens per trial as a function of reasoning effort level, one line per model, in one labeled row for the primary experiment and one for the ablation experiment. The dagger (†) marks the authoring model, Claude Opus 4.6†: it helped author the stimulus components, so its results may reflect familiarity with its own output patterns rather than the capability being measured. In both experiments, mean reasoning tokens per trial never decreased as a model's reasoning effort level rose, but the peak each model reached varied roughly sevenfold across models. In the primary experiment those peaks ran from 1,076 tokens per trial for Claude Opus 4.6† at max reasoning effort to 7,611 for GPT-5.2 at xhigh reasoning effort, with GPT-5.2 consuming about three times the next-highest peak (2,613 for GPT-5-mini at high reasoning effort). The ablation experiment held the same ordering with every peak lower, from 925 tokens per trial for Claude Opus 4.6† to 6,946 for GPT-5.2.}
\label{art:supplementary_figure_s6}
\end{figure}

%% file: items/token_composition_paired.tex
\begin{figure}[tbp]
\centering
\includegraphics[alt={Token Composition by Configuration},width=\textwidth,height=0.88\textheight,keepaspectratio]{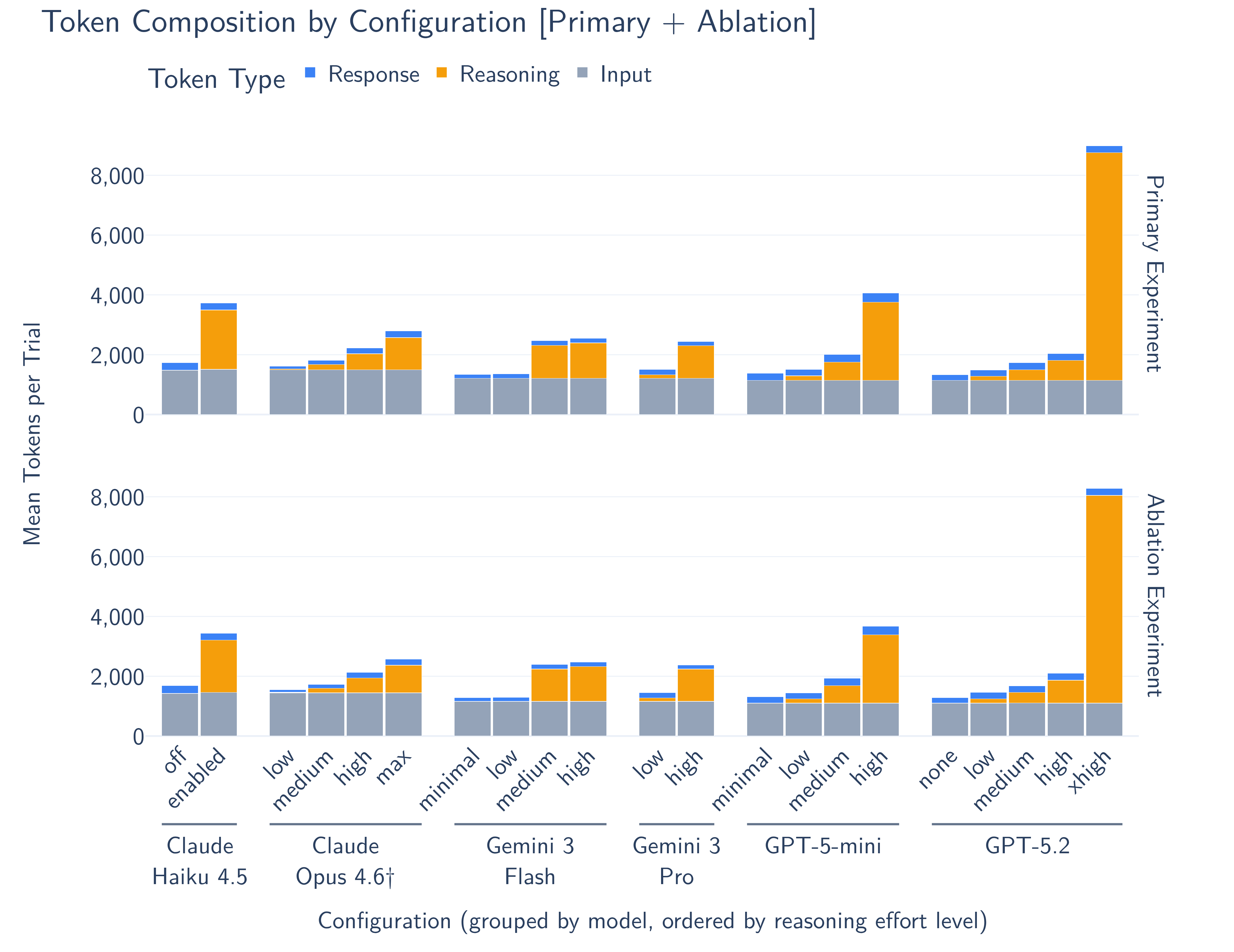}
\ReftCaption{Supplementary~Figure~S7}{Token Composition by Configuration}{Mean per-trial token counts decomposed into \textit{input tokens} (slate), \textit{reasoning tokens} (amber), and \textit{response tokens} (blue), stacked as a single bar per configuration, with configurations grouped by model, in one labeled row for the primary experiment and one for the ablation experiment. The dagger (†) marks the authoring model, Claude Opus 4.6†: it helped author the stimulus components, so its results may reflect familiarity with its own output patterns rather than the capability being measured. Reasoning tokens were the largest of the three components in only 3 of the 21 configurations in the primary experiment, namely Claude Haiku 4.5 with reasoning enabled, GPT-5-mini at high reasoning effort, and GPT-5.2 at xhigh reasoning effort; input tokens were the largest in the other 18. GPT-5.2 at xhigh reasoning effort was the extreme case, with reasoning tokens making up 84.7\% of its per-trial total in the primary experiment (4.6 million reasoning tokens across the 600 trials for that configuration) and 83.9\% in the ablation experiment.}
\label{art:supplementary_figure_s7}
\end{figure}

%% file: items/marginal_cost_per_percentage_point_paired.tex
\begin{figure}[tbp]
\centering
\includegraphics[alt={Marginal Cost per Percentage Point of Balanced Accuracy},width=\textwidth,height=0.88\textheight,keepaspectratio]{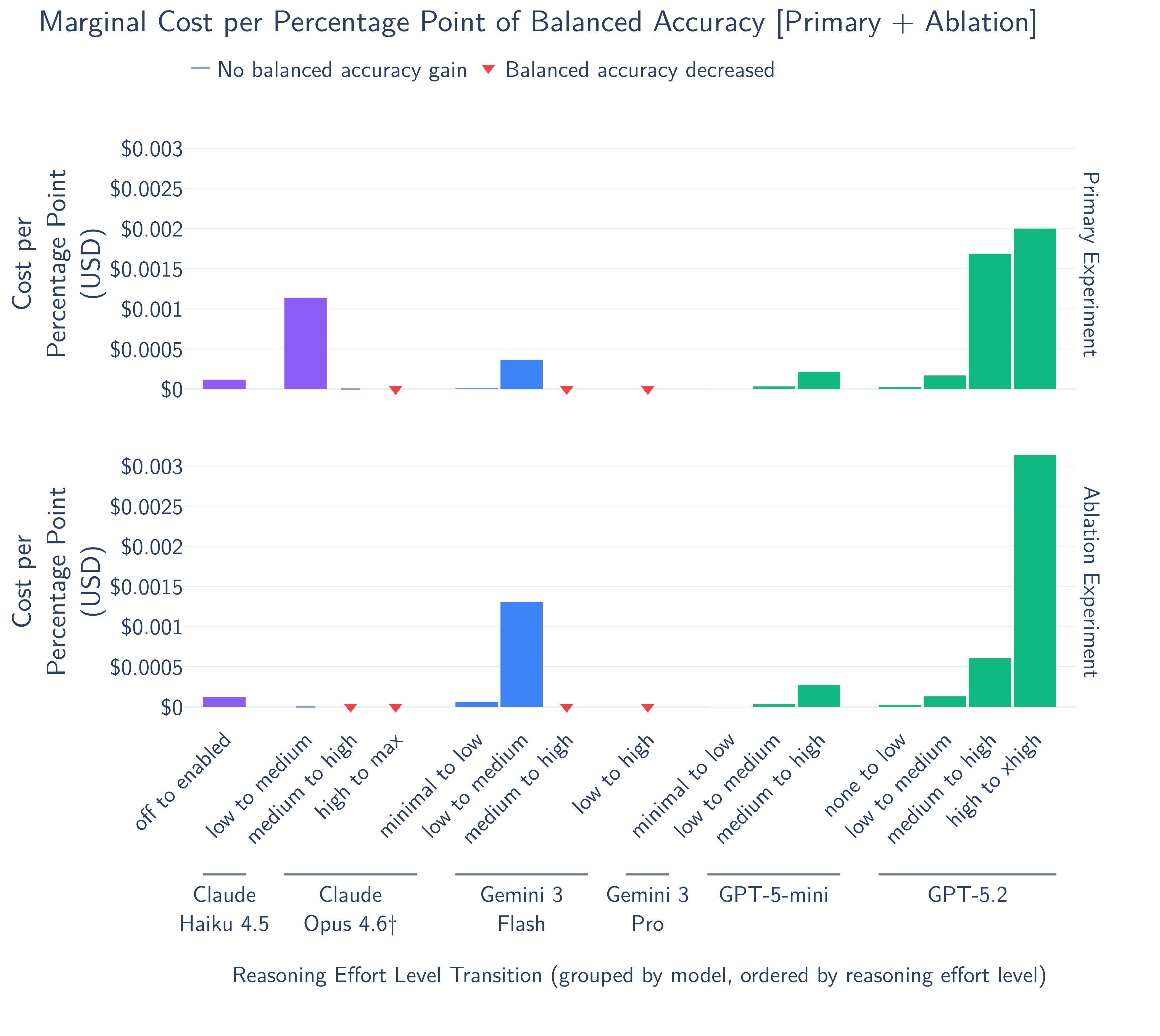}
\ReftCaption{Supplementary~Figure~S8}{Marginal Cost per Percentage Point of Balanced Accuracy}{Marginal cost per percentage point of balanced accuracy for every adjacent reasoning effort level transition within each graduated-reasoning model, in one labeled row for the primary experiment and one for the ablation experiment. Improving transitions render as colored bars (colored by provider); transitions with no change in balanced accuracy render as horizontal dashes at zero; transitions where the upper level reduced balanced accuracy render as downward triangles at zero. The dagger (†) marks the authoring model, Claude Opus 4.6†: it helped author the stimulus components, so its results may reflect familiarity with its own output patterns rather than the capability being measured. GPT-5.2's step from high to xhigh reasoning effort was the primary experiment's most expensive improving transition, at \$0.0020 per percentage point. Three transitions decreased balanced accuracy in the primary experiment: Claude Opus 4.6† from high to max reasoning effort, Gemini 3 Pro from low to high reasoning effort, and Gemini 3 Flash from medium to high reasoning effort. In the ablation experiment, four transitions decreased balanced accuracy: the same three plus Claude Opus 4.6† from medium to high reasoning effort.}
\label{art:supplementary_figure_s8}
\end{figure}

%% file: items/failure_mode_breakdown_correct_draft_paired.tex
\begin{figure}[tbp]
\centering
\includegraphics[alt={Failure Modes by Raw Count, Correct-Draft},width=\textwidth,height=0.88\textheight,keepaspectratio]{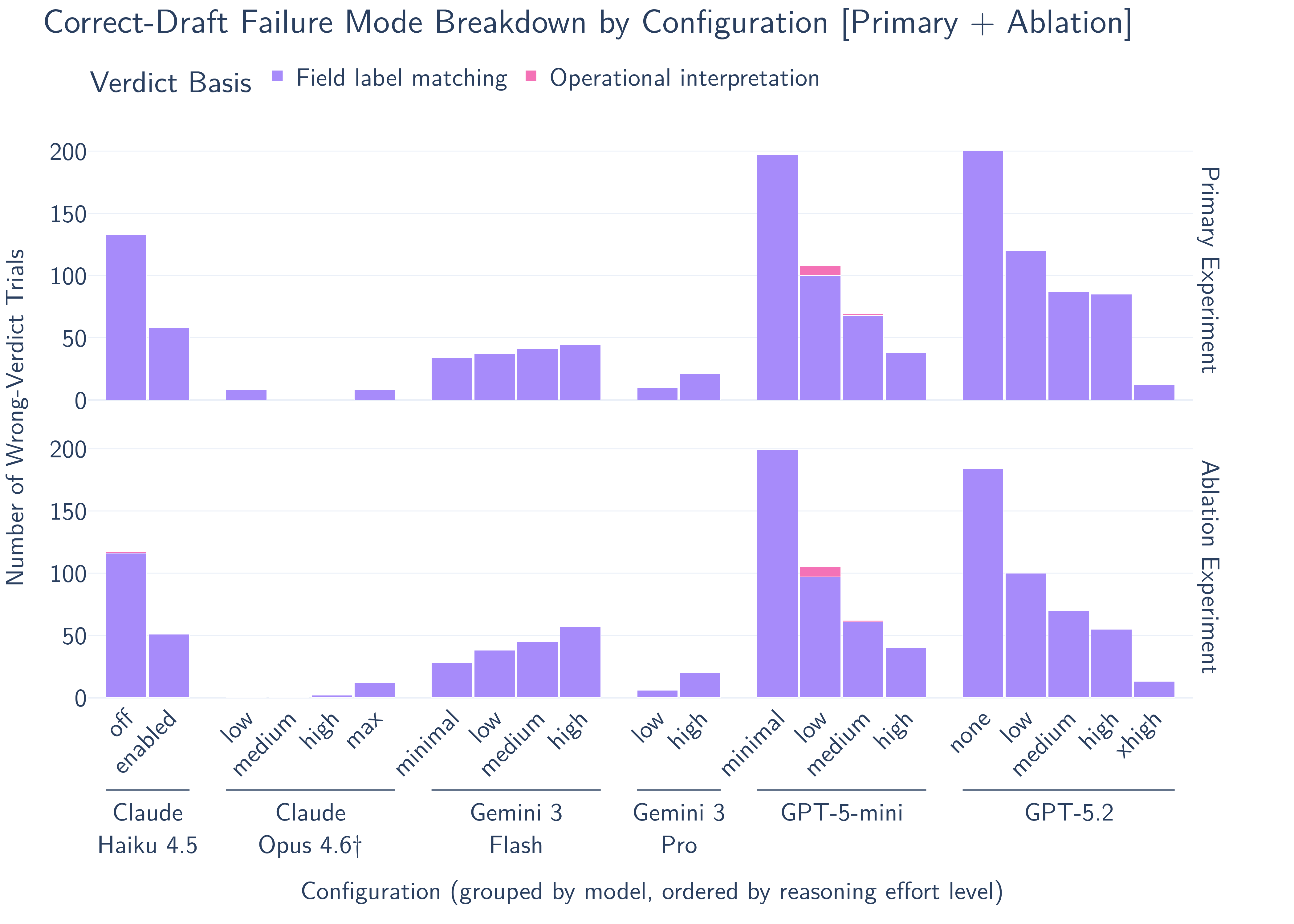}
\ReftCaption{Supplementary~Figure~S9a}{Failure Modes by Raw Count, Correct-Draft}{Per-configuration count of correct-draft trials whose verdict did not match ground truth, stacked by what determined the verdict (\textit{operational interpretation}, pink, upper segment of each bar, versus \textit{field label matching}, violet, lower segment), in one labeled row for the primary experiment and one for the ablation experiment. The dagger (†) marks the authoring model, Claude Opus 4.6†: it helped author the stimulus components, so its results may reflect familiarity with its own output patterns rather than the capability being measured. In the primary experiment, 1,301 of 1,310 correct-draft trials whose verdict did not match ground truth were governed by field label matching; in the ablation experiment, 1,194 of 1,204; in both experiments, trials in which the operational interpretation governed the verdict were vanishingly few in almost every configuration.}
\label{art:supplementary_figure_s9a}
\end{figure}

%% file: items/failure_mode_breakdown_incorrect_draft_paired.tex
\begin{landscape}
\centering
\includegraphics[alt={Failure Modes by Raw Count, Incorrect-Draft},width=\linewidth,height=0.87\textwidth,keepaspectratio]{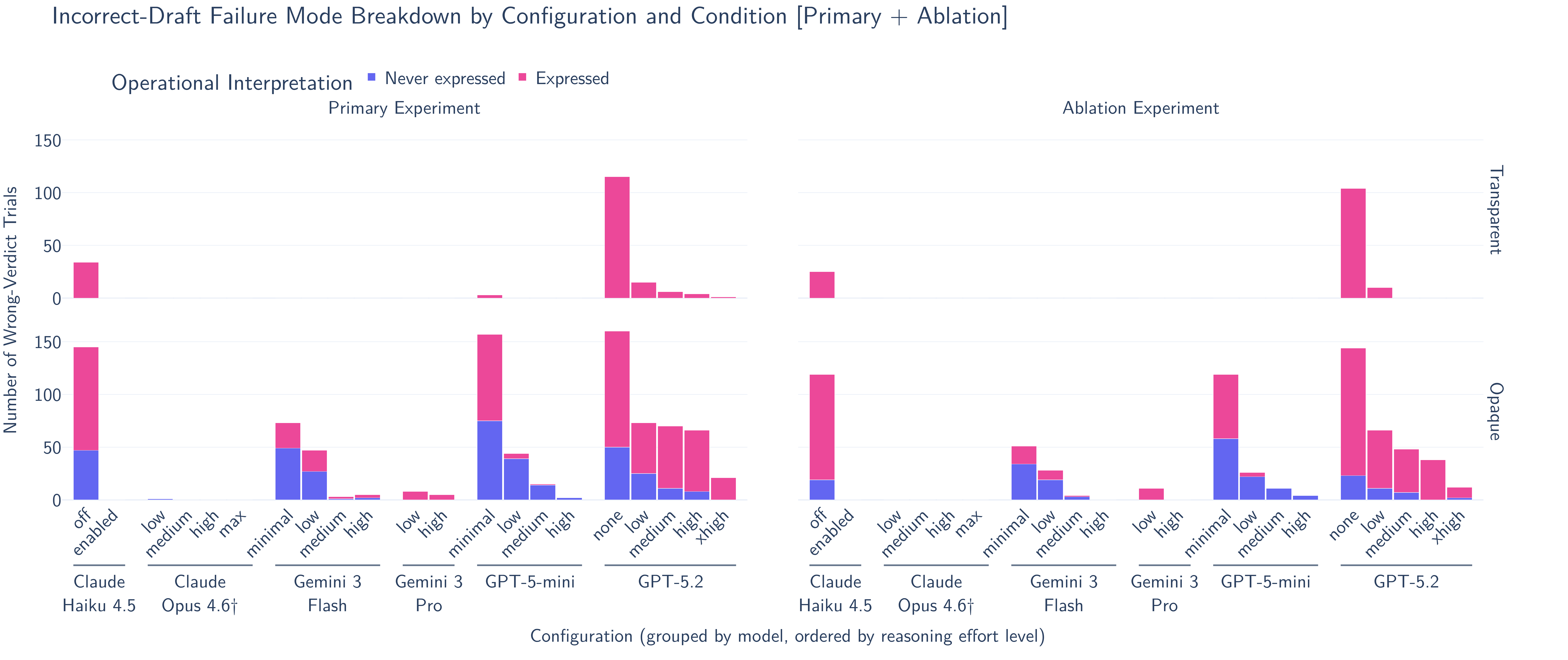}
\ReftCaptionOf{Supplementary~Figure~S9b}{Failure Modes by Raw Count, Incorrect-Draft}{Per-configuration count of incorrect-draft trials whose verdict did not match ground truth, stacked by whether the meta-evaluator \textit{expressed the operational interpretation} (pink, upper segment of each bar) or \textit{never expressed it} (indigo, lower segment), displayed as one row per condition (transparent above opaque) against two labeled columns, one for the primary experiment and one for the ablation experiment. The pink segment counts every trial in which the interpretation was expressed, whether or not it went on to determine the verdict. Among those trials, the operational interpretation determined the verdict on 2 transparent and 23 opaque trials in the primary experiment and on 3 transparent and 41 opaque trials in the ablation experiment. The dagger (†) marks the authoring model, Claude Opus 4.6†: it helped author the stimulus components, so its results may reflect familiarity with its own output patterns rather than the capability being measured. In both experiments the meta-evaluator expressed the operational interpretation on every transparent trial whose verdict did not match ground truth: 178 of 178 in the primary experiment and 139 of 139 in the ablation experiment. On opaque stimuli it never expressed that interpretation on 351 of 895 such trials in the primary experiment and on 213 of 681 in the ablation experiment.}
\label{art:supplementary_figure_s9b}
\end{landscape}

%% file: items/failure_mode_dominance_heatmap_correct_paired.tex
\begin{figure}[tbp]
\centering
\includegraphics[alt={Failure Mode Dominance Heatmap, Correct-Draft},width=\textwidth,height=0.88\textheight,keepaspectratio]{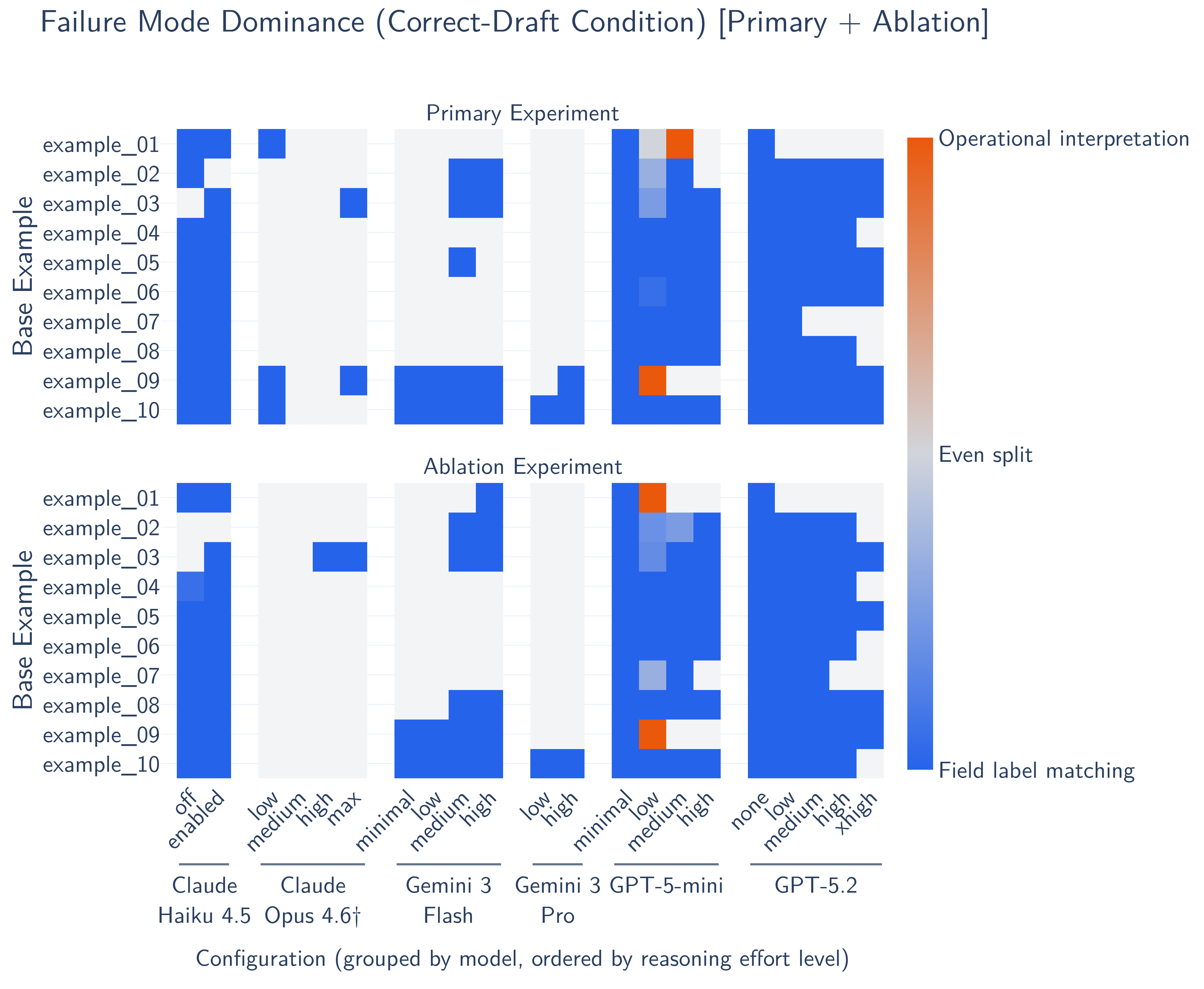}
\ReftCaption{Supplementary~Figure~S10a}{Failure Mode Dominance Heatmap, Correct-Draft}{Which failure mode dominated each example-by-configuration pairing on correct-draft stimuli, in one labeled matrix for the primary experiment and one for the ablation experiment. The dagger (†) marks the authoring model, Claude Opus 4.6†: it helped author the stimulus components, so its results may reflect familiarity with its own output patterns rather than the capability being measured. Only trials whose verdict did not match ground truth are classified, and each is scored by whether field label matching determined the verdict or the operational interpretation did. A cell's position between the two ends of the scale is the split of its classified trials between those two modes, and the palest cells, paler than the scale's midpoint gray, are pairings with no such trials to classify. Field label matching dominated 113 of the 116 pairings that had classified trials in the primary experiment and 109 of 111 in the ablation experiment; the operational interpretation dominated 2 pairings in each, and 1 pairing in the primary experiment was an even split. That pattern follows from correct-draft governing rates of 0.69\% in the primary experiment (9 of 1,310 trials) and 0.83\% in the ablation experiment (10 of 1,204).}
\label{art:supplementary_figure_s10a}
\end{figure}

%% file: items/failure_mode_dominance_heatmap_transparent_paired.tex
\begin{figure}[tbp]
\centering
\includegraphics[alt={Failure Mode Dominance Heatmap, Transparent},width=\textwidth,height=0.88\textheight,keepaspectratio]{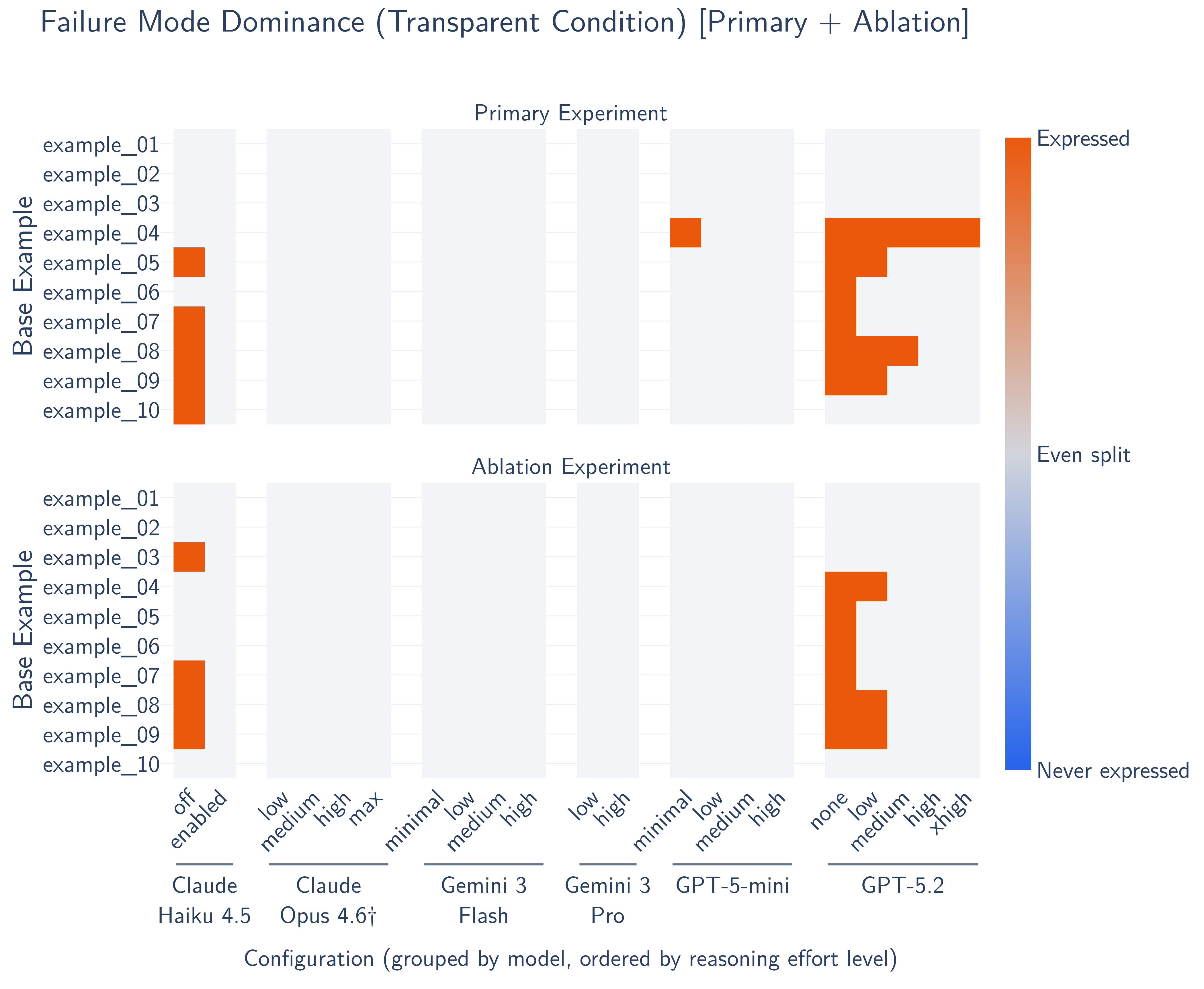}
\ReftCaption{Supplementary~Figure~S10b}{Failure Mode Dominance Heatmap, Transparent}{Which failure mode dominated each example-by-configuration pairing on transparent stimuli, in one labeled matrix for the primary experiment and one for the ablation experiment. The dagger (†) marks the authoring model, Claude Opus 4.6†: it helped author the stimulus components, so its results may reflect familiarity with its own output patterns rather than the capability being measured. Only trials whose verdict did not match ground truth are classified, and each is scored by whether the meta-evaluator expressed the operational interpretation, whether or not it went on to determine the verdict. A cell's position between the two ends of the scale is the split of its classified trials between those two modes, and the palest cells, paler than the scale's midpoint gray, are pairings with no such trials to classify. On transparent stimuli the meta-evaluator expressed the operational interpretation on every classified trial, a 100\% articulation rate in both experiments, so every pairing with classified trials sits at the expressed end of the scale: 20 pairings in the primary experiment and 13 in the ablation experiment. The failure on these stimuli was therefore never an absence of the operational interpretation; it lay in what the meta-evaluator did after expressing that interpretation.}
\label{art:supplementary_figure_s10b}
\end{figure}

%% file: items/failure_mode_dominance_heatmap_opaque_paired.tex
\begin{figure}[tbp]
\centering
\includegraphics[alt={Failure Mode Dominance Heatmap, Opaque},width=\textwidth,height=0.88\textheight,keepaspectratio]{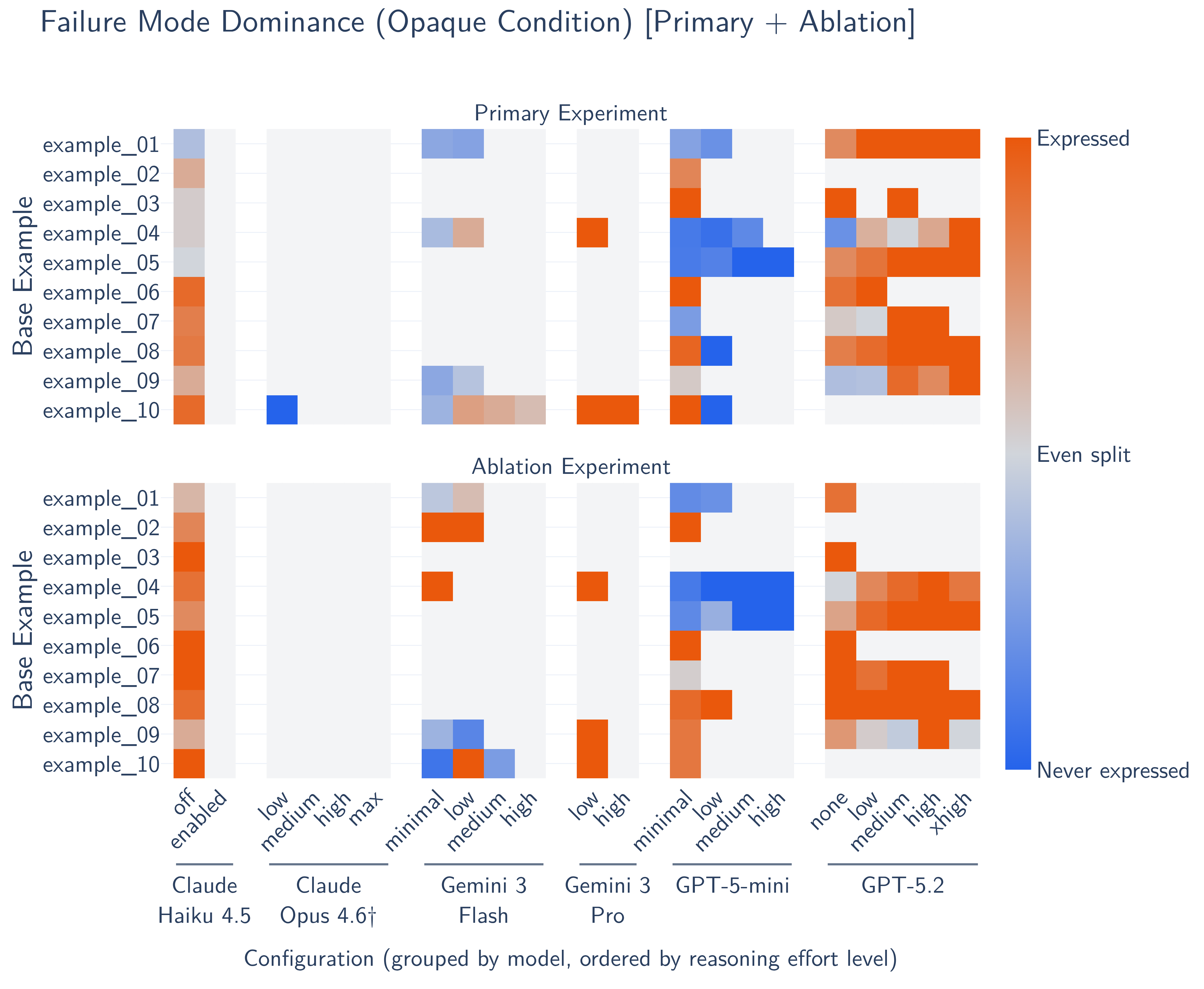}
\ReftCaption{Supplementary~Figure~S10c}{Failure Mode Dominance Heatmap, Opaque}{Which failure mode dominated each example-by-configuration pairing on opaque stimuli, in one labeled matrix for the primary experiment and one for the ablation experiment. The dagger (†) marks the authoring model, Claude Opus 4.6†: it helped author the stimulus components, so its results may reflect familiarity with its own output patterns rather than the capability being measured. Only trials whose verdict did not match ground truth are classified, and each is scored by whether the meta-evaluator expressed the operational interpretation, whether or not it went on to determine the verdict. A cell's position between the two ends of the scale is the split of its classified trials between those two modes, and the palest cells, paler than the scale's midpoint gray, are pairings with no such trials to classify. In the primary experiment, expressing the operational interpretation dominated 49 of the 75 pairings that had classified trials, never expressing it dominated 23, and 3 were even splits; the corresponding counts across the ablation experiment's 67 such pairings were 49, 16, and 2. Among the 63 pairings with classified trials in both experiments, 34 moved toward expressed, 12 moved toward never expressed, and 17 did not move, consistent with the opaque articulation rate rising from 60.78\% in the primary experiment (544 of 895 classified trials) to 68.72\% in the ablation experiment (468 of 681).}
\label{art:supplementary_figure_s10c}
\end{figure}